\documentclass{article} %
\usepackage{iclr2025_conference,times}

\usepackage[utf8]{inputenc} %
\usepackage[T1]{fontenc}    %
\usepackage{hyperref}       %
\usepackage{url}            %
\usepackage{booktabs}       %
\usepackage{amsfonts}       %
\usepackage{nicefrac}       %
\usepackage{microtype}      %
\usepackage{xcolor}         %

\usepackage{wrapfig}
\usepackage{multicol,lipsum}
\usepackage{subcaption}

\usepackage{multirow}
\usepackage{algorithm}
\usepackage{algorithmic}
\usepackage{enumitem}
\usepackage{amsmath}
\usepackage{amssymb}
\usepackage{mathtools}
\usepackage{amsthm}
\definecolor{blue-violet}{rgb}{0.54, 0.17, 0.89}
\definecolor{caribbeangreen}{rgb}{0.0, 0.8, 0.6}
\definecolor{deepmagenta}{rgb}{0.8, 0.0, 0.8}
\definecolor{amber(sae/ece)}{rgb}{1.0, 0.49, 0.0}
\definecolor{applegreen}{rgb}{0.55, 0.71, 0.0}
\definecolor{cornflowerblue}{rgb}{0.39, 0.58, 0.93}
\definecolor{ao(english)}{rgb}{0.0, 0.5, 0.0}
\definecolor{lightmagenta}{HTML}{e799d9}
\definecolor{lightblue}{HTML}{7ba4db}
\definecolor{lightcaribbeangreen}{HTML}{a0ced9}
\definecolor{urlcolor}{HTML}{3e8976}
\hypersetup{colorlinks,linkcolor={blue},citecolor={ao(english)},urlcolor={black}} 

\renewcommand{\algorithmiccomment}[1]{#1}   %

\usepackage[capitalize,noabbrev]{cleveref}

\theoremstyle{plain}
\newtheorem{theorem}{Theorem}[section]

\theoremstyle{definition}
\newtheorem{definition}[theorem]{Definition}

\theoremstyle{remark}

\usepackage[toc,page,header]{appendix}
\usepackage{minitoc}

\title{Optimistic Games \\ for Combinatorial Bayesian Optimization \\ with Application to Protein Design}

\author{Melis Ilayda~Bal\textsuperscript{\textmd{1}}\thanks{Corresponding author. Code is available at \url{https://github.com/melisilaydabal/gameopt}.}  , Pier Giuseppe Sessa\textsuperscript{\textmd{2}}, Mojm\'{i}r Mutn\'{y}\textsuperscript{\textmd{3}}, Andreas Krause\textsuperscript{\textmd{3}}\\
    \textsuperscript{\textmd{1}}Max Planck Institute for Intelligent Systems, T\"{u}bingen, Germany \\
    \textsuperscript{\textmd{2}}Google DeepMind  \quad \textsuperscript{\textmd{3}}ETH Z\"{u}rich \\
    \texttt{mbal@tuebingen.mpg.de, piergs@google.com} \\ \texttt{mmutny@inf.ethz.ch, krausea@ethz.ch}
}

\iclrfinalcopy %
\begin{document}
\doparttoc %
\faketableofcontents %

\maketitle

\begin{abstract}
  
Bayesian optimization (BO) is a powerful framework to optimize black-box expensive-to-evaluate functions via sequential interactions. 
In several important problems (e.g. drug discovery, circuit design, neural architecture search, etc.), though, such functions are defined over large \emph{combinatorial and unstructured} spaces. This makes existing BO algorithms not feasible due to the intractable maximization of the acquisition function over these domains. 
To address this issue, we propose \textsc{GameOpt}, a novel game-theoretical approach to combinatorial BO.
\textsc{GameOpt} establishes a cooperative game between the different optimization variables, and selects points that are game \emph{equilibria} of an upper confidence bound acquisition function. These are stable configurations from which no variable has an incentive to deviate -- analog to local optima in continuous domains. Crucially, this allows us to efficiently break down the complexity of the combinatorial domain into individual decision sets, making \textsc{GameOpt} scalable to large combinatorial spaces. %
We demonstrate the application of \textsc{GameOpt} to the challenging \emph{protein design} problem and validate its performance on four real-world protein datasets. Each protein can take up to $20^{X}$ possible configurations, where $X$ is the length of a protein, making standard BO methods infeasible.
Instead, our approach iteratively selects informative protein configurations and very quickly discovers highly active protein variants compared to other baselines.
\end{abstract}

\section{Introduction}
\label{sec:intro}
\vspace{-0.2cm}
Many scientific and engineering problems such as drug discovery \citep{drug-discovery}, neural architecture search \citep{neural-architecture-search}, or circuit design \citep{circuit-design} require optimization of expensive-to-evaluate black-box functions over combinatorial unstructured spaces involving binary, integer-valued, and categorical variables. 
As a concrete example, consider the \emph{protein design} problem, \textit{i.e.}, finding the optimal amino acid sequence to maximize the functional capacity (fitness) of the protein. Such fitness functions are highly complex, one can, in most cases, only be elucidated from real-world protein synthesis experiments. 
Moreover, exhaustive exploration is infeasible for both traditional lab methods
and computational techniques \citep{NavigatingFitLandscape-krause} due to \textit{combinatorial explosion}: %
a typical protein has $300$ amino acid sites, each to be filled with one of twenty natural amino acids, yielding $20^{300}$ candidate variants.\looseness=-1

\looseness -1 Bayesian optimization (BO) is an established framework for optimizing black-box functions with the goal of minimizing the number of evaluations needed to certify optimality \citep{bo-1975}. BO constructs a probabilistic surrogate model as a representation of the underlying black-box function, e.g., using Gaussian Processes (GPs) \citep{GP}. Then, it iteratively selects the next evaluations typically by maximizing a designated acquisition function. The BO framework has proven to be very powerful and successful in a variety of real-world problems including material discovery~\citep{bo-material}, adaptive experimental design~\citep{bo-adaptive-exp}, or drug discovery \citep{chembo, bo-seq-ae}.
When considering combinatorial domains, however, standard BO methods are intractable since maximizing the acquisition function requires an exhaustive search over the whole combinatorial space (\textit{e.g.} of size $20^{300}$ in the context of proteins) without further assumptions.

\begin{figure*}[t]
    \centering  
    {\includegraphics[width=1\linewidth,trim=20 230 0 0,clip]{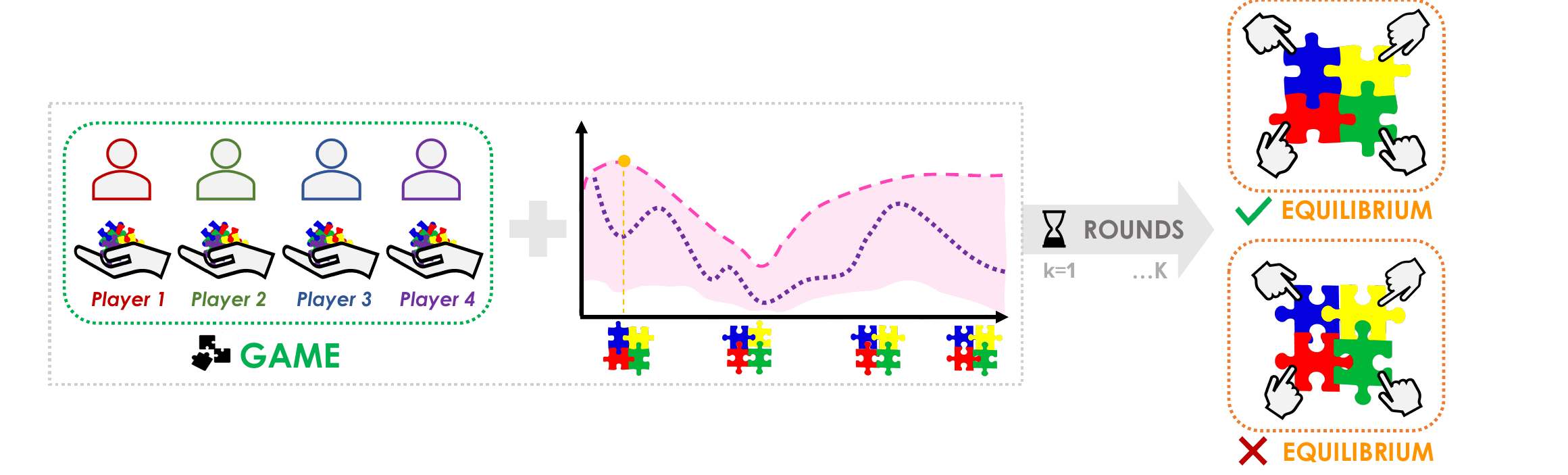}}
\setlength{\abovecaptionskip}{-10pt}
\setlength{\belowcaptionskip}{-15pt}
 \caption{Illustration of \textsc{GameOpt}. \textsc{GameOpt} defines a game among the decision variables, where game rewards are represented by the upper confidence bound (UCB) function. This decouples the combinatorial decision space into individual decision sets and allows \textsc{GameOpt} to \emph{tractably} compute game equilibria. These can be thought of as local optima of the AF in unstructured domains.}
\label{fig:game-opt}
\end{figure*}

To address this challenge, we propose \textsc{GameOpt}, a novel game-theoretical framework for combinatorial BO.  
To circumvent the intractable maximization of an acquisition function, \textsc{GameOpt} defines a cooperative game between the discrete domain variables and, at each interaction round, selects informative points to be game \emph{equilibria} of the acquisition function. These are stable configurations from which no player (variable) has the incentive to deviate. They can be thought of as a formalization of a notion of local optima of the acquisition function in unstructured domains (i.e., domains lacking a lattice structure). Crucially, these can be computed employing well-known equilibrium finding subroutines, effectively simulating a repeated game among the players. In this work, we utilize the Upper Confidence Bound (UCB) acquisition function, which represents an \emph{optimistic} estimate of the underlying objective and was shown to efficiently balance exploration with exploitation~\citep{gpopt-bandit-krause}. For an overview of the method, see Figure \ref{fig:game-opt}. 

\vspace{-0.2cm}
\paragraph{Contributions} 
We make the following main contributions:
\begin{itemize}[left=0cm]
  \setlength\itemsep{-0.2em}
  \vspace{-0.2cm}
    \item We propose \textsc{GameOpt}, a \textbf{novel} game-theoretical BO framework for large combinatorial and unstructured search spaces.
    \textsc{GameOpt} computes informative evaluation points as the equilibria (\textit{i.e.}, local optima) of a cooperative game between the discrete variables. This overcomes the scalability issues of maximizing acquisition functions over combinatorial domains and provides a \textbf{tractable} optimization. \textsc{GameOpt} is a \textbf{flexible} procedure where the resulting per-iteration game can be solved by any readily available game strategy or solver. 
    \item \looseness -1  Under common kernel regularity assumptions, we bound the sample complexity of \textsc{GameOpt}, quantifying the gap between the computed equilibria (of the surrogate UCB function) and those of the underlying unknown objective. Given target accuracy level $\epsilon$, \textsc{GameOpt} returns $\epsilon$-approximate equilibria after $T=\Omega(\gamma_T \epsilon^{-2})$ iterations, where $\gamma_T$ is the kernel-dependent maximum information gain~\citep{gpopt-bandit-krause}.
    \item We apply \textsc{GameOpt} to the challenging \textbf{protein design} problem, involving search spaces of categorical inputs. There, \textsc{GameOpt} advances the protein design process by mimicking natural evolution via a game between protein sites. We \textbf{experimentally} validate its performance on several \textbf{real-world} protein design problems based on human binding protein GB1~\citep{gb1-4-dataset, gb1-55-dataset}, iron-dependent halogenase \citep{halogenase-dataset} and green-fluorescent protein \citep{gfp, gfp-dataset}.
    \textsc{GameOpt} converges consistently faster, \textit{i.e.}, it requires fewer BO iterations to identify highly binding protein variants compared to baseline methods such as classical directed evolution.
\vspace{-0.2cm}
\end{itemize}
\vspace{-0.2cm}
\section{Problem Statement and Background}\vspace{-0.2cm}
\label{sec:problem-statement-background}
\paragraph{Problem statement}
We consider the problem of optimizing a costly-to-evaluate, black-box function $f:\mathcal{X} \to \mathbb{R}$ over a combinatorial unstructured space $\mathcal{X}$ without a lattice form. Suppose each element $x \in \mathcal{X}$ can be represented by $n$ discrete variables $(x^1, x^2, \dots, x^n)$ ($n$-dimensional), where each $x^i$ takes values from a set $\mathcal{X}^{(i)}$, 
this makes the domain of $n \geq 1$ variables
$\mathcal{X} = \mathcal{X}^{(1)} \times \dots \mathcal{X}^{(n)}$. Assuming $|\mathcal{X}^{(i)}|=d, \: \forall i$, the size of the combinatorial space $\mathcal{X}$ is $d^n$. 
However, the proposed \textsc{GameOpt} framework can also operate under varying $|\mathcal{X}^{(i)}|$ sizes, as we detail in Appendix~\ref{app:grouping-players-different-dimensions}.

As a concrete motivating example, consider the protein design problem (Section~\ref{sec:analysis}). There, $f(x)$ represents the fitness value of the designed protein sequence $x$. The search space size is $|\mathcal{X}| = 20^{n}$, with $n$ protein sites and $|\mathcal{X}^{(i)}|=20$ amino acid choices per site. Moreover, a (noisy) evaluation $f(x)$ is a labor-intensive process, requiring extensive efforts and specialized laboratory equipment. \looseness=-1

\paragraph{Gaussian Processes (GPs)} \looseness -1 \vspace{-0.1cm}
Bayesian Optimization~\citep{bo-1975} is a versatile framework for optimizing complex, noisy, and expensive-to-evaluate functions. BO leverages Bayesian inference to model the underlying function with a surrogate, \textit{e.g.}, a Gaussian Process (GP) and iteratively selects evaluation points that are the most informative in terms of reducing uncertainty or enhancing model performance.
\looseness -1

Formally, a Gaussian Process $\mathcal{GP}(\mu(\cdot),k(\cdot,\cdot))$ over domain $\mathcal{X}$ is specified by a prior mean function $\mu(x): \mathcal{X} \to \mathbb{R}$ and a covariance function $k(x,x^{\prime}): \mathcal{X} \times \mathcal{X} \to \mathbb{R}$, denoted by
$
f(x) \sim \mathcal{GP}(\mu(x),k(x,x^{\prime}))
$,
where $f(x)$ represents the function value at input $x$. 

Given a set of observed data points $X_{t}$ up to iteration $t$ and their corresponding vector of noisy observations $Y_{t} = f(X_{t}) + \epsilon_{t} $ with Gaussian noise $\epsilon_{t} \sim \mathcal{N}(0, \sigma_{t}^{2})$, and a GP prior defined by $\mathcal{GP}(\mu_{t}(x),k_{t}(x,x^{\prime}))$, the posterior distribution of the GP at iteration $t+1$ given new observations $X_{\dagger}$ is again Gaussian $p(f_{t+1} \mid X_{t}, X_{\dagger}, Y_{t}) = \mathcal{N}(\mu_{t+1}, \sigma^{2}_{t+1})$ with posterior mean and variance~\citep{GP}.
\looseness=-1
\vspace{-0.1cm}
\paragraph{Bayesian Optimization (BO)} \looseness -1 \vspace{-0.2cm}
To maximize $f$, BO algorithms iteratively select evaluation points so as to balance exploration and exploitation. At each iteration the method selects the maximizer of an \emph{acquisition function}, for example, the widely-adopted Upper-Confidence Bound (UCB) \citep{gpopt-bandit-krause} function. Given a $\mathcal{GP}$ model at iteration $t$, the UCB function is defined as
\begin{equation}
\label{eqn:ucb}
\text{UCB}_{t}(\mathcal{GP},x) = \mu_{t}(x) + \beta_t \sigma_{t}(x),
\end{equation}
where $\mu(x)$ and $\sigma(x)$ are the posterior mean and standard deviation at point $x$ according to $\mathcal{GP}$, and $\beta_t \in \mathbb{R}$ is a confidence parameter influencing the width of the set that can be selected to ensure the validity of the confidence set.
The UCB function defines an \emph{optimistic} estimate of the underlying objective $f$, and can effectively balance exploration (\textit{i.e.}, favoring points with large uncertainty $\sigma_{t}(x)$) with exploitation (\textit{i.e.}, selecting points with large posterior mean $\mu_{t}(x)$).\looseness=-1

While standard BO methods can efficiently optimize $\text{UCB}(\mathcal{GP},\cdot)$ in efficiently enumerable or continuous domains, they become very soon intractable in the case of combinatorial unstructured domains, such as the space of possible amino acid sequences.
In the next section, we propose \textsc{GameOpt}, a novel BO approach that circumvents such prohibitive difficulty.\looseness=-1
\vspace{-0.2cm}
\begin{algorithm}[t]
\caption{\textsc{GameOpt}}
\label{alg:optimistic-game}
\begin{algorithmic}[1]
\STATE {\bfseries Input:} GP prior $\mathcal{GP}^0(\mu_{0}, k(\cdot,\cdot))$, initial data $\mathcal{D}_{0}=\{(x_{i}, y_{i}=f(x_{i})+\epsilon)\}$, batch size $B \in \mathbb{N}$,  $M \in \mathbb{N} > B$, parameter $\beta$.
\FOR{iteration $t = 1,2, \dots, T$}
\STATE Construct game with reward function $\text{UCB}(\mathcal{GP}^{t-1}, \beta, \cdot): \prod_{i=1}^n \mathcal{X}^{(i)} \to \mathbb{R}$
\STATE Compute $M$ equilibria $\{x_{t,i}\}_{i=1}^M$ of the above.  \hfill {\color{amber(sae/ece)} */ Equilibrium-finding subroutine}
\STATE Select batch of top $B$ equilibria $\{x_{t,i}\}_{i=1}^B$ according to $\text{UCB}(\mathcal{GP}^{t-1},\beta, \cdot)$. \hfill {\color{amber(sae/ece)} */ Filtering}
\STATE Obtain evaluations $y_{t,i} = f(x_{t,i}) + \epsilon_{t,i}, \quad \forall i =1,\ldots, B$  
\STATE Update $\mathcal{D}_{t} \gets \mathcal{D}_{t-1} \cup  \{(x_{t,i}, y_{t,i})\}_{i=1}^B$
\STATE Posterior update of model $\mathcal{GP}^t$  with $\mathcal{D}_{t}$.
\ENDFOR
\end{algorithmic}
\end{algorithm}
\section{ \textsc{GameOpt} algorithm }
\label{sec:proposed-approach}
\vspace{-0.2cm}
In a nutshell, the proposed \textsc{GameOpt} (Optimistic Games) approach circumvents the combinatorial optimization of the $\textsc{UCB}$ function by defining a \emph{cooperative game} among the  $n$ input variables and computes the associated equilibria as candidate evaluation points.  
Formally, at each iteration $t$, \textsc{GameOpt} defines a cooperative game \citep{game-theory} involving $\mathcal{N} = \{1, \ldots, n\}$ players, each player $i$ taking actions in the discrete set $\mathcal{X}^{(i)}$. In such a game, the players' interests are aligned towards the goal of maximizing the function $\textsc{UCB}(\mathcal{GP}^t, \cdot): \prod_{i=1}^n \mathcal{X}^{(i)}\rightarrow \mathbb{R}$, where $\mathcal{GP}^t$ is the current GP estimate at iteration $t$. Thus, it can be interpreted as an optimistic game with respect to the true unknown $f$. In such a game, the goal of the players is to compute game (Nash) \emph{equilibria}, defined as follows. \looseness -1

\begin{definition}[Nash equilibrium~\citep{Nash1951}] Let $r^i: \mathcal{X} \rightarrow \mathbb{R}$ be the reward function of each player~$i$, defined over joint configuration $x$.
A joint strategy profile \(x_\text{eq} = (x_\text{eq}^1, \ldots, x_\text{eq}^n)\) is a Nash equilibrium if, for every player $i \in \mathcal{N}$, $r^i(x_\text{eq}^i, x_\text{eq}^{-i}) \geq r^i(x^i, x_\text{eq}^{-i}), \forall x^{i} \in \mathcal{X}^{(i)}$, where $x_\text{eq}=(x_{\text{eq}}^{i}, x_{\text{eq}}^{-i})$, and $x_{\text{eq}}^{-i}$ is the joint equilibrium strategy of all players except $i$.
\end{definition}

The existence of such equilibrium point(s) is guaranteed since players and actions are finite~\citep{game-theory}. Moreover, because players' reward functions are aligned and coincide with $\textsc{UCB}(\mathcal{GP}^t, \cdot)$, efficient polynomial-time equilibrium-finding methods can be employed, such as Iterative Best-Response (\textsc{IBR}), where players update their actions sequentially, or simultaneous multiplicative weights updates such as the \textsc{Hedge}~\citep {hedge} algorithm. 
We report these two possible strategies in Algorithms~\ref{alg:opt-game-ibr} and \ref{alg:opt-game-hedge} in Section~\ref{sec:equilibrium-subroutines}. 
Intuitively, equilibria are computed by breaking down the complex decision space into individual decision sets, as illustrated in Figure~\ref{fig:game-opt}. Mathematically, we refer to this operation as $\arg\operatorname{eq}$, returns the joint configuration(s) which form equilibria according to UCB, \looseness=-1
\begin{equation}\label{eq:argeq}
    x_{\text{eq}} = \arg\operatorname{eq}_{x^i \in \mathcal{X}^{(i)}; i\in \mathcal{N}} \textsc{UCB}(\mathcal{GP}^t, (x^1, \dots, x^n)).
\end{equation}

\vspace{-0.15cm}
Our overall approach is summarized in Algorithm~\ref{alg:optimistic-game}. In practice,  we compute $M>1$ equilibria and subselect a batch of top $B<M$ equilibria according to the $\textsc{UCB}(\mathcal{GP}^t, \cdot)$ criterion. 
Subsequently, such a batch is evaluated by $f$, the GP model is updated accordingly, and a new game with an updated reward function is defined at the next iteration based on the updated posterior.\looseness=-1

\paragraph{A form of local optimality}  \looseness -1 \vspace{-0.1cm}
Within \textsc{GameOpt}, each player strategically selects actions to maximize their collective payoff, much like seeking local optima in a continuous multi-dimensional function (see Figure~\ref{fig:game-opt}). In continuous optimization, a local optimum is a point, where there is no direction that leads to an improvement, similarly, as in our framework there is not a player that can unilaterally improve the value of the collective pay-off. In essence, seeking equilibria is analogous to seeking local optima of a continuous acquisition function, and our game-based approach allows us to effectively pinpoint them within an unstructured combinatorial space.
We remark that \textsc{GameOpt} computes equilibria of the current $\textsc{UCB}(\mathcal{GP}^t, \cdot)$ function which, as we show in Section~\ref{sec:analysis}, are better and better approximations of equilibria of the unknown objective $f$.
\vspace{-0.2cm}
\paragraph{Price of Anarchy} \looseness -1
But how good are equilibria compared to the global optimum? The quality of equilibria (also known as the efficiency of the game) can be quantified via the game-theoretic notion of Price of Anarchy (PoA)~\citep{Christodoulou2005}, defined as the ratio between the worst equilibrium and the global optimum, \textit{i.e.}, $\text{PoA}:={\min_{\mathbf{x} \in \mathcal{E}}f(\mathbf{x})}/{\max_{\mathbf{x}}f(\mathbf{x})}$ where $\mathcal{E}$ is the set of all equilibria of $f$. PoA has been extensively studied for various classes of games and can sometimes be upper-bounded given further assumptions on $f$. As an example, in case $f$ is a submodular function (over binary, integer, or continuous domains), PoA is guaranteed to be at least 0.5~\citep{Vetta02,sessa19bounding}. Although such a PoA guarantee does not readily apply to our setting, we believe similar ones could be proved for the case of unstructured domains — though this is beyond the scope of our work. 
In practice, given an unknown function $f$ (such as the protein’s fitness function in our experiments of Section~\ref{sec:analysis}), not all equilibria may achieve high function values (i.e. PoA can be very low). Nevertheless, \textsc{GameOpt} computes \emph{multiple} equilibria ($M>1$) at each iteration and selects only the top $B$ according to their UCB value. We believe this is a key form of robustness that can effectively filter out suboptimal equilibria and empower \textsc{GameOpt}'s experimental performance. 
\vspace{-0.2cm}
\subsection{Equilibrium finding subroutines}  
\label{sec:equilibrium-subroutines} \looseness -1 
\vspace{-0.2cm}
We present a set of established algorithms for finding an equilibrium of the game introduced in Eq.~\eqref{eq:argeq}. \looseness -1

\paragraph{Iterative best responses} \looseness -1 \vspace{-0.15cm} One possible subroutine for Algorithm~\ref{alg:optimistic-game} is Iterative Best Response (IBR) procedure as provided in Algorithm \ref{alg:opt-game-ibr}. 
Concretely, under the cooperative game setting outlined in Section~\ref{sec:proposed-approach} and given $\mathcal{GP}$-predicted UCB function, each player \textit{iteratively} selects the response that maximizes the value of the game given that the other players play the joint strategy from the previous round. Each player is sequentially selected to play their best response in a round-robin fashion. Because action space is finite, this procedure is guaranteed to converge to a local maximum of the UCB function \textit{i.e.}, an equilibrium of the underlying game ~\citep{game-theory}.
\vspace{-0.2cm}
\paragraph{Multiplicative weights updates} \looseness -1  Alternatively, we can compute game equilibria letting players \textit{simultaneously} act according to a multiplicative weights update algorithm such as \textsc{Hedge}~\citep{hedge}, see ~\cref{alg:opt-game-hedge}. We can cast equilibrium computation as an instance of \emph{adversarial online learning} among multiple learners~ \citep{pred-learning-games}. Here, each player selects a strategy based on their available options and, after observing the joint payoff, players' strategies are re-weighted based on past performance. Through repeated rounds of play and re-weighting, the empirical frequency of play forms a coarse correlated equilibrium (CCE) (a weaker notion of Nash equilibrium), see e.g.~\citep{pred-learning-games}, while convergence to pure Nash equilibria is also guaranteed in some cases~\citep{congestion-games, mw-congestion}. \vspace{-0.2cm}
\begin{figure}[t]
\vspace{-0.5cm}
\begin{minipage}[t]{0.45\textwidth}
\begin{algorithm}[H]
\caption{Iterative Best Response (\textsc{IBR})}
\label{alg:opt-game-ibr}
\begin{algorithmic}[1]
\STATE {\bfseries Input:} Domain $\mathcal{X}$, payoff (reward) \\ $r: \mathcal{X} \to \mathbb{R}$, players $\mathcal{N}$.
\STATE $\bold{x}^{br}_{0} \gets$ random joint strategy,  $ \bold{x}^{br}_{0} \in \mathcal{X}$.
\FOR[\hfill{\color{amber(sae/ece)} */ BR game}]{round $k=1,\dots,K$}
\STATE $\mathcal{X}^{br}_{k} \gets \Big\{(x^{i,br}, x_{k-1}^{-i, br}), \text{ such that } \newline  x^{i,br} = \arg\max\limits_{x\in \mathcal{X}^{(i)}} r(x, x_{k-1}^{-i, br}) \Big\}_{i=1}^n$ 
\STATE Play $\bold{x}^{br}_{k} \gets \arg \max_{x_{k} \in \mathcal{X}^{br}_{k}} [r(x_{k})]$
\ENDFOR
\STATE \textbf{return} $\bold{x}^{br}_{K}$  \hfill {\color{amber(sae/ece)} */ Equilibrium}
\end{algorithmic}
\end{algorithm} 
\end{minipage}
\hspace{0.05cm}
\begin{minipage}[t]{0.54\textwidth}
\begin{algorithm}[H]
\caption{Simultaneous \textsc{Hedge}}
\label{alg:opt-game-hedge}
\begin{algorithmic}[1]
\STATE {\bfseries Input:} Domain $\mathcal{X} = \prod_{i=1}^{n}\mathcal{X}^{(i)} \text{ with } \mid \mathcal{X}^{(i)} \mid = W$, payoff $r: \mathcal{X} \to \mathbb{R}$, players $\mathcal{N}$, parameter $\eta$.
\STATE Initialize weights $\boldsymbol{w}_{1} \gets \frac{1}{W}[1,\dots, 1] \in \mathbb{R}^{\mid \mathcal{N} \mid \times W}$
\FOR[\hfill{\color{amber(sae/ece)} */ Compute CCE}]{round $k = 1, \dots K$} 
\STATE Sample $x^{i}_{k} \sim \boldsymbol{w}^{i}_{1}, \forall i \in \mathcal{N} $
\STATE Set joint strategy $\bold{x}_{k} \gets \{ x^{i}_{k} \}_{i \in \mathcal{N}}$
\FOR[\hfill{\color{amber(sae/ece)} */ Players' payoff}]{player $i \in \mathcal{N}$} 
\STATE $\ell_{x^{-i}_{k}} \gets  [r(x^{j,-i}_{k})]_{\forall j \in \mathcal{X}^{(i)}}$, where \newline $x^{j,-i}_{k} = {j} \cup \{ x^{i^{\prime}}_{k} \}_{i^{\prime} \in \mathcal{N} \setminus \{i\}}, \forall j \in \mathcal{X}^{(i)}$ 
\STATE Set $\boldsymbol{w}^{i}_{k+1} \propto \boldsymbol{w}^{i}_{k} \text{ exp}(\eta \ell_{x^{-i}_{k}})$ 
\ENDFOR
\ENDFOR
\STATE \textbf{return} $\text{Uniform}\{\bold{x}_1, \ldots, \bold{x}_K\}$  \hfill {\color{amber(sae/ece)} */ Equilibrium}
\end{algorithmic}
\end{algorithm}
\end{minipage}
\vspace{-0.5cm}
\end{figure}
\vspace{-0.2cm}
\subsection{Related work} \looseness -1 \vspace{-0.2cm}
\label{sec:related-work-comb-bo}
While there exist rather few works in the area \citep{bounce-bo-combo-mix}, existing combinatorial BO methods either target surrogate modeling with discrete variables \citep{bo-combinatorial, bo-combinatorial-graph, dealing-categorical-bayesopt, bayesopt-set-kernels, bo-dictionary} or optimizing acquisition function within discrete spaces \citep{bo-combinatorial, opt-discrete-expensive-eval, bayesopt-over-hybrid, mercer-features-comb-bayesopt, antibody-design-combo}.
However, they often require a parametric surrogate model with higher-order interaction specifications for combinatorial structures \citep{bo-combinatorial} or domain-specific knowledge \citep{opt-discrete-expensive-eval}.
In contrast, \textsc{GameOpt} relies on a non-parametric surrogate model, without the need for domain-specific knowledge. 

Closest to ours is \citep{bo-prob-reparam}, which also targets optimizing the acquisition function in high-cardinality discrete/mixed search spaces via a probabilistic reparameterization (PR) that maximizes the expectation of the acquisition function.
However, PR fails at being tractable since it requires evaluating the expectation over the joint distribution of all decision variables, requiring combinatorially many elements to be summed. An accurate estimate would require extensive sampling without special structural assumptions.
In contrast, \textsc{GameOpt} treats each variable \emph{independently} (potentially in parallel) within the game, keeping the values of the remaining variables fixed during each strategy update. 
We use PR as a baseline to evaluate our approach in Section \ref{sec:analysis}, and demonstrate improved performance of our method on protein design problems.

Further, a body of research has focused on BO over continuous (latent) spaces \citep{gomez2018automatic, eriksson2019scalable, tripp2020sample, deshwal2021combining, maus2022local, bo-seq-ae, zhang2023learning}. These methods learn continuous sequence embeddings and optimize with gradient-based techniques by utilizing deep generative models.
However, the primary problem we address in our study is the intractable acquisition function optimization over \textit{large combinatorial} search spaces, specifically tackling the challenge of exhaustive exploration. In line with this, we select our baselines accordingly and include a comparison with some latent space optimizers only as additional experimental evaluation for insight. \looseness -1

Recently, the interplay between BO and game theory has been explored by the line of works \citep{sessa2019no, sessa2022efficient, combo-expert-advice, han2024no}, but its connection with combinatorial BO is novel. \looseness=-1
\vspace{-0.2cm}
\section{Sample-Complexity Guarantees}
\vspace{-0.2cm}
In this section, we derive sample-complexity guarantees for \textsc{GameOpt}. Namely, we characterize the number of interaction rounds $T$ required to reach approximate equilibria (\textit{i.e.}, local optima) of the true function $f$. For simplicity, we assume \textsc{GameOpt} is run with batch size $B=1$, though our results can be generalized to larger $B$.

The obtained guarantees are based on standard regret bounds of Bayesian optimization adapted to our equilibrium finding goal. These are characterized by the widely utilized notion of maximum information gain~\citep{gpopt-bandit-krause}: 
\begin{equation}
\gamma_t = \frac{1}{2} \log  \left|I_t + \sigma^{-1}\boldsymbol{K}_t \right|.
\end{equation}
This is a kernel-dependent ($\boldsymbol{K}_t$) quantity that quantifies the maximal uncertainty reduction about $f$ after $t$ observations. 
Further, to characterize our sample complexity, we define the notion of $\epsilon$-approximate Nash equilibrium.

\begin{definition}[$\epsilon$-approximate Nash equilibrium] A strategy profile $\tilde{x}_\text{eq}$ is a $\epsilon$-approximate (Nash) equilibrium of $f$ if, for each $i\in \mathcal{N}$, $f(\tilde{x}_\text{eq}) \geq f(x^{i}, \tilde{x}_\text{eq}^{-i}) - \epsilon$, $\forall x^{i} \in \mathcal{X}^{(i)}$.
\end{definition} 

In the next main theorem, we provide a lower bound on the number of iterations $T$ to reach approximate equilibria. After $T$ rounds, we assume \textsc{GameOpt} returns $x_{T^\star}$ with:
$$T^\star := \arg\min_{t\in [T]} \max_{i, x^i} \left[ \text{UCB}(\mathcal{GP}^t, x^i, x^{-i}_t) - \text{LCB}(\mathcal{GP}^t, x_t) \right], $$
where LCB is the lower confidence bound function $\text{LCB}(\mathcal{GP}^t, x) = \mu_t(x) - \beta_t \sigma_t(x)$. That is, among the selected points $x_1, \ldots, x_T$, $x_{T^\star}$ is the one that guarantees the minimum worst-case single-player deviation. The deviation above is computed according to the UCB and with respect to the LCB, thus representing an upper bound on the actual deviation in terms of $f$. We can affirm the following.
\begin{theorem}[Sample complexity of \textsc{GameOpt}]\label{thm:thm_sample_complexity}
    Assume $f$ satisfies the regularity assumptions of Section~\ref{sec:problem-statement-background}, and \textsc{GameOpt} is run with confidence width $\beta_t = 2n\log\left(\sup_{i\in \mathcal{N}}|\mathcal{X}_i|\frac{t^2\pi^2}{6\delta}\right)$. Then, with probability at least $1-\delta$ and for a given accuracy $\epsilon \geq 0$, the strategy $x_{T^\star}$ returned by \textsc{GameOpt} is a $\epsilon$-approximate Nash equilibrium when 
    \begin{equation}
        T \geq \Omega\left(\frac{\beta_T \gamma_T}{\epsilon^2}\right).
    \end{equation}
\end{theorem}\vspace{-0.2cm}
An equivalent interpretation of the above result is as follows: After $T$ iterations, \textsc{GameOpt} returns an $\epsilon_T$-approximate Nash equilibrium of $f$, with approximation factor $\epsilon_{T} \leq \mathcal{O}(T^{-\frac{1}{2}} \sqrt{\beta_T \gamma_{T}})$. Note that the latter bound is the typical rate of convergence of BO algorithms~\citep{gpopt-bandit-krause} to the global maximizer. Instead, in our combinatorial BO setup --where global optimization is intractable-- it corresponds to the rate of convergence to equilibria. A more explicit convergence guarantee can be obtained by employing existing bounds for $\gamma_T$ which are known for commonly used kernels ~\citep{gpopt-bandit-krause}.
E.g., for squared exponential kernels $\gamma_T= \mathcal{O}(\log(T)^{nd})$ where $d$ is the dimension of each input space $\mathcal{X}^{(i)}$ for each player $i \in \mathcal{N}$, with $| \mathcal{N}| = n$. 
\vspace{-0.2cm}
\section{Application to Protein Design} \looseness -1 \vspace{-0.2cm}
\label{sec:analysis}
In this section, we specialize the \textsc{GameOpt} framework to protein design, a problem defined over the space of possible amino acid sequences. Note that such domains are highly combinatorial (their size grows exponentially with the sequence length) and unstructured (i.e. they lack a lattice structure). In this context, computing game equilibria follows the natural principle of promoting beneficial mutants and mirrors the proteins' mutation and selection process. 
In Algorithms~\ref{alg:evolution-opt-ibr} and ~\ref{alg:evolution-opt-hedge} (Appendix~\ref{app:evolution-opt}), we provide a detailed elaboration of \textsc{GameOpt} for protein design using equilibrium-finding methods. We showcase its performance in four real-world protein datasets.

In the protein design context, \textsc{GameOpt} establishes a cooperative game among the different protein sites $i \in \{1,\ldots, n\}$, where $n$ is the length of the protein sequence. 
Each site $i$ chooses an amino acid from the set with $\mid \mathcal{X}^{(i)} \mid = 20$, 
where the switching between amino acid choices can be thought of as biological \textit{mutation}. The joint objective of the players is to converge to a highly rewarding protein sequence, as measured by the GP-predicted optimistic score for the fitness function. This mirrors the \textit{selection} phase in \textit{evolutionary search}, providing a \textit{directed} approach to protein optimization. Below, we discuss related work in the area.

\textbf{Related work on evolutionary search.} \looseness -1 
A considerable line of works \citep{directedevol-arnold1998, cma-evolution, directedevol-pfl, machine-directed-evolution, opt-discrete-expensive-eval, odbo-dataset-paper, evol-bayesopt-constrained-mo} centers around evolutionary search algorithms for optimizing black-box functions. Within combinatorial amino-acid sequence spaces, 
the highly regarded technique, 
\textit{directed evolution} \citep{directedevol-arnold1998, directedevol-pfl}, draws inspiration from natural evolution and identifies local optima through a series of repeated random searches, characterized by controlled iterative cycles of mutation and selection. Expanding upon this, machine learning-guided variants \citep{machine-directed-evolution, wittmann2021advances, NavigatingFitLandscape-krause, population-seq-design}  mitigate the sample-inefficiency and intractability concerns associated with directed evolution. In general, these methods are not data-driven in the sense they do not use the whole extent of the past data and focus on the best variant found so far or a selection of thereof and propose a random search from thereon. Alternatively, even if allowed to adapt to the past outcomes, they tend to restrict themselves to very small search spaces \citep{halogenase-dataset}. Instead, our approach uses all past data to create a UCB estimate of the fitness landscape and utilize it to simulate a cooperative evolution in problems where even the whole sequence of the protein can be optimized. Moreover, compared to such methods, \textsc{GameOpt} mimics evolution \emph{within each interaction} using the surrogate UCB function.
\vspace{-0.2cm}
\subsection{Datasets} \looseness -1 
\label{sec:datasets}
We empirically evaluate \textsc{GameOpt} on real-world protein design problems, specifically focusing on the following instances: protein G domain B1, \emph{GB1}, binding affinity to an antibody IgG-FC (KA), examined on two distinct datasets, \emph{GB1(4)} \citep{gb1-4-dataset} and \emph{GB1(55)} \citep{gb1-55-dataset}, characterized by sequence lengths of $4$ and $55$, respectively; three critical amino acid positions in an iron/$\alpha$-ketoglutarate-dependent halogenase with sequence length $3$ \citep{halogenase-dataset}; and \textit{Aequorea victoria} green-fluorescent protein (\emph{GFP}) of length $238$ \citep{gfp, gfp-dataset}.
The former \emph{GB1} dataset is fully combinatorial, \textit{i.e.}, covering fitness measurements of $20^{4}$ variants. Here, each protein site is treated as a player in the \textsc{GameOpt}. 
The latter is non-exhaustive, including only 2-point mutations of \emph{GB1}. Thus, an MLP having $R^2=0.93$ on a test set is trained and treated as the ground truth fitness for the fully combinatorial dataset. For \emph{GB1(55)}, we also consider a modified setup where ``only'' $10$ sites can be mutated. 
Similarly, \emph{Halogenase} and \emph{GFP} are also non-exhaustive involving fitness measurements for $605$ and $35,584$ unique variants, respectively. 
To obtain the complete protein fitness landscape, we once again construct oracles for these datasets, utilizing MLPs achieving $R^2=0.96$ and $R^2=0.90$ on their respective test sets. In the case of the \emph{Halogenase} dataset, each protein site is treated as a player, while for the \emph{GFP} dataset, $6$ and $8$ sites are designated as players.
Further experimental details are in Appendix~\ref{app:experiment-details}. 

\vspace{-0.2cm}
\subsection{Experimental setup}
\label{sec:exp-setup}\looseness -1 
In all experiments, we use a GP surrogate with an RBF kernel for GP-based methods. The RBF specifies lengthscales for each input variable separately -- sometimes known as ARD kernels \citep{GP}. To handle categorical inputs to the GP surrogate, we employ feature embeddings as representations for these inputs using the ESM-1v transformer protein language model by \citep{ESM-1v}. The prior mean for the GP is pre-defined as the average log fitness value over the whole dataset.
Kernel hyperparameters are optimized prior to the start of optimization and remain fixed throughout the BO iterations; specifically,  lengthscales are optimized over the training set at the start of each replication using Bayesian evidence, and the outputscale is fixed to the difference between the maximum fitness value observed in the dataset \& mean. In other words, we also fit a prior mean. 
A consistent 
observation noise of $0.0004$ is maintained for each training example. Moreover, we use batch size $B=5$.
In Appendix~\ref{app:experiment-details}, we provide the (hyper)parameter settings (see Table~\ref{tab:params-and-hyperparams}) and the detailed setup for the experiments.
\vspace{-0.2cm}
\subsection{Baselines}  \looseness -1 
We benchmark \textsc{GameOpt} against the following baselines: 
\begin{enumerate}[left=0.2cm]
  \setlength\itemsep{-0.1em}
    \item \textsc{GP-UCB}~\citep{gpopt-bandit-krause} selecting --at each iteration-- the best $B$ points in terms of UCB value. Note that this is feasible (though computationally expensive) only for the \emph{GB1(4)} and \emph{Halogenase} datasets, while it is prohibitive for \emph{GB(55)} and \emph{GFP} 
    \item \textsc{IBR-Fitness}, which mimics directed evolution \citep{directedevol-arnold1998} through a series of local searches on the fitness landscape, iteratively selecting the $B$ best-responses based on log fitness criterion 
    \item \textsc{PR} \citep{bo-prob-reparam}, a state-of-the-art discrete/mixed BO approach picking $B$ points using the expected UCB criterion 
    \item \textsc{Random} baseline randomly sampling $B$ random sequences at each iteration.
\end{enumerate}Further details and pseudo codes of such baselines are in Appendix~\ref{app:baselines}.

We assess our method using two key metrics: convergence speed and sampled batch diversity w.r.t. past, (\textit{i.e.}, the degree of distinctiveness among newly acquired samples in comparison to the original data point particularly in the context of the input space) for BO evaluation. The latter can also be regarded as the measure of exploration. 
Convergence speed is tracked by the log fitness value of the best-so-far discovered protein variant across BO iterations. We monitor the diversity of the sampled batch concerning the past across BO iterations through $(1)$ the average Hamming distance between the executed variant and the proposed variant from the previous iteration (pairwise distance) and $(2)$ the average Hamming distance of the executed variant from the nearest initial training point.

In Appendix~\ref{app:experiment-results}, we provide additional performance metrics such as the fraction of global optima discovered, the fraction of discovered solutions above a fitness threshold, cumulative maximum, and mean pairwise Hamming distances. Moreover, we compare with discrete local search methods~\citep{botorch} in Appendix~\ref{app:comp-w-dls} and report their respective runtimes in Appendix~\ref{app:runtimes}.
\looseness=-1
\begin{figure*}[t]
    \centering  
    \begin{subfigure}{0.33\textwidth}
    \centering
    {\includegraphics[width=1\linewidth,trim=10 10 0 10,clip]{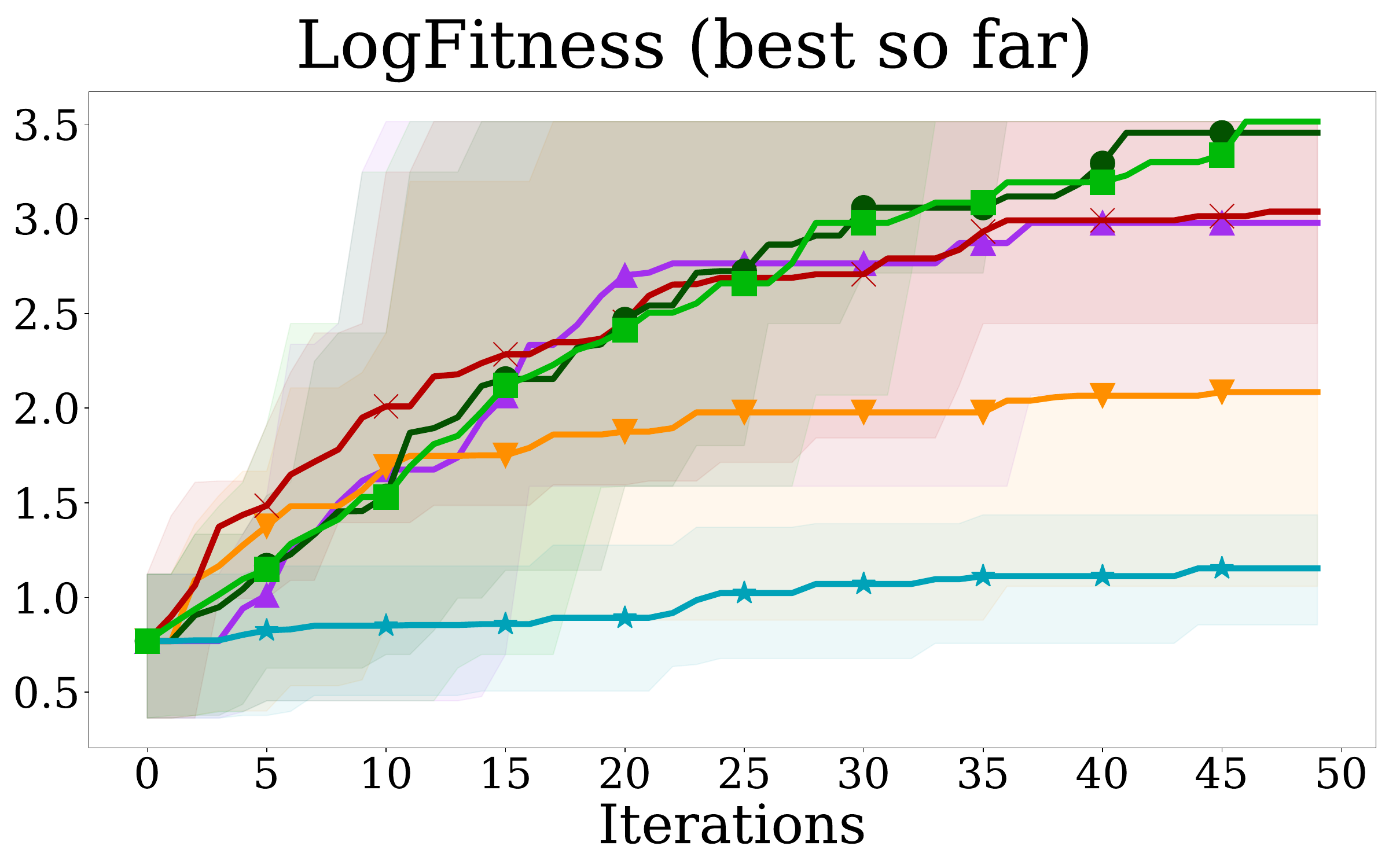}}
    \caption{\emph{Halogenase}, $n=3$, $18$ reps.}
    \label{fig:avg-perf-plot-d}
    \end{subfigure}%
    \begin{subfigure}{0.33\textwidth}
    \centering
    {\includegraphics[width=1\linewidth,trim=10 10 0 10,clip]{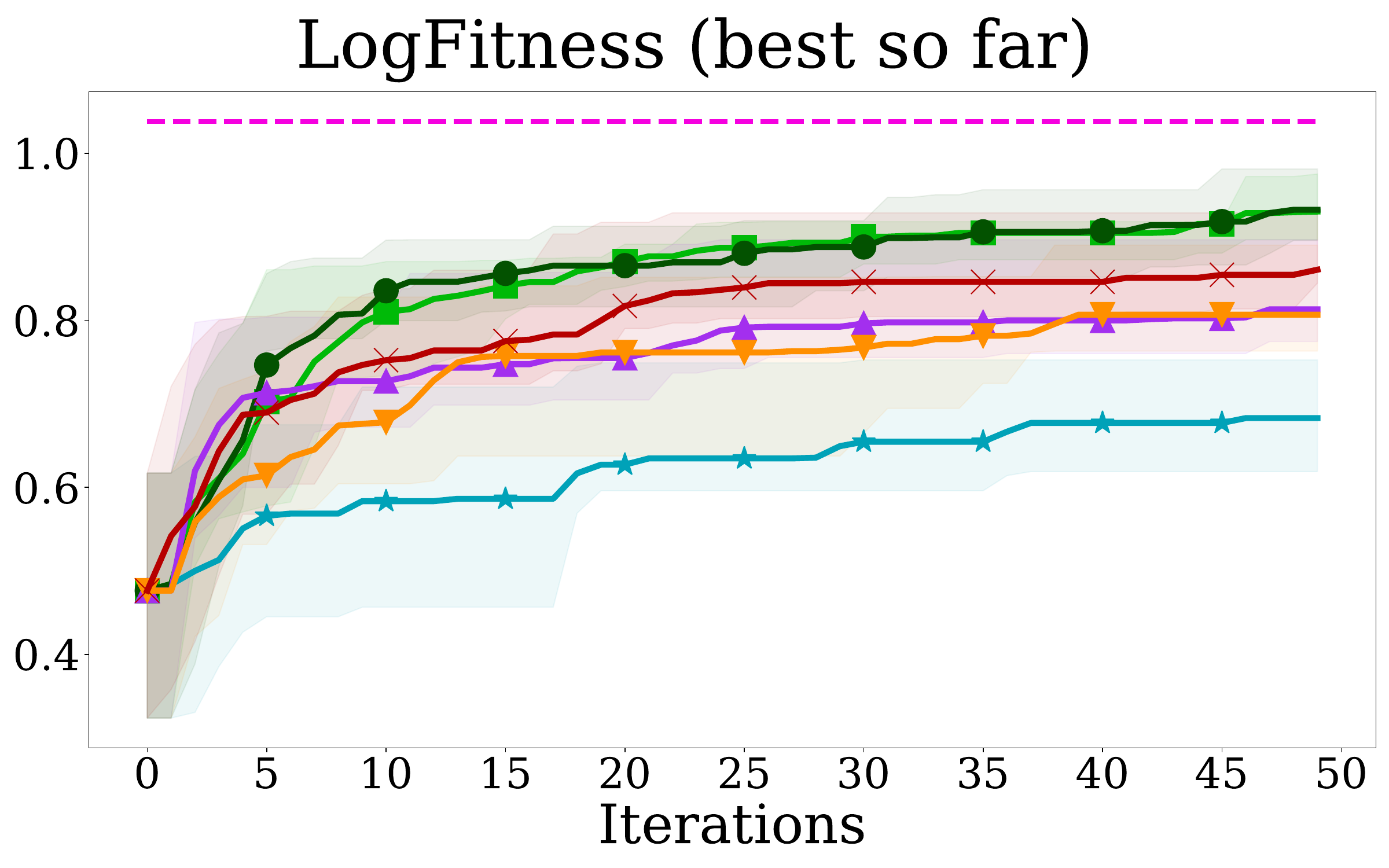}}
    \caption{ \emph{GB1(4)}, $n=4$, $18$ reps.}
    \label{fig:avg-perf-plot-a}
    \end{subfigure}%
    \begin{subfigure}{0.33\textwidth}
    \centering
    {\includegraphics[width=1\linewidth,trim=10 10 0 10,clip]{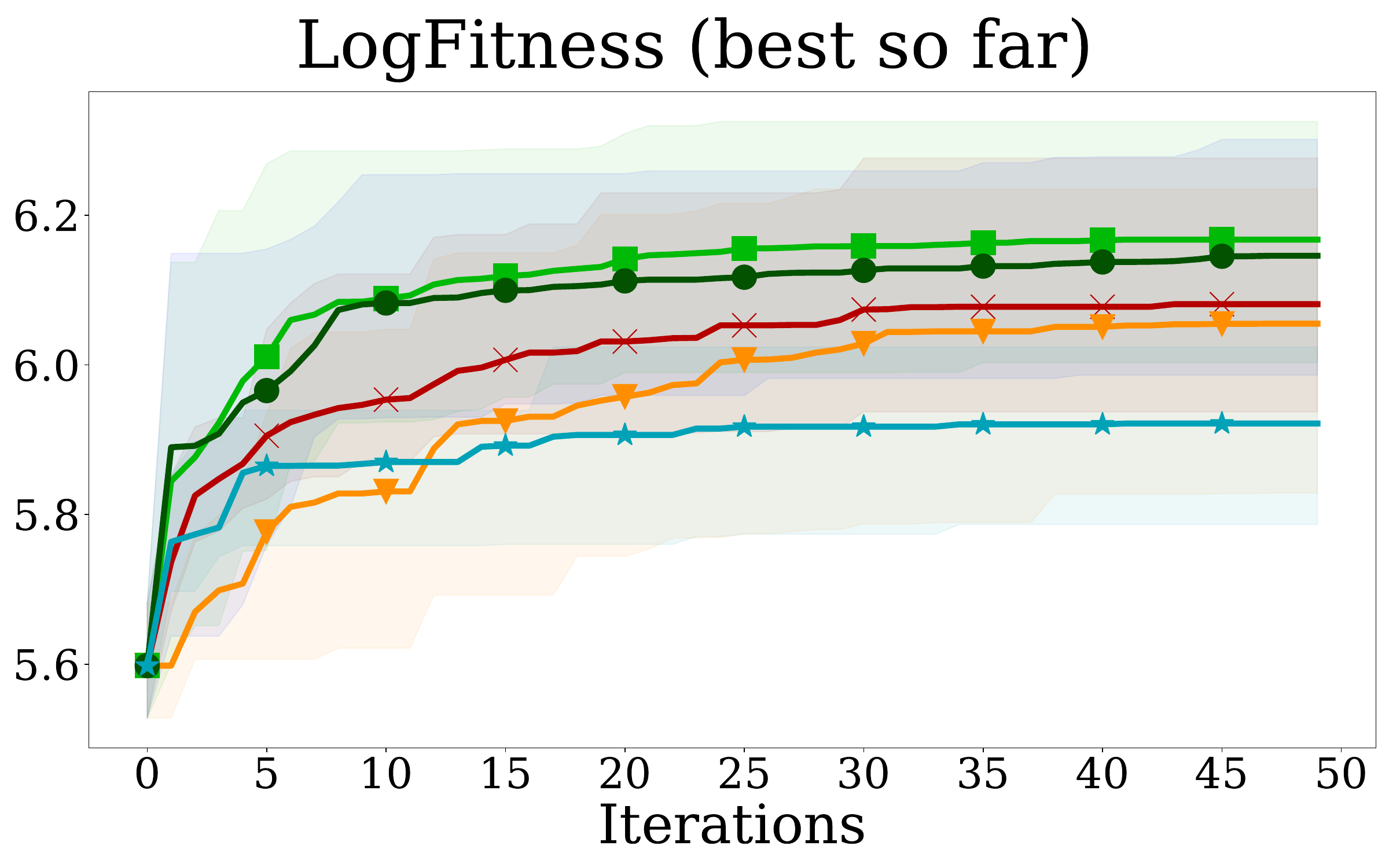}}
    \caption{\emph{GFP}, $n=6$, $10$ reps.}
    \label{fig:avg-perf-plot-h}
    \end{subfigure}%
    \\
    \begin{subfigure}{0.33\textwidth}
    \centering
    {\includegraphics[width=1\linewidth,trim=10 10 0 10,clip]{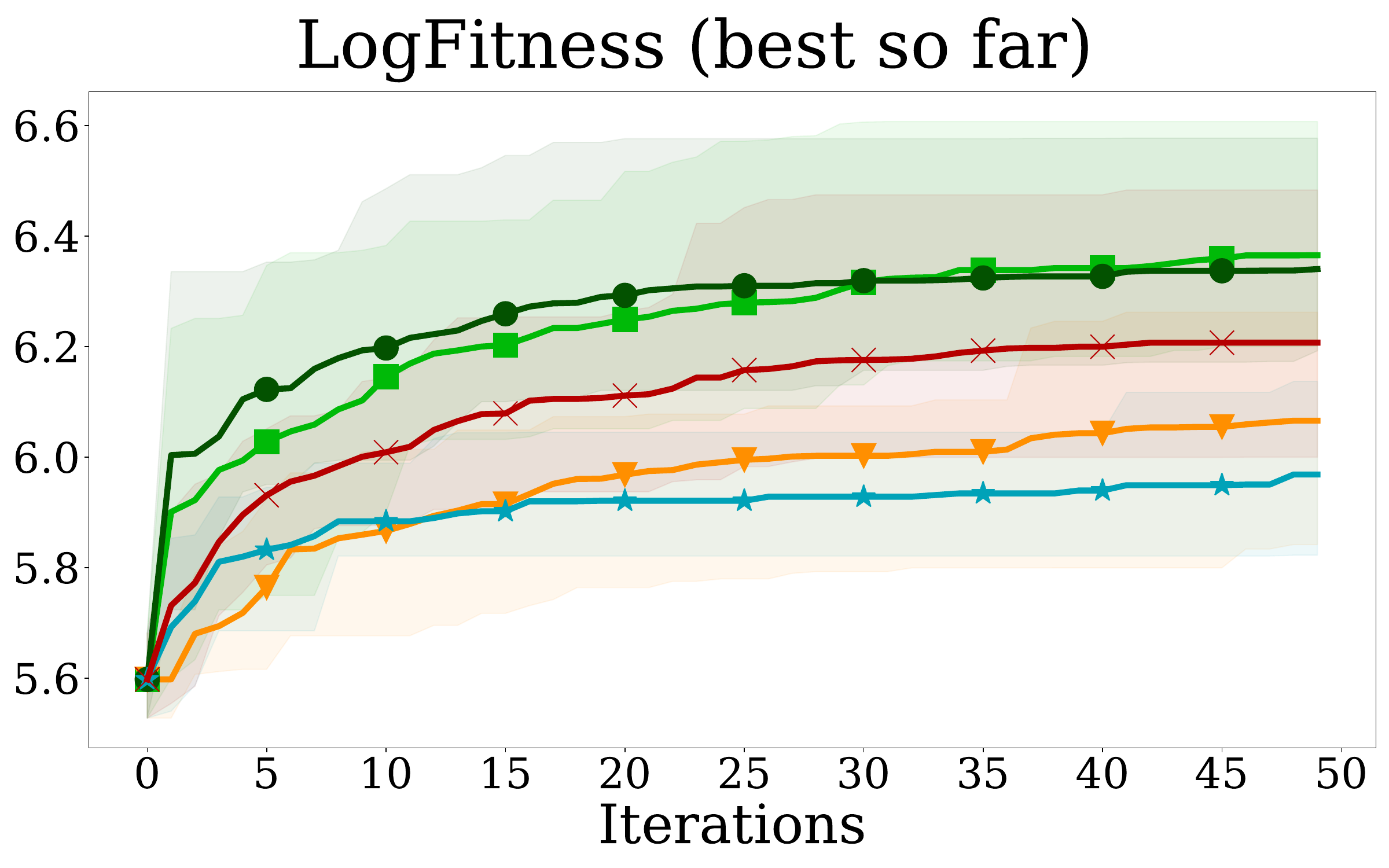}}
    \caption{\emph{GFP}, $n=8$, $10$ reps.}
    \label{fig:avg-perf-plot-e}
    \end{subfigure}%
    \begin{subfigure}{0.33\textwidth}
    \centering
    {\includegraphics[width=1\linewidth,trim=10 10 0 10,clip]{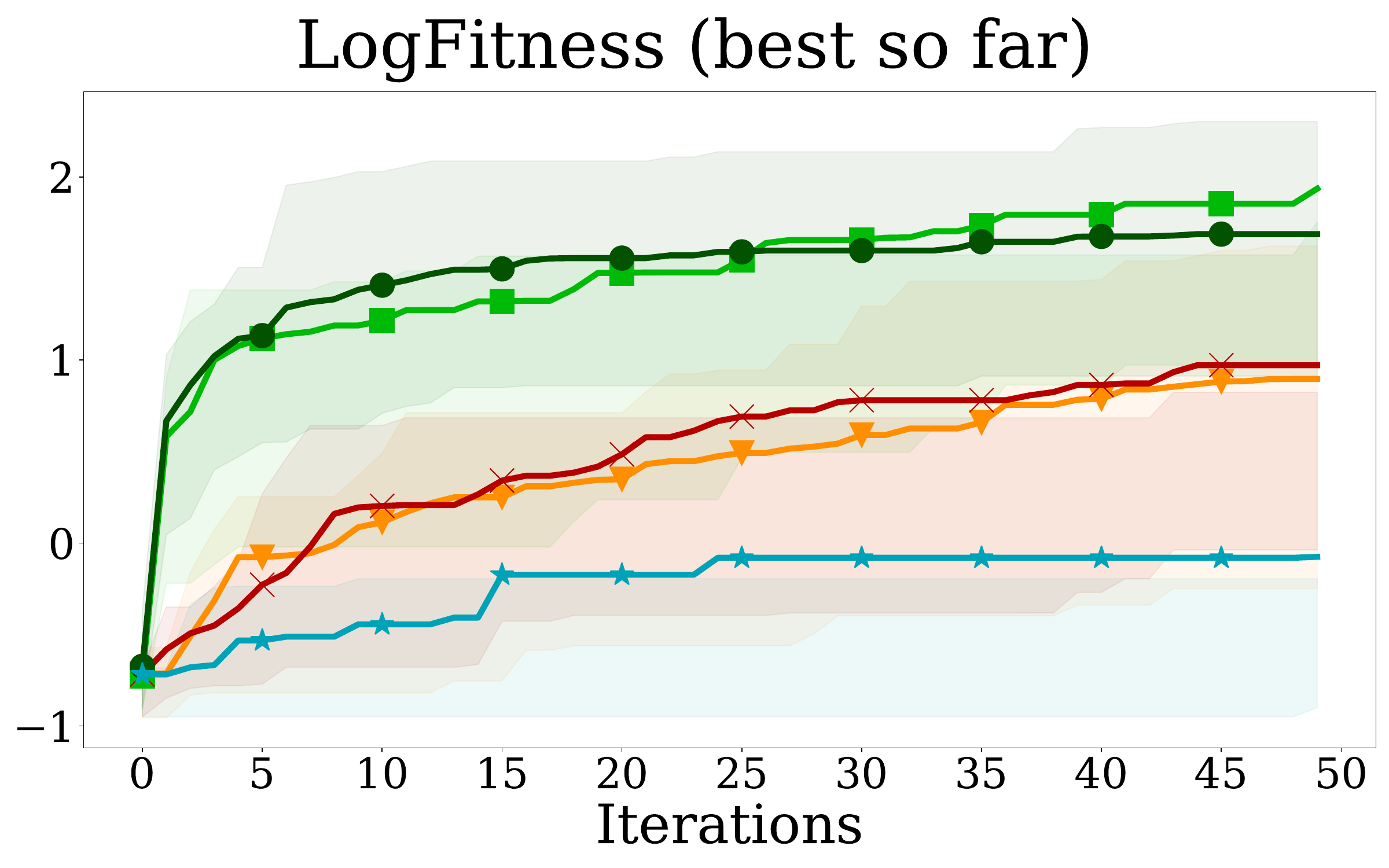}}
    \caption{\emph{GB1(55)}, $n=10$, $10$ reps.}
    \label{fig:avg-perf-plot-c}
    \end{subfigure}%
    \begin{subfigure}{0.33\textwidth}
    \centering
    {\includegraphics[width=1\linewidth,trim=10 10 0 10,clip]{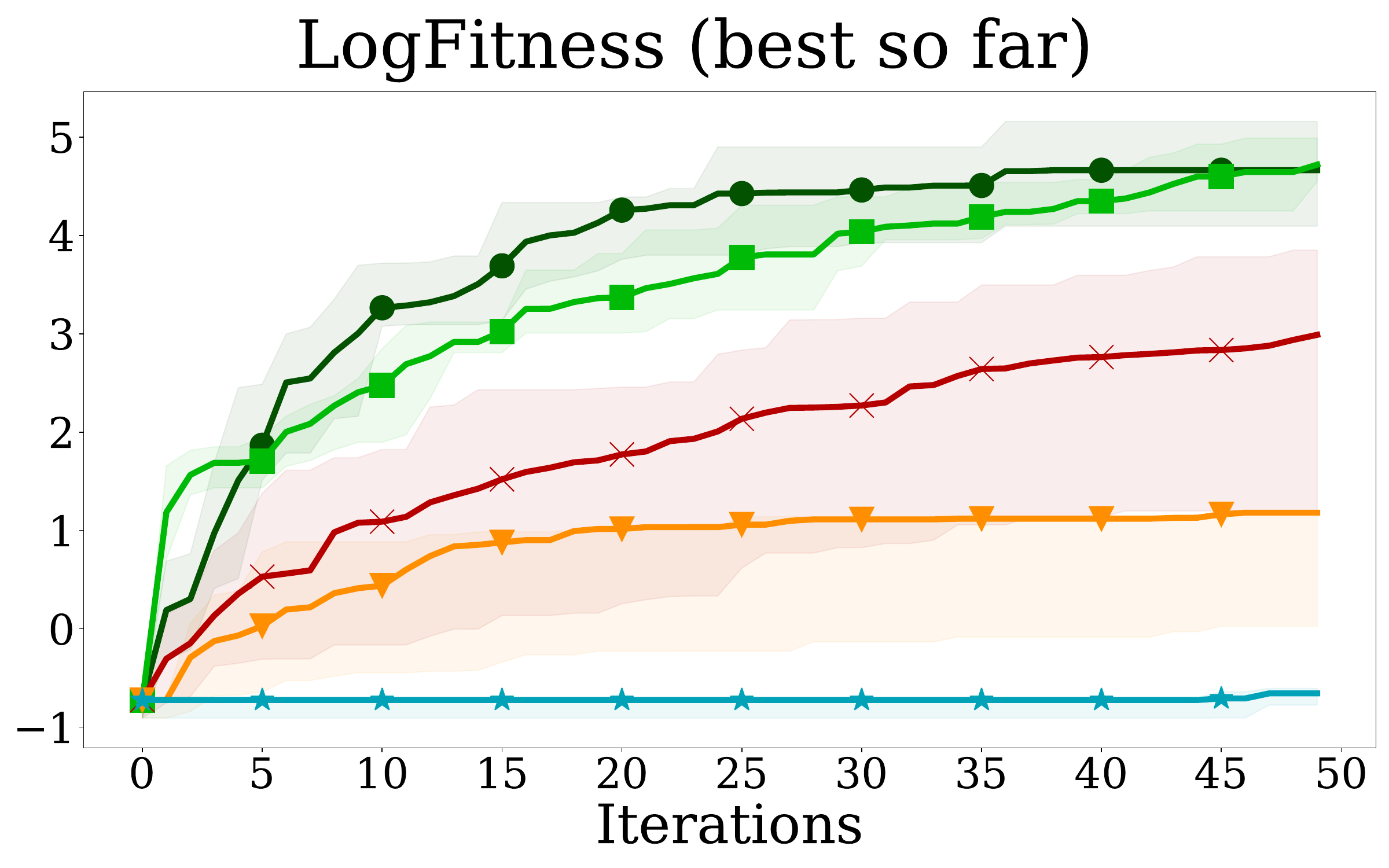}}
    \caption{\emph{GB1(55)}, $n=55$, $10$ reps.}
    \label{fig:avg-perf-plot-b}
    \end{subfigure}%
    \\
    \begin{subfigure}{1\textwidth}
    \centering
    {\includegraphics[width=1\linewidth,trim=40 20 5 15,clip]{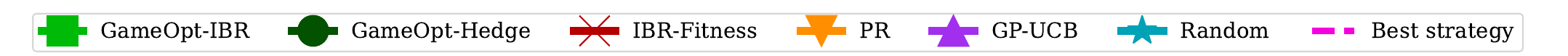}}
    \label{fig:subfig1}
    \end{subfigure}%
\setlength{\belowcaptionskip}{-15pt}
 \setlength{\abovecaptionskip}{-10pt}
 \caption{Convergence speed of methods in terms of log fitness value of the best-so-far protein across BO iterations, under batch size $B =5$.  Results are averaged over replications initiated with different training sets: $100$ protein variants for \emph{Halogenase} and \emph{GB1(4)}, and $1000$ protein variants for other domains.
 Error bars are interquartile ranges averaged over replications. 
 In all experiments \textsc{GameOpt} with \textsc{Ibr} and \textsc{Hedge} subroutines discover better protein sequences at a much faster rate.
 }
\label{fig:avg-perf-plot}
\end{figure*}

\vspace{-0.2cm}
\subsection{Results} 
\vspace{-0.2cm}
\label{sec:results}
\textsc{GameOpt}, with \textsc{Ibr} and \textsc{Hedge} equilibrium computation subroutines, consistently outperform baselines across all experiments, discovering higher fitness protein sequences faster (see Figure~\ref{fig:avg-perf-plot}).

\textbf{Results for \emph{Halogenase}. }\looseness=-1 
While initially surpassed by \textsc{IBR-Fitness}, \textsc{PR}, and \textsc{GP-UCB}, \textsc{GameOpt} variants converge faster to higher log fitness proteins than baselines. Notably, \textsc{IBR-Fitness} performs best-responses on the true log fitness function, whereas \textsc{GameOpt-Ibr} simulates best-response dynamics directly on the UCB model, allowing to compute equilibria at each iteration. Additionally, although \textsc{GP-UCB} performs comparably, it incurs higher computational demands. Furthermore, \textsc{PR} and \textsc{Random} perform poorly in the \emph{Halogenase} setting.
\begin{figure*}[t]
    \centering  
    \begin{subfigure}{0.33\textwidth}
    \centering
    {\includegraphics[width=1\linewidth,trim=0 10 0 0,clip]{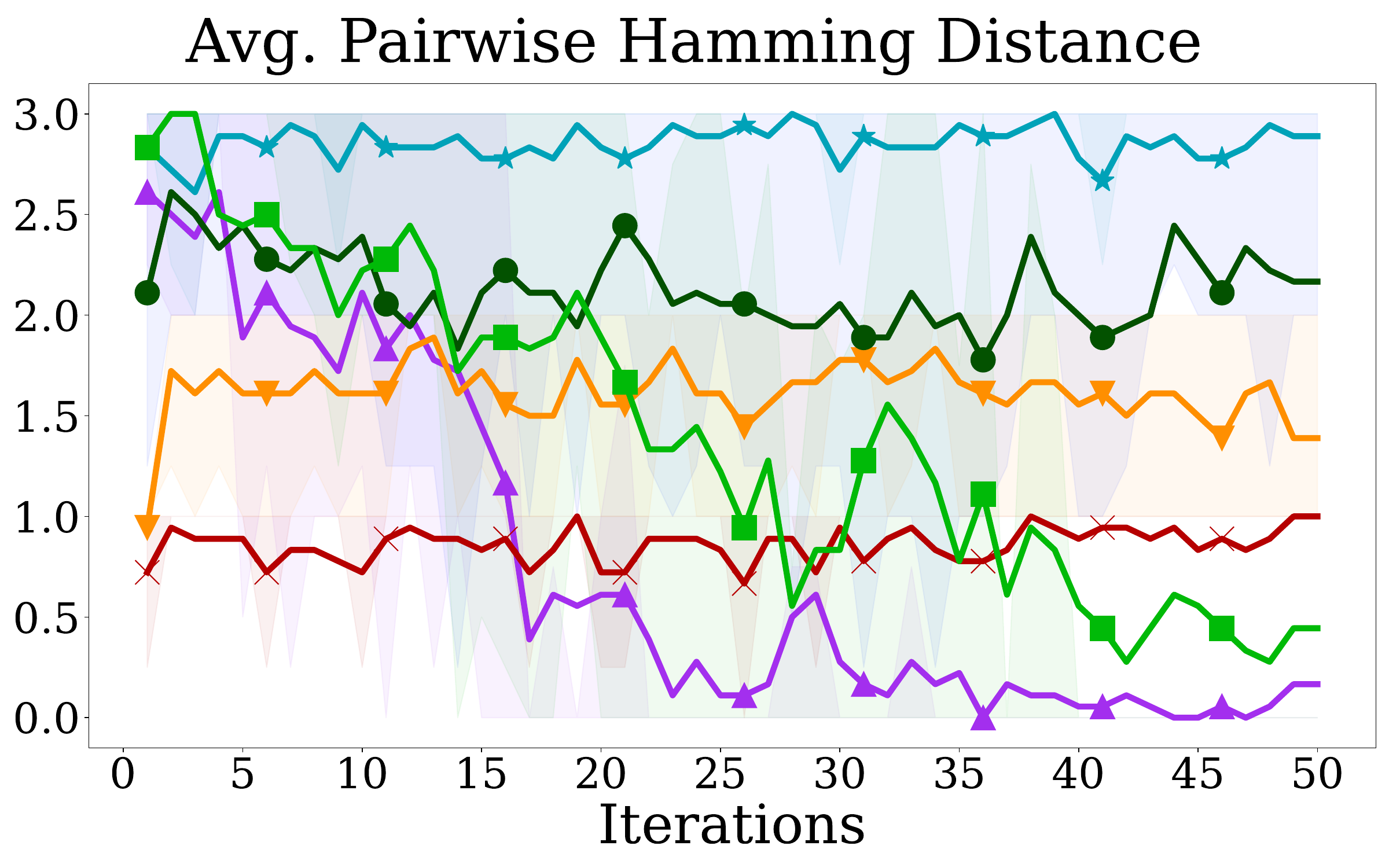}}
    \caption{\emph{Halogenase}, $n=3$, $18$ reps.}
    \label{fig:avg-perf-plot-hamming-c}
    \end{subfigure}%
    \begin{subfigure}{0.33\textwidth}\centering
    {\includegraphics[width=1\linewidth,trim=0 10 0 0,clip]{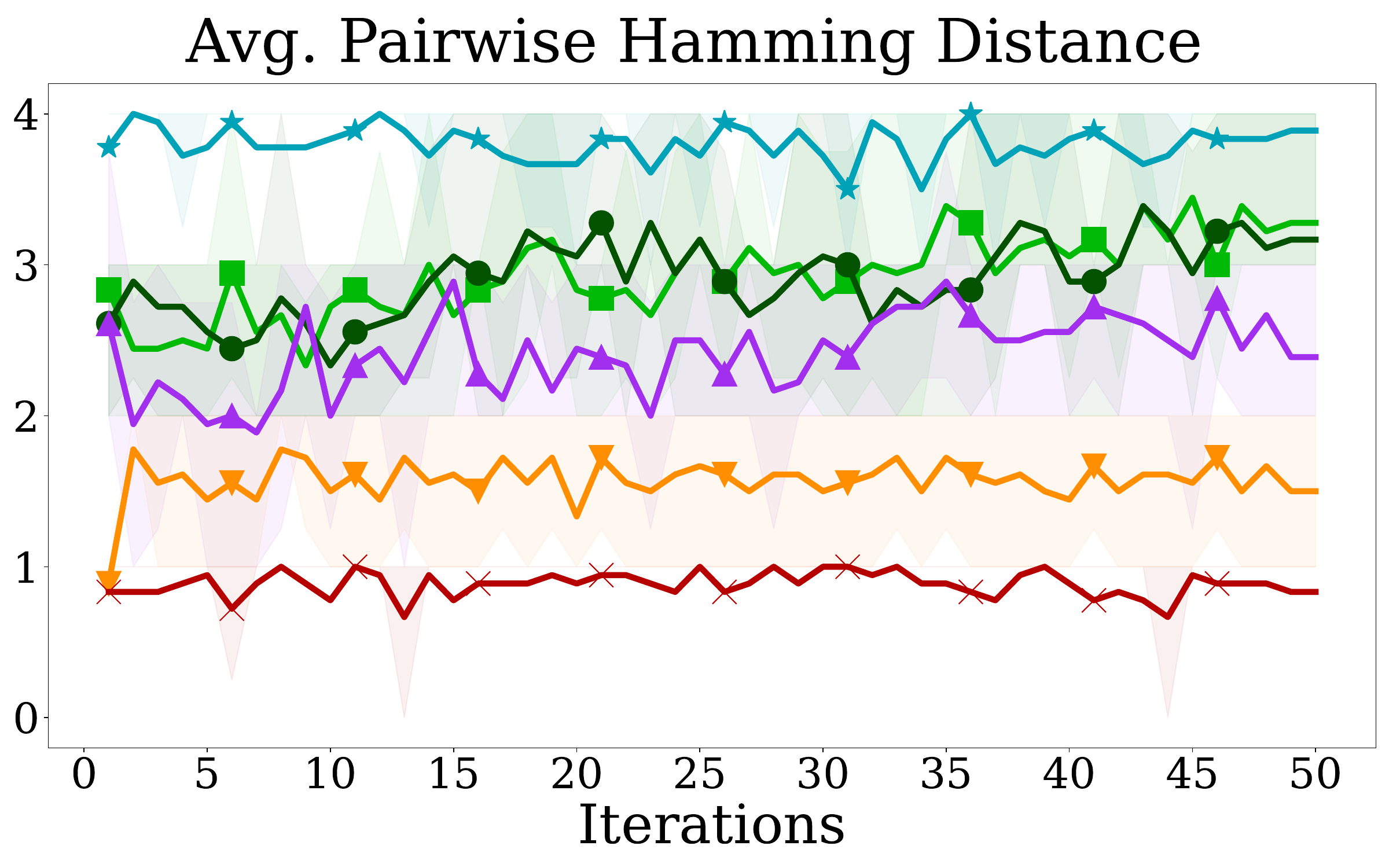}}
    \caption{\emph{GB1(4)}, $n=4$, $18$ reps.}
    \label{fig:avg-perf-plot-hamming-e}
    \end{subfigure}%
    \begin{subfigure}{0.33\textwidth}
    \centering
    {\includegraphics[width=1\linewidth,trim=0 10 0 0,clip]{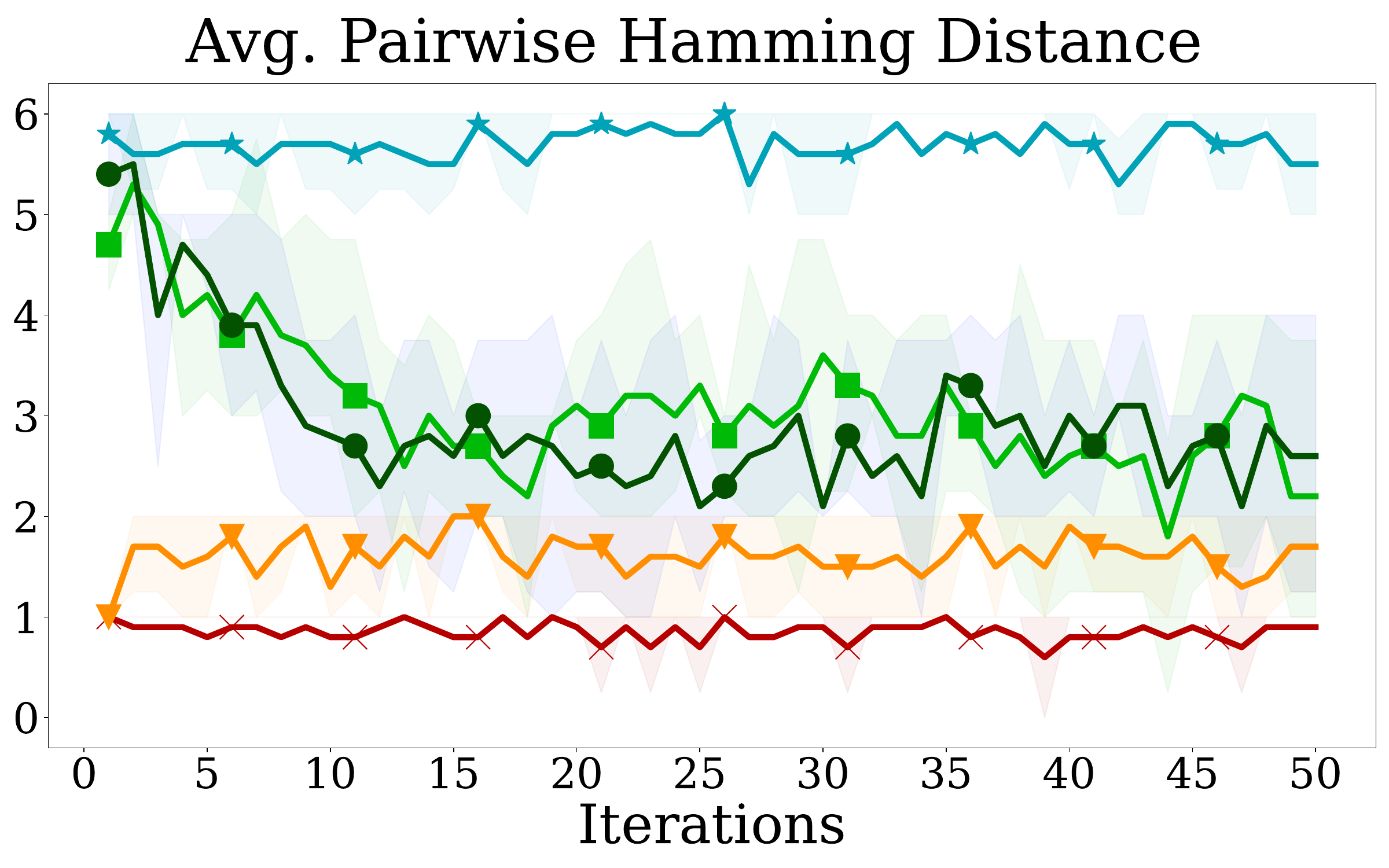}}
    \caption{\emph{GFP}, $n=6$, $10$ reps.}
    \label{fig:avg-perf-plot-hamming-h}
    \end{subfigure}%
    \\      
    \begin{subfigure}{0.33\textwidth}
    \centering
    {\includegraphics[width=1\linewidth,trim=0 10 0 0,clip]{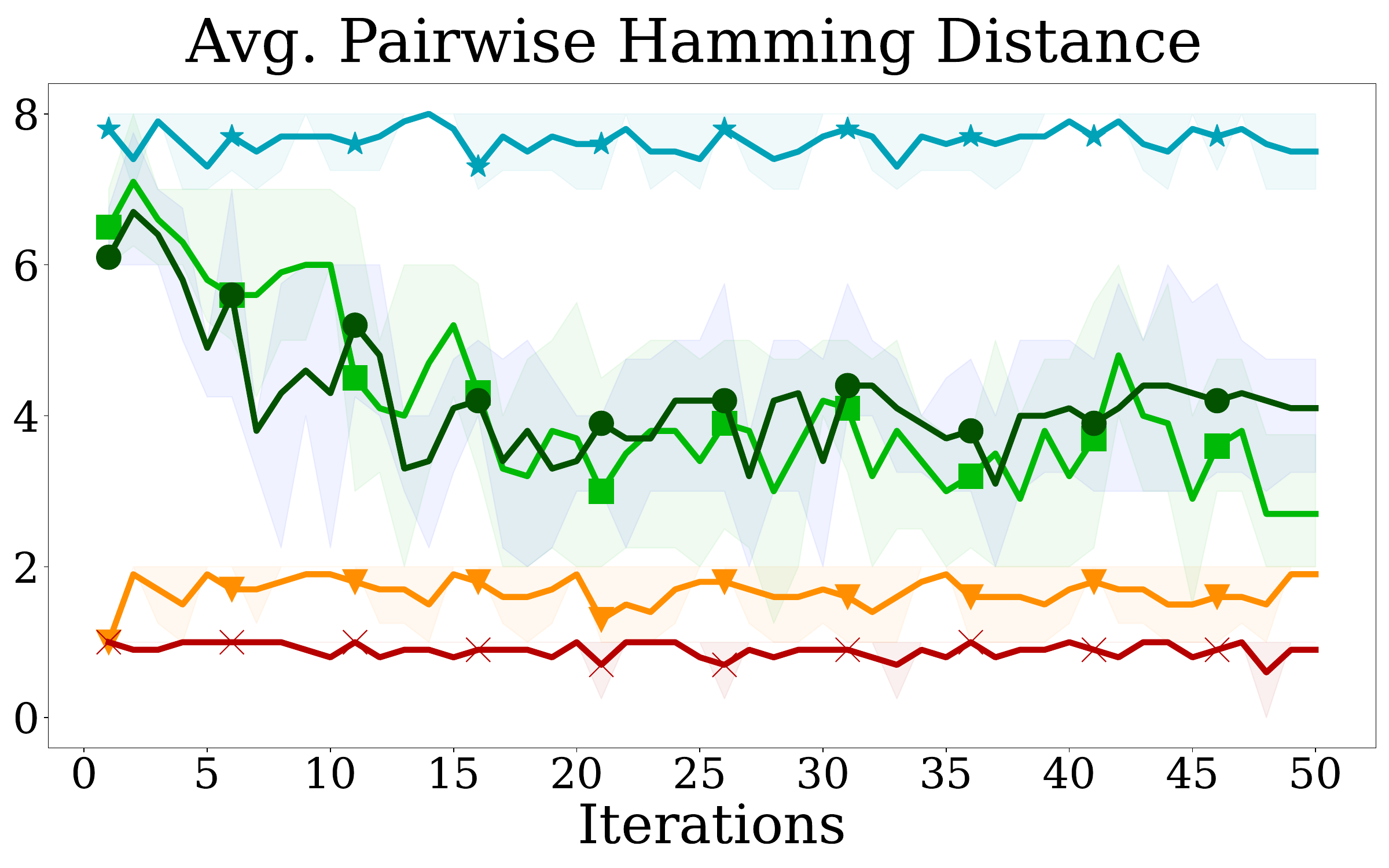}}
    \caption{\emph{GFP}, $n=8$, $10$ reps.}
    \label{fig:avg-perf-plot-hamming-d}
    \end{subfigure}%
    \begin{subfigure}{0.33\textwidth}
    \centering
    {\includegraphics[width=1\linewidth,trim=0 10 0 0,clip]{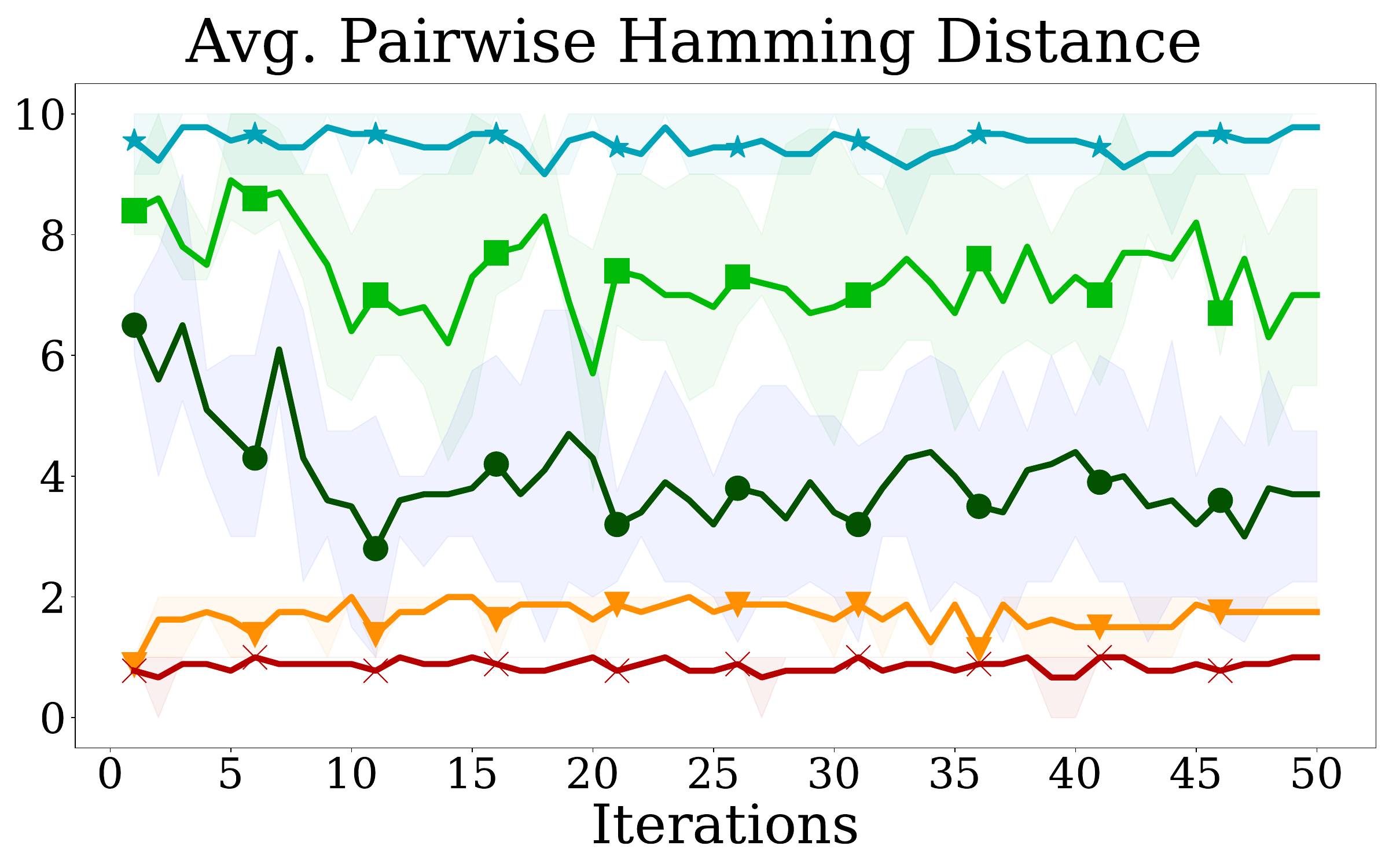}}
    \caption{\emph{GB1(55)}, $n=10$, $10$ reps.}
    \label{fig:avg-perf-plot-hamming-g}
    \end{subfigure}%
    \begin{subfigure}{0.33\textwidth}
    \centering
    {\includegraphics[width=1\linewidth,trim=0 10 0 0,clip]{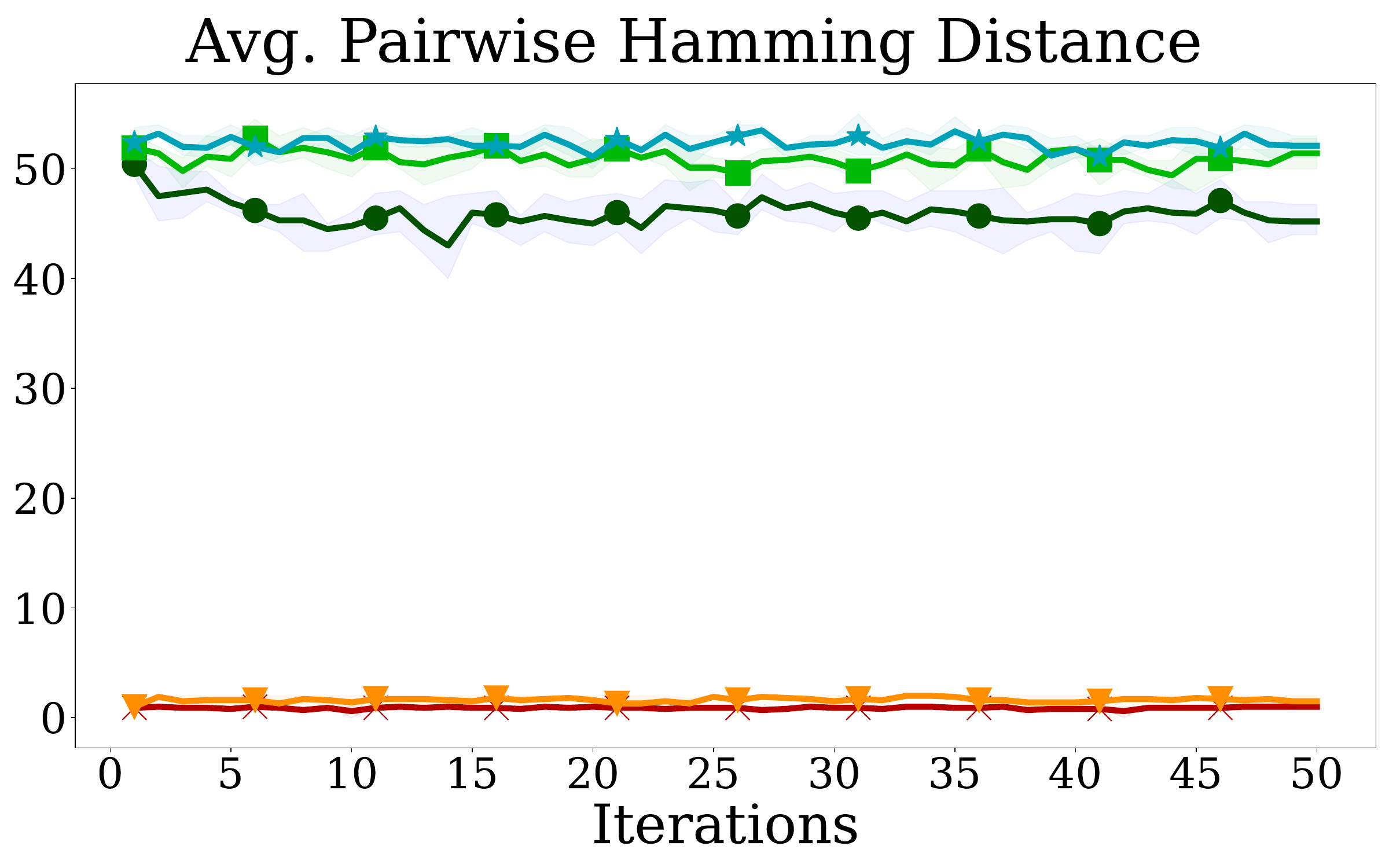}}
    \caption{\emph{GB1(55)}, $n=55$, $10$ reps.}
    \label{fig:avg-perf-plot-hamming-f}
    \end{subfigure}%
    \\
    \begin{subfigure}{1\textwidth}
    \centering
    {\includegraphics[width=1\linewidth,trim=40 20 5 15,clip]{figures/experiment_results/legend_5.pdf}}
    \label{fig:subfig1}
    \end{subfigure}%
\setlength{\belowcaptionskip}{-10pt}
\setlength{\abovecaptionskip}{-15pt}
 \caption{Sampled batch diversity relative to past, measured via mean Hamming distance between executed and proposed variants from the previous iteration (pairwise distance), under batch size $B=5$. 
 Results are averaged over replications, and error bars show interquartile ranges.
 In all experiments \textsc{GameOpt} consistently samples a rather diverse batch of evaluation points w.r.t. past proposed variants. This enhanced exploration of the search space contributes to its strong performance compared to the baseline methods. \looseness=-1}
\label{fig:avg-perf-plot-hamming-2}
\vspace{-0.2cm}
\end{figure*}

\textbf{Results for \emph{GB1(4)}. } \looseness=-1
Similarly, \textsc{GameOpt} approaches steadily surpass \textsc{PR} and \textsc{GP-UCB} while exploring superior protein sequences efficiently. Initially trailing \textsc{IBR-Fitness}, \textsc{GameOpt} approaches prove more adept at exploring and sampling diverse points (see Figure~\ref{fig:avg-perf-plot-hamming} in Appendix~\ref{app:experiment-results}).

The baseline \textsc{PR}  is not very competitive and also comes with higher computational demands. 
As highlighted in Section \ref{sec:related-work-comb-bo}, \textsc{PR} relies on the expected UCB as the acquisition function, requiring expectation computation across players set and amino acid choices. This makes its performance contingent on accurately estimating expected UCB through combinatorially many sequence samples.
In contrast, \textsc{GameOpt} efficiently finds stable outcomes by breaking down the combinatorial search space into individual decision sets, resulting in a more manageable process. \looseness=-1

Finally, of particular observation is the subpar performance of \textsc{GP-UCB} and \textsc{Random}.
A detailed analysis of \textsc{GP-UCB}'s performance is presented in Appendix~\ref{app:experiment-results-gp-ucb}, where it is observed that the efficacy of \textsc{GP-UCB} heavily relies on the quality of the initial GP surrogate. In contrast, \textsc{GameOpt} demonstrates robustness in overcoming the limitations of a model initialized with a limited amount of data, thereby enhancing its sample efficiency. Further discussion on the performance of methods can be found in Appendix~\ref{app:experiment-results}. \looseness=-1

\textbf{Results for \emph{GFP}. } \looseness -1  The complexity of the problem positively correlates with the performance gap between \textsc{GameOpt} and baselines. In the protein search space with $6$ and $8$ amino acid decisions, \textsc{GameOpt} with either subroutine excels in identifying high-log fitness protein sequences even from the start. Figure~\ref{fig:avg-perf-plot-hamming-2} 
further demonstrates \textsc{GameOpt}'s consistent exploration of diverse batches and Figure~\ref{fig:avg-perf-plot-hamming} in Appendix~\ref{app:sample-batch-diversity} shows its high rate of exploration. 

\textbf{Results for \emph{GB1(55)}. } \looseness -1 In both versions of the most complex problem domain, \textsc{GameOpt} demonstrates superior performance. As the decision flexibility (\textit{i.e.}, the number of players) increases from $10$ to $55$, the performance gap against baselines widens. Furthermore, \textsc{GameOpt} achieves incomparable batch diversity concerning past, both with respect to the initial training and previously executed protein sequences, as shown in Figures~\ref{fig:avg-perf-plot-hamming-2} and ~\ref{fig:avg-perf-plot-hamming} in Appendix~\ref{app:sample-batch-diversity}. 
\vspace{-0.2cm}
\subsection{Further discussion and limitations} 
\label{sec:further-limitations}
\looseness -1 \vspace{-0.2cm}
In our experiments, we compare to \textsc{IBR-Fitness}, which simulates currently employed strategies in the iterative protein optimization literature. This is by no means the only methodology applied in this field, and a comprehensive comparison is beyond the scope of this work.
\looseness -1 
\vspace{-0.2cm}
\paragraph{Batch size.} \looseness -1
Similar to the previous point, in the experiments presented in Section~\ref{sec:results}, we use a batch size of $B=5$. While we acknowledge this is a restrictive setup given the technological needs of common screenings, nevertheless, our results should be transferable and applicable irrespective of the batch size which differs to each laboratory setting.
To support this, we include batch size ablations, $B=\{1,3,5,10,50\}$, in Appendix~\ref{app:batch-size-ablation} (Figure~\ref{fig:batch-size-plot}), demonstrating that \textsc{GameOpt} variations consistently outperform baseline methods in discovering better protein sequences across various batch sizes.
\looseness -1
\vspace{-0.2cm}
\paragraph{Learning rate.} \looseness -1
In Appendix~\ref{app:lr-ablation-hedge}, we detail learning rate ablations for \textsc{GameOpt-Hedge}.
\looseness -1
\vspace{-0.2cm}
\paragraph{Efficient vs better optimization of the acquisition function.} \looseness -1
The primary benefit of \textsc{GameOpt} is its ability to efficiently optimize the acquisition function by adopting a game-theoretical perspective. Unlike GP-UCB, our focus is on \textit{tractable} optimization in large combinatorial spaces rather than improving optimization. This efficiency is vital for navigating such spaces, where \textsc{GameOpt} exploits game equilibria to enhance exploration.
Appendix~\ref{app:ucb-values-of-sampled-batches} (see Figure~\ref{fig:ucb-values-of-sampled-batch}) provides an analysis of UCB values over BO iterations. While \textsc{GameOpt} is initially outperformed by GP-UCB in collecting higher UCB-valued batches, its exploration strategy ultimately leads to more efficient optimization.
\looseness -1
\vspace{-0.2cm}
\paragraph{Different acquisition functions.} \looseness -1
We design the \textsc{GameOpt} framework using UCB as the acquisition function (also the game reward function), with its sample complexity derived accordingly.  
To assess broader applicability, we also evaluate \textsc{GameOpt} with alternative acquisition functions, such as expected improvement \citep{jones1998efficient}. 
As shown in Appendix~\ref{app:ei-acquisition}, Figure~\ref{fig:ei-af-plot}, our empirical comparison with GP-based methods reveals that \textsc{GameOpt} consistently outperforms these baselines across all performance metrics.
\looseness -1
\vspace{-0.2cm}
\paragraph{Comparison with latent space optimizers.} \looseness -1
In line with the primary focus of this study, we compare \textsc{GameOpt} to baselines that employ acquisition function optimizers operating \textit{directly} on large \textit{combinatorial} search spaces \citep{dreczkowski2024framework}, addressing the challenge of exhaustive exploration.
As discussed in Section~\ref{sec:related-work-comb-bo}, another line of work follows BO over continuous spaces by learning a latent representation via deep generative models. 
While a comprehensive comparison with these methods is beyond our scope, we provide additional insights in Appendix~\ref{app:comp-w-latent-space-optimizers}, where we empirically compare \textsc{GameOpt} against Naïve LSBO-(L-BFGS-B) \citep{gomez2018automatic}, using a second-order gradient-based optimizer, and LADDER \citep{deshwal2021combining}. The results highlight \textsc{GameOpt}'s effectiveness— not only in finding higher-fitness sequences tractably but also in bypassing the limitations of latent space optimizers, particularly their dependency on a decoder.
\looseness -1
\vspace{-0.2cm}
\paragraph{Sequence-based kernels.} \looseness -1
To show the applicability of \textsc{GameOpt} under various kernel choices, we provide additional analysis using string kernels \citep{moss2020boss} in Appendix~\ref{app:comp-w-latent-space-optimizers}. 
Specifically, we consider a structure-coupled kernel designed by combining an RBF kernel with a sub-sequence string kernel \citep{moss2020boss}. In this setting, \textsc{GameOpt} demonstrates faster convergence to higher log fitness protein sequences, further emphasizing its adaptability across different kernel configurations.
\looseness -1
\vspace{-0.2cm}
\section{Conclusions} \looseness -1  \vspace{-0.2cm}
We introduced \textsc{GameOpt}, a novel \textit{tractable} game-theoretical approach to combinatorial BO that leverages game equilibria of a cooperative game between discrete inputs of a costly-to-evaluate black-box function to tractably optimize the acquisition function over combinatorial and unstructured spaces, and select informative points. Empirical analysis on challenging protein design problems showed that \textsc{GameOpt} surpassed baselines in terms of convergence speed, consistently identifying better protein variants more quickly, thereby being more resource-efficient. 
\textsc{GameOpt} is a versatile framework, allowing for exploration with different acquisition functions or mixed equilibrium concepts. As for future work, an adaptive grouping of players could be further investigated. 
\vspace{-0.2cm}
\section*{Acknowledgements}
MIB acknowledges the support of the Max Planck Graduate Center for Computer and Information Science. The preliminary experiments were made possible through the support of the Max Planck Institute for Software Systems compute cluster.

\bibliography{gameopt}
\bibliographystyle{iclr2025_conference}

\newpage
\appendix

\addcontentsline{toc}{section}{Appendix} %
\part{Appendix} %
\parttoc %
\newpage

We gather here the technical proofs, the details on \textsc{GameOpt}'s application to protein design, and additional experiment results complementing the main paper.

\section{Proof of Theorem~\ref{thm:thm_sample_complexity}}
\label{app:proof-thm4-2}
The proof relies on the main confidence lemma~\citep[Lemma 5.1]{gpopt-bandit-krause} which states that, when the confidence width is set as $\beta_t=  2n\log\left(\sup_{i\in [n]}|\mathcal{X}_i|\frac{t^2\pi^2}{6\delta}\right) \geq 2\log\left(|\prod_{i=1}^n \mathcal{X}_i|\frac{t^2\pi^2}{6\delta}\right)$, then with probability at least $(1-\delta)$,
\begin{equation}\label{eq:confidence_lemma}
   \mu_t(x) - \beta_t \sigma_t(x) \leq f(x) \leq  \mu_t(x) +  \beta_t \sigma_t(x),\quad \forall x\in \mathcal{X} ,\quad \forall t\geq 1.
\end{equation}
In other words, $\text{UCB}(\mathcal{GP}^t,\cdot)$ and $\text{LCB}(\mathcal{GP}^t,\cdot)$ are upper and lower bound functions with high probability. For simplicity, we will use the notation $\text{UCB}_t(\cdot)$ and $\text{LCB}_t(\cdot)$ for $\text{UCB}(\mathcal{GP}^t,\cdot)$ and $\text{LCB}(\mathcal{GP}^t,\cdot)$, respectively.

Next, we show that after $T$ iterations the strategy reported by \textsc{GameOpt}: $x_{T^\star}$, with $T^\star = \arg\min_{t\in [T]} \max_{i, x^i} \text{UCB}_t(x^i, x^{-i}_t) - \text{LCB}_t(x_t)$, is a $\epsilon_T$-approximate Nash equilibrium of $f$ with $\epsilon_T \leq \mathcal{O}(T^{-0.5}\sqrt{\gamma_T})$. The theorem statement then follows by a simple inversion of the aforementioned bound.

By definition, $x_{T^\star}$ is a $\epsilon_T$-approximate Nash equilibrium of $f$ when $\epsilon_T = \max_{i, x^i} f(x^i, x_{T^\star}^{-i}) - f(x_{T^\star})$, i.e. $\epsilon_T$ upper bounds all possible single-player deviations. We can bound the worst-case single-player deviation with probability  $(1-\delta)$ by the following chain of inequalities:

\begin{align}
   \epsilon_T = \max_{i, x^i} f(x^i, x_{T^\star}^{-i}) - f(x_{T^\star}) & \leq \max_{i, x^i} \text{UCB}_{T^\star}(x^i, x_{T^\star}^{-i}) - \text{LCB}_{T^\star}(x_{T^\star}) \\
    & \leq \frac{1}{T}\sum_{t=1}^T \max_{i, x^i} \text{UCB}_t(x^i, x^{-i}_t) - \text{LCB}_t(x_t) \\
    & = \frac{1}{T}\sum_{t=1}^T \max_{i, x^i} \text{UCB}_t(x^i, x^{-i}_t) - \text{UCB}_t(x_t) + \frac{2}{T} \sum_{t=1}^T \beta_t \sigma_t(x_t)\\
    & \leq \frac{2}{T} \sum_{t=1}^T \beta_t \sigma_t(x_t) \leq \mathcal{O}(T^{-0.5} \sqrt{\beta_T \gamma_T}).
\end{align}
The first inequality follows from the confidence lemma~\eqref{eq:confidence_lemma}. The second one, by the fact that $ \max_{i, x^i} \text{UCB}_{T^\star}(x^i, x_{T^\star}^{-i}) - \text{LCB}_{T^\star}(x_{T^\star}) \leq \max_{i, x^i} \text{UCB}_t(x^i, x^{-i}_t) - \text{LCB}_t(x_t), \forall t$, by definition of $T^{\star}$. The last inequality holds because, at each iteration $t$, $x_t$ is an equilibrium of the $\text{UCB}_t$ function. Finally, the last inequality is from~\citep[Lemma 5.4]{gpopt-bandit-krause}. \hfill \qed

\section{\textsc{GameOpt} for Protein Design} 
\label{app:evolution-opt}

The core concept of the \textsc{GameOpt} framework is inspired by the principles of natural evolution.
In protein design, achieving equilibrium of a cooperative game over protein sites mirrors the iterative mutation and selection process in evolution. 
Where it converges to beneficial mutant sequences, can be thought of as equilibrium of the game. 
Given that protein search spaces align well with the domain \textsc{GameOpt} works on, we introduce a specialized version of \textsc{GameOpt}, tailored for protein design applications. 

\begin{algorithm}[t!]
\caption{\textsc{GameOpt-Ibr} for Protein Design}
\label{alg:evolution-opt-ibr}
\begin{algorithmic}[1]
\STATE {\bfseries Input:} GP prior $\mathcal{GP}^0(\mu_{0}, k(\cdot,\cdot))$, initial data $\mathcal{D}_{0}=\{(x_{i}, y_{i}=f(x_{i})+\epsilon)\}$, protein sites $\mathcal{N}$, batch size $B \in \mathbb{N}$,  $M \in \mathbb{N} > B$, parameter $\beta$.
\FOR{iteration $t = 1,2, \dots, T$}
\STATE Construct game with reward function $\text{UCB}(\mathcal{GP}^{t-1}, \beta, \cdot): \prod_{i=1}^n \mathcal{X}^{(i)} \to \mathbb{R}$
\FOR{$m=1,2,\dots,M$}
\STATE $\bold{x}^{br}_{0} \gets$ random starting protein sequence,  $ \bold{x}^{br}_{0} \in \mathcal{X}$
\FOR[\hfill{\color{amber(sae/ece)} */ BR game}]{round $k=1,2,\dots,K$}
\STATE $\mathcal{X}^{br}_{k} \gets \Big\{(x^{i,br}, x_{k-1}^{-i, br}), \text{ such that }x^{i,br} = \arg\max_{x\in \mathcal{X}^{(i)}} \text{UCB}(x, x_{k-1}^{-i, br}) \Big\}_{i=1}^n$ 
\STATE Play $\bold{x}^{br}_{k} \gets \arg \max_{\bold{x}_{k} \in \mathcal{X}^{br}_{k}} [\text{UCB}(x_{k})]$
\ENDFOR
\STATE Collect equilibrium protein sequence $x_{t,m} \gets \bold{x}^{br}_{K}$
\ENDFOR
\STATE Select batch of top $B$ equilibrium protein sequences  $\{x_{t,i}\}_{i=1}^B$ according to $\text{UCB}(\mathcal{GP}^{t-1},\beta, \cdot)$. \hfill {\color{amber(sae/ece)} */ Filtering}
\STATE Obtain fitness evaluations $y_{t,i} = f(x_{t,i}) + \epsilon_{t,i}, \quad \forall i =1,\ldots, B$  
\STATE Update $\mathcal{D}_{t} \gets \mathcal{D}_{t-1} \cup  \{(x_{t,i}, y_{t,i})\}_{i=1}^B$
\STATE Posterior update of model $\mathcal{GP}^t$  with $\mathcal{D}_{t}$ 
\ENDFOR
\end{algorithmic}
\end{algorithm}

\begin{algorithm}[t!]
\caption{\textsc{GameOpt-Hedge} for Protein Design}
\label{alg:evolution-opt-hedge}
\begin{algorithmic}[1]
\STATE {\bfseries Input:} GP prior $\mathcal{GP}^0(\mu_{0}, k(\cdot,\cdot))$, initial data $\mathcal{D}_{0}=\{(x_{i}, y_{i}=f(x_{i})+\epsilon)\}$, protein sites $\mathcal{N}$, batch size $B \in \mathbb{N}$,  $M \in \mathbb{N} > B$, parameters $\beta, \eta$.
\FOR{iteration $t = 1,2, \dots, T$}
\STATE Construct game with reward function $\text{UCB}(\mathcal{GP}^{t-1}, \beta, \cdot): \prod_{i=1}^n \mathcal{X}^{(i)} \to \mathbb{R}$
\FOR{$m=1,2,\dots,M$}
\STATE Initialize weights $\boldsymbol{w}_{1} \gets \frac{1}{W}[1,\dots, 1] \in \mathbb{R}^{\mid \mathcal{N} \mid \times W}$
\FOR[\hfill{\color{amber(sae/ece)} */ Simultaneous Hedge}]{round $k=1,2,\dots,K$}
\STATE Sample $x^{i}_{k} \sim \boldsymbol{w}^{i}_{1}, \forall i \in \mathcal{N} $
\STATE Set joint strategy $\bold{x}_{k} \gets \{ x^{i}_{k} \}_{i \in \mathcal{N}} $
\FOR[\hfill{\color{amber(sae/ece)} */ Players' payoff}]{player $i \in \mathcal{N}$} 
\STATE $\ell_{x^{-i}_{k}} \gets  [v(x^{j,-i}_{k})]_{\forall j \in \mathcal{X}^{(i)}}$, where \newline $x^{j,-i}_{k} = {j} \cup \{ x^{i^{\prime}}_{k} \}_{i^{\prime} \in \mathcal{N} \setminus \{i\}}, \forall j \in \mathcal{X}^{(i)}$ 
\STATE Set $\boldsymbol{w}^{i}_{k+1} \propto \boldsymbol{w}^{i}_{k} \text{ exp}(\eta \ell_{x^{-i}_{k}})$ 
\ENDFOR
\ENDFOR
\STATE Collect equilibrium protein sequence $x_{t,m} \gets \bold{x}_{K}$
\ENDFOR
\STATE Select batch of top $B$ equilibrium protein sequences  $\{x_{t,i}\}_{i=1}^B$ according to $\text{UCB}(\mathcal{GP}^{t-1},\beta, \cdot)$. \hfill {\color{amber(sae/ece)} */ Filtering}
\STATE Obtain fitness evaluations $y_{t,i} = f(x_{t,i}) + \epsilon_{t,i}, \quad \forall i =1,\ldots, B$  
\STATE Update $\mathcal{D}_{t} \gets \mathcal{D}_{t-1} \cup  \{(x_{t,i}, y_{t,i})\}_{i=1}^B$
\STATE Posterior update of model $\mathcal{GP}^t$  with $\mathcal{D}_{t}$ 
\ENDFOR
\end{algorithmic}
\end{algorithm}

\section{Baselines} 
\label{app:baselines}

In Section~\ref{sec:analysis}, we empirically evaluate \textsc{GameOpt} against existing baselines which we detail next. These include
\textsc{IBR-Fitness}, inspired by directed evolution (Algorithm~\ref{alg:ibr-fitness}), \textsc{Random} (Algorithm~\ref{alg:random}), which samples evaluation points randomly, and \textsc{PR}, an optimizer of expected UCB \citep{bo-prob-reparam}. We further compared \textsc{GameOpt} with discrete local search methods in \Cref{app:comp-w-dls}.

\begin{algorithm}[ht!]
\caption{\textsc{Iterative Best Response-Fitness (Ibr-Fitness)}}
\label{alg:ibr-fitness}
\begin{algorithmic}[1]
\STATE {\bfseries Input:} Domain $\mathcal{X}$, fitness function $f: \mathcal{X} \to \mathbb{R}$, players $\mathcal{N}$, initial data $\mathcal{D}_{0}=\{(x_{i}, y_{i}=f(x_{i})+\epsilon)\}$, batch size $B \in \mathbb{N}$.
\STATE $\bold{x}^{br}_{0} \gets$ random joint strategy,  $ \bold{x}^{br}_{0} \in \mathcal{X}$
\FOR{iteration $t = 1,2, \dots, T$}
\STATE Randomly selected $B$ players $\in \mathcal{N}$ generates BRs $\{x_{t,i}\}_{i=1}^B$ w.r.t. $\bold{x}^{br}_{t-1}$ based on $f(\cdot)$
\STATE Obtain evaluations $y_{t,i} = f(x_{t,i}) + \epsilon_{t,i}, \quad \forall i =1,\ldots, B$  
\STATE Update $\mathcal{D}_{t} \gets \mathcal{D}_{t-1} \cup  \{(x_{t,i}, y_{t,i})\}_{i=1}^B$
\STATE Play $\bold{x}^{br}_{t} \gets \arg \max_{x_{t,i} \in \{x_{t,i}\}_{i=1}^B} y_{t,i}$
\ENDFOR
\STATE \textbf{return} $\bold{x}^\star_{T} \gets \arg \max _{(x,y) \in \mathcal{D}_{T}} y$  \hfill {\color{amber(sae/ece)} */ Best-so-far}
\end{algorithmic}
\end{algorithm}

\begin{algorithm}[ht!]
\caption{\textsc{Random}}
\label{alg:random}
\begin{algorithmic}[1]
\STATE {\bfseries Input:} Domain $\mathcal{X}$, $f: \mathcal{X} \to \mathbb{R}$, initial data $\mathcal{D}_{0}=\{(x_{i}, y_{i}=f(x_{i})+\epsilon)\}$, batch size $B \in \mathbb{N}$.
\FOR{iteration $t = 1,2, \dots, T$}
\STATE Randomly generate batch of $B$ points $\{x_{t,i}\}_{i=1}^B, \forall x_{t,i} \in \mathcal{X}$ 
\STATE Obtain evaluations $y_{t,i} = f(x_{t,i}) + \epsilon_{t,i}, \quad \forall i =1,\ldots, B$  
\STATE Update $\mathcal{D}_{t} \gets \mathcal{D}_{t-1} \cup  \{(x_{t,i}, y_{t,i})\}_{i=1}^B$
\ENDFOR
\STATE \textbf{return} $\bold{x}^\star_{T} \gets \arg \max _{(x,y) \in \mathcal{D}_{T}} y$  \hfill {\color{amber(sae/ece)} */ Best-so-far}
\end{algorithmic}
\end{algorithm}

\section{Experiment Details} 
\label{app:experiment-details}

We set the (hyper)parameters for the experiments as in Table~\ref{tab:params-and-hyperparams}.

\begin{table}[h]
\caption{Hyperparameter values.}
\centering
\resizebox{1\linewidth}{!}{
\begin{tabular}{c|l|l}
\specialrule{1.5pt}{1pt}{1pt} 
\textbf{Symbol} & \textbf{Hyperparameter}     & \textbf{Value} \\ 
\specialrule{1.5pt}{1pt}{1pt} 
$T$  & The number of active learning (BO) iterations  & $50$  \\ 
$K$ & The number of game rounds & $40$ for \emph{GFP}, $200$ for \emph{Halogenase} and \emph{GB1(55)}, and $400$ for \emph{GB1(4)}; \\
& & $50$ for \textsc{GameOpt-Ibr} in \emph{GB1(55)} \& $n=10$ domain\\
& & $100$ for \textsc{GameOpt-Ibr} in \emph{GB1(55)} \& $n=55$ domain\\
$n = \mid \mathcal{N} \mid$        & The number of players                                     & $3$ for \emph{Halogenase}, $4$ for \emph{GB1(4)}, $6$ and $8$ for \emph{GFP}, $10$ and $55$ for \emph{GB1(55)}        \\ 
$\mid \mathcal{D}_{0} \mid$ & The number of samples in training set & $100$ for \emph{Halogenase} and \emph{GB1(4)}, and $1000$ for \emph{GFP} and \emph{GB1(55)} \\
$\eta$ & Learning rate & Optimized over the set \{$0.1, 0.5, 0.9, 1.0, 1.5, 2.0$\}\\
$\epsilon$ & Observation noise for each training example & $0.0004$  \\
$l$ & RBF kernel lengthscale & Optimized offline\\
$\beta$ & The UCB tuning parameter & $2$ \\
$B$ & Batch size per BO iteration & 5 \\
\specialrule{1.5pt}{1pt}{1pt} 
\end{tabular}
}
\label{tab:params-and-hyperparams}
\end{table}

\paragraph{\emph{GB1(4)}}
The dataset \citep{gb1-4-dataset} is fully combinatorial, \textit{i.e.}, encompassing fitness measurements of $20^{4}$ variants with $4$ sites. In this context, each protein site is treated as a player in the cooperative game of \textsc{GameOpt}, with $\mathcal{N}=\{1, \dots, 4\}$. Additionally, we also analyzed the effect of player grouping inspired by \textit{epistasis} phenomenon in protein design and provided the analysis in Appendix~\ref{app:experiment-results}. 

We train the GP surrogate by utilizing a small portion of the dataset, specifically $0.0625\%$, consisting of $100$ protein variants. 
Since existing literature does not provide common ground feature embeddings as representations for the \emph{GB1(4)} variants, we use chemical descriptors \citep{machine-assisted-de} to extract $60$ feature embeddings using a training set of size $1000$ protein variants with LASSO method. We apply $k$-fold cross-validation with $k=18$ different train/test dataset partitions. Following this, we evaluate the performance of our approach over $18$ replications. In each replication, we initialize the GP surrogate-based baseline methods with the same initial GP model as our approach. We also use the same initial protein sequence for comparison within that replicate but employ different initial points across replications. We set the starting joint strategy as the protein sequence having the highest log fitness value in the training set. The prior mean of the GP is fixed at $1.0162$. For the kernel hyperparameters, $60$ lengthscales are defined for each feature dimension and optimized offline at the beginning of a replication; outputscale is set to $0.02169$.

\paragraph{\emph{GB1(55)}} 
We experiment on the non-exhaustive dataset \emph{GB1(55)} that only includes 2-point mutations throughout the entire $55$ residues of the \emph{GB1} protein resulting in $535,917$ variants \citep{gb1-55-dataset} and consider two settings: $55$ and $10$ number of players. 

\paragraph{\emph{GB1(55)} with $\bold{55}$ Players}
In this context, we treat each protein site as a player in the \textsc{GameOpt}, thus, $\mathcal{N}=\{1, \dots, 55\}$.

As the dataset is not completely combinatorial, we do not have access to measured fitness values for all $20^{55}$ variants. To overcome this, we employ a Deep Neural Network-based (DNN) \textit{oracle} to predict fitness scores using feature embeddings associated with the protein sequences.
We again opt to feature embeddings as the representation for categorical input of GP surrogate. Unlike \emph{GB1(4)}, we utilize the \texttt{ESM-1v} protein language model from \texttt{esm} introduced by \cite{ESM-1v}, specifically designed for predicting protein variant effects and can be used to extract embeddings. With \texttt{ESM-1v}, we represent a sequence through a $1280$ dimensional feature embedding vector. 
We train the \textit{oracle} with supervised learning,
using the training set having $(477\,854 \times 1280, 477\,854)$ feature \& label pairs.
Obtaining the exhaustive version of the \emph{GB1(55)} dataset, we train the GP surrogate using \texttt{ESM-1v} feature embeddings of $1000$ randomly generated protein variants and corresponding \textit{oracle}-predicted fitness scores for $10$ replications.

\paragraph{\emph{GB1(55)} with $\bold{10}$ players}
To further analyze the performance of \textsc{GameOpt} compared to the other baselines, we consider the setting where among the $55$ sites, only $10$ most significant protein sites can be mutated. 

We employ the same protein language model for embeddings and \textit{oracle} to predict fitness scores. However, the choice of $10$ players among ${55\choose 10}$ possibilities is a strategic decision that affects the design performance. For this, we define the significance of a protein site considering the average variation in the fitness scores in the dataset. Concretely, we use Algorithm \ref{alg:compute-signif-sites} and select $\mathcal{N}=\{21,24,35,39,41,45,46,47,48,50\}$ sites as the players. We treat the rest of the protein sequence, \textit{i.e.}, sites that do not correspond to players as fixed.

\begin{algorithm}[ht!]
\caption{\textsc{ComputeMostSignificantSites}}
\label{alg:compute-signif-sites}
\begin{algorithmic}[1]
\STATE {\bfseries Input:} Dataset $\mathcal{D}=(x_{i},y_{i})_{i=1}^{N}$, players $K$, protein sequence length $L$, amino acids set $\mathcal{A}$.
\STATE Initialize $players \gets \emptyset$, ${site\_score}^{k}_{a} \gets \emptyset$  and ${site\_var}^{k} \gets 0, \forall k \in \{1,\dots,L\}, a \in \mathcal{A}$ 
\FOR{each pair $(x_{i},y_{i}) \in \mathcal{D}$}
\FOR{each site $k \in \{1,\dots,L\}$}
\STATE Set amino acid in site $k$ as $a \gets x_{i}^{k}$
\STATE Append ${site\_score}^{k}_{a} \gets {site\_score}^{k}_{a} \cup \{y_{i}\}$
\ENDFOR
\ENDFOR
\STATE ${site\_score}^{k} \gets \{ {site\_score}^{k}_{a} \}_{a \in \mathcal{A}}, \forall k \in \{1,\dots,L\}$
\STATE ${site\_var}^{k} \gets stdev({site\_score}^{k}), \forall k \in \{1,\dots,L\}$
\STATE \textbf{return} $K$ sites having highest ${site\_score}$ as \textit{players}
\end{algorithmic}
\end{algorithm}

\paragraph{\emph{Halogenase}} \emph{Halogenase} is a non-exhaustive dataset involving fitness measurements for $605$ unique variants. To obtain the complete protein fitness landscape, we again construct an oracle, \textit{i.e.}, an MLP having $R^2=0.96$ on a test set and experiment on a setting where each protein site is a player.

\paragraph{\emph{GFP} with $\bold{6}$ players} The \textit{Aequorea victoria} green-fluorescent protein dataset only includes fitness measurements of $35,584$  variants corresponding to mixed mutations of some positions on a $238$ length sequence.
For the fully combinatorial protein fitness landscape, we construct an oracle, \textit{i.e.}, an MLP with test $R^2=0.90$ and choose $6$ positions: $\mathcal{N}=\{10,18,22,37,67,78\}$ that have the largest number of mutations in the original dataset as the players of the \textsc{GameOpt}.

\paragraph{\emph{GFP} with $\bold{8}$ players} To set the players in this setting, we identified $8$ positions that have the largest number of mutations in the original dataset: $\mathcal{N}=\{10,18,22,37,67,78,196,112\}$.

\clearpage

\section{Additional Experimental Results} 
\label{app:experiment-results}

\subsection{Sample batch diversity}
\label{app:sample-batch-diversity}
We also evaluate our method against baselines in terms of performance metric: sampled batch diversity concerning the past. It is measured via the mean Hamming distance of executed points at each BO iteration to the $(1)$ closest initial training point and $(2)$ the proposed point at the previous iteration (pairwise distance).

\begin{figure*}[ht!]
    \centering  
    \begin{subfigure}{0.33\textwidth}
    \centering
    {\includegraphics[width=1\linewidth,trim=0 10 0 0,clip]{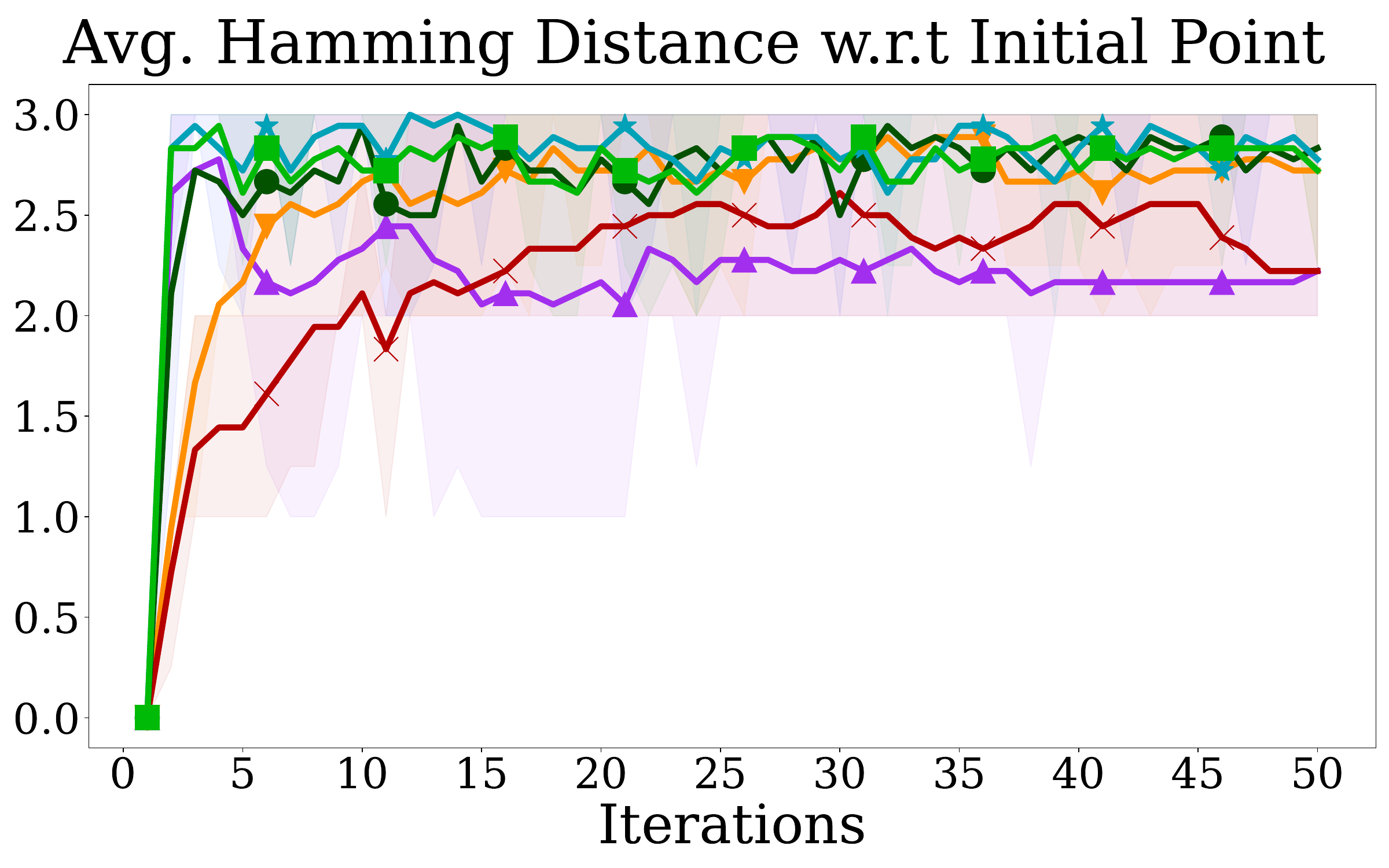}}
    \caption{\emph{Halogenase}, $n=3$, $18$ reps.}
    \label{fig:avg-perf-plot-hamming-d}
    \end{subfigure}%
    \begin{subfigure}{0.33\textwidth}
    \centering
    {\includegraphics[width=1\linewidth,trim=0 10 0 0,clip]{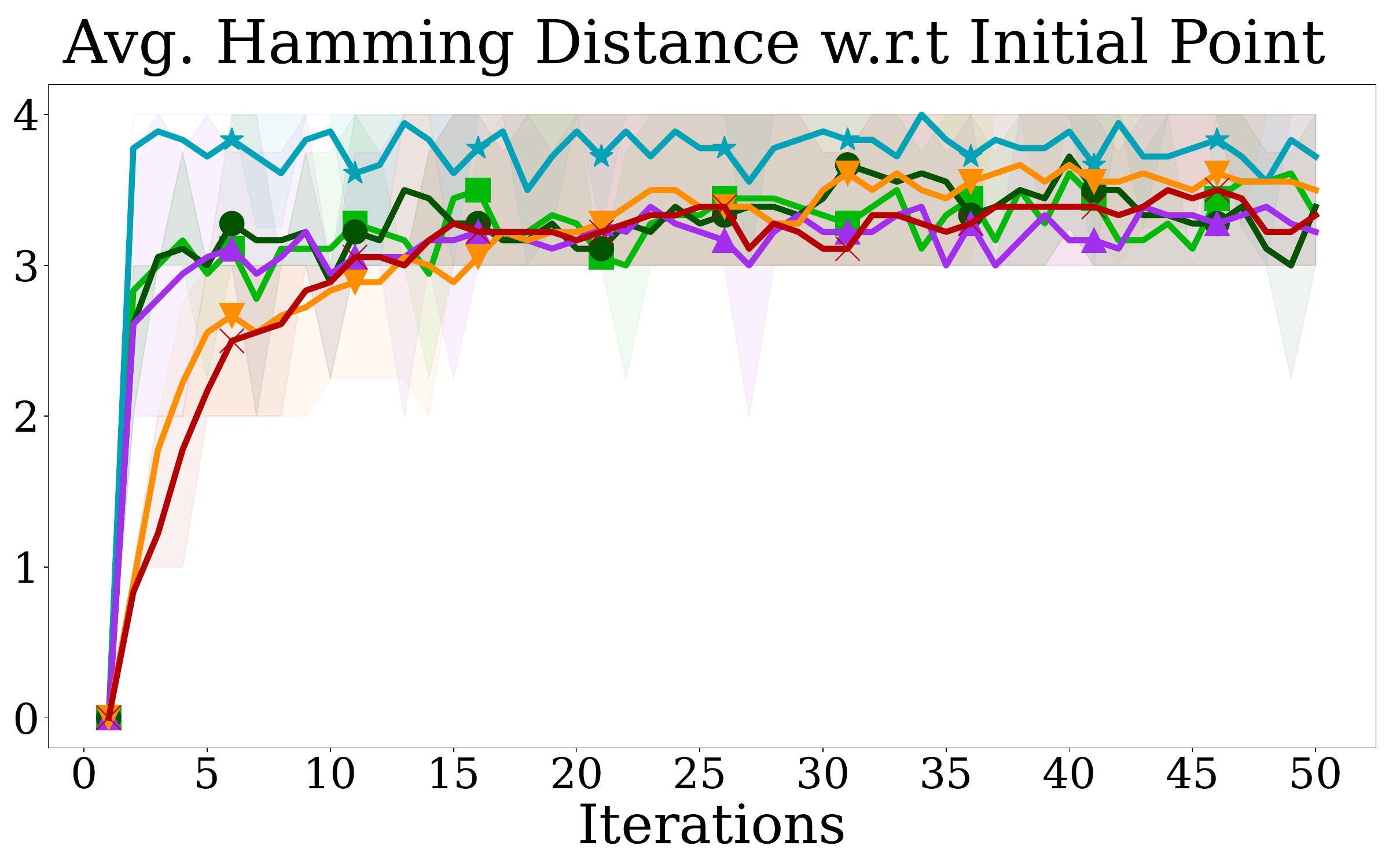}}
    \caption{\emph{GB1(4)}, $n=4$, $18$ reps.}
    \label{fig:avg-perf-plot-hamming-a}
    \end{subfigure}%
        \begin{subfigure}{0.33\textwidth}
    \centering
    {\includegraphics[width=1\linewidth,trim=0 10 0 0,clip]{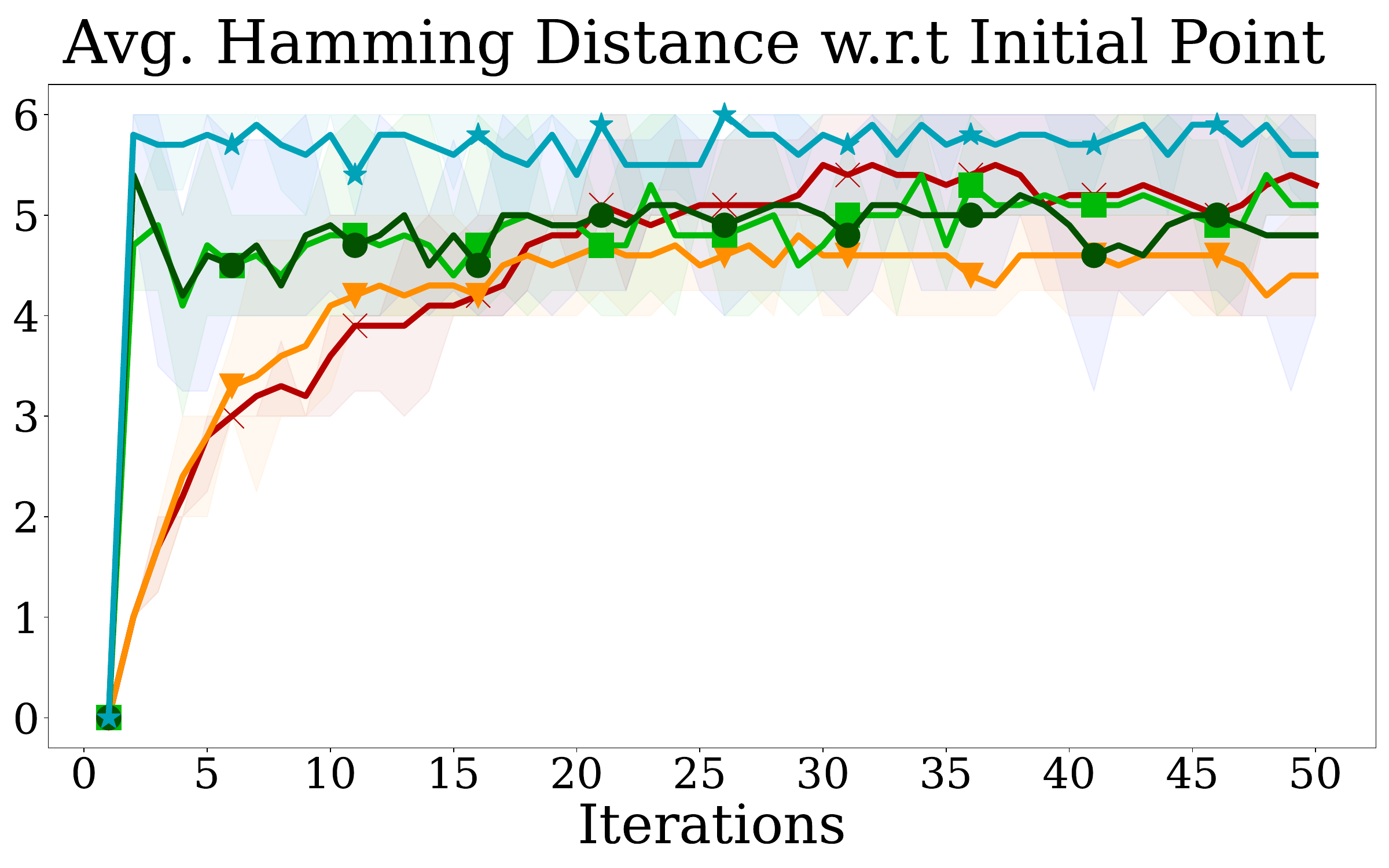}}
    \caption{\emph{GFP}, $n=6$, $10$ reps.}
    \label{fig:avg-perf-plot-hamming-h}
    \end{subfigure}%
    \\
    \begin{subfigure}{0.33\textwidth}
    \centering
    {\includegraphics[width=1\linewidth,trim=0 10 0 0,clip]{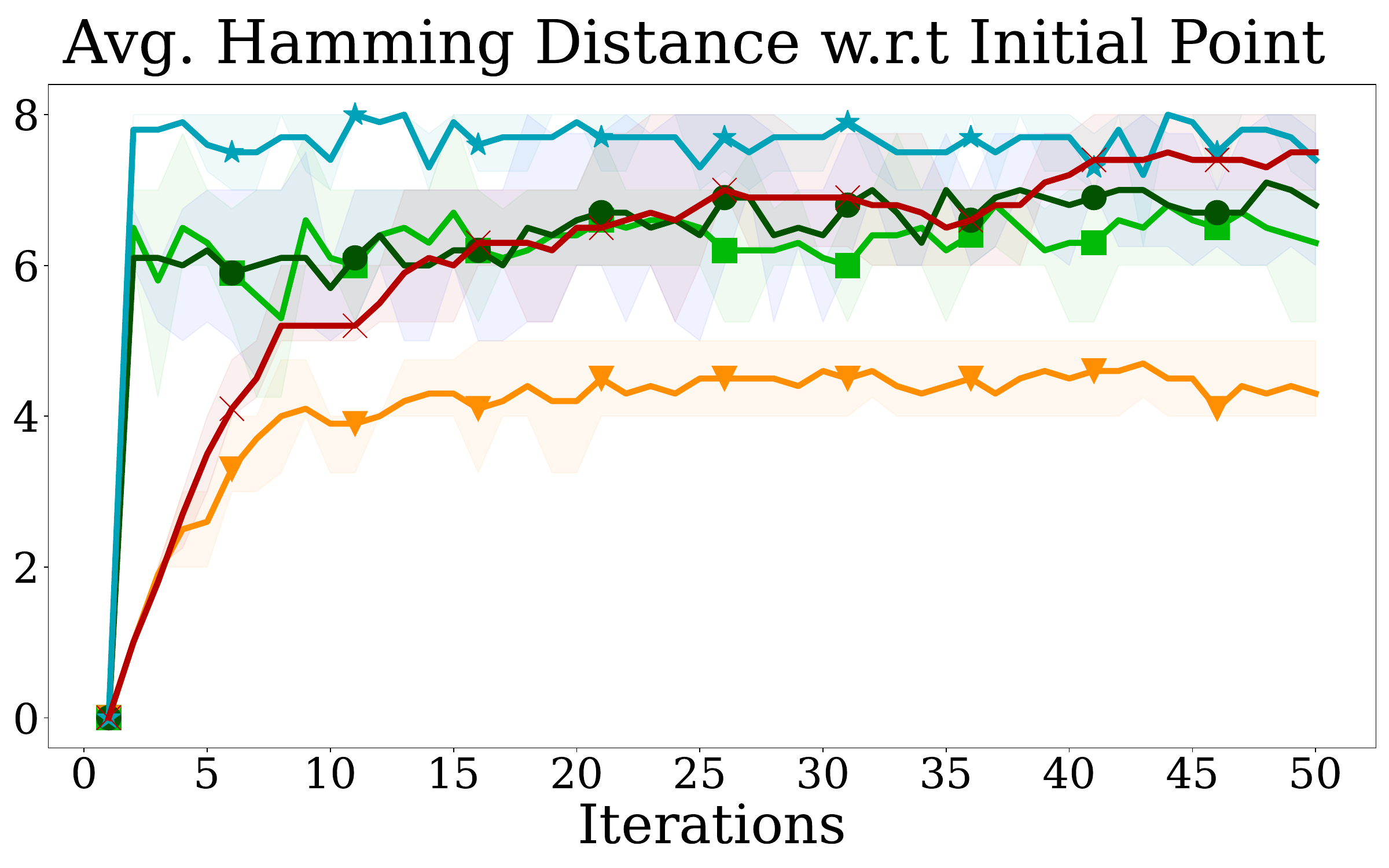}}
    \caption{\emph{GFP}, $n=8$, $10$ reps.}
    \label{fig:avg-perf-plot-hamming-e}
    \end{subfigure}%
    \begin{subfigure}{0.33\textwidth}
    \centering
    {\includegraphics[width=1\linewidth,trim=0 10 0 0,clip]{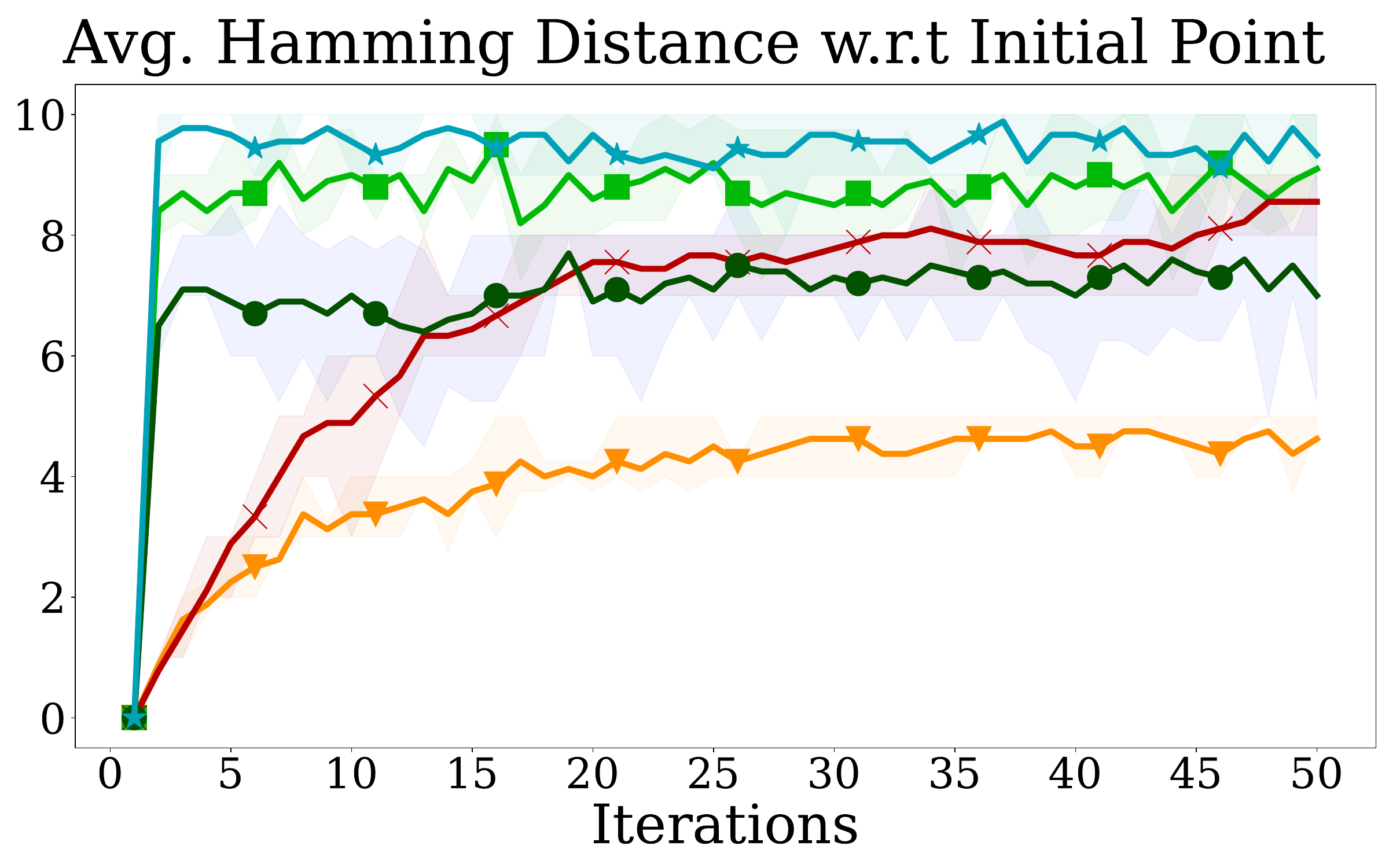}}
    \caption{\emph{GB1(55)}, $n=10$, $10$ reps.}
    \label{fig:avg-perf-plot-hamming-c}
    \end{subfigure}%
        \begin{subfigure}{0.33\textwidth}
    \centering
    {\includegraphics[width=1\linewidth,trim=0 10 0 0,clip]{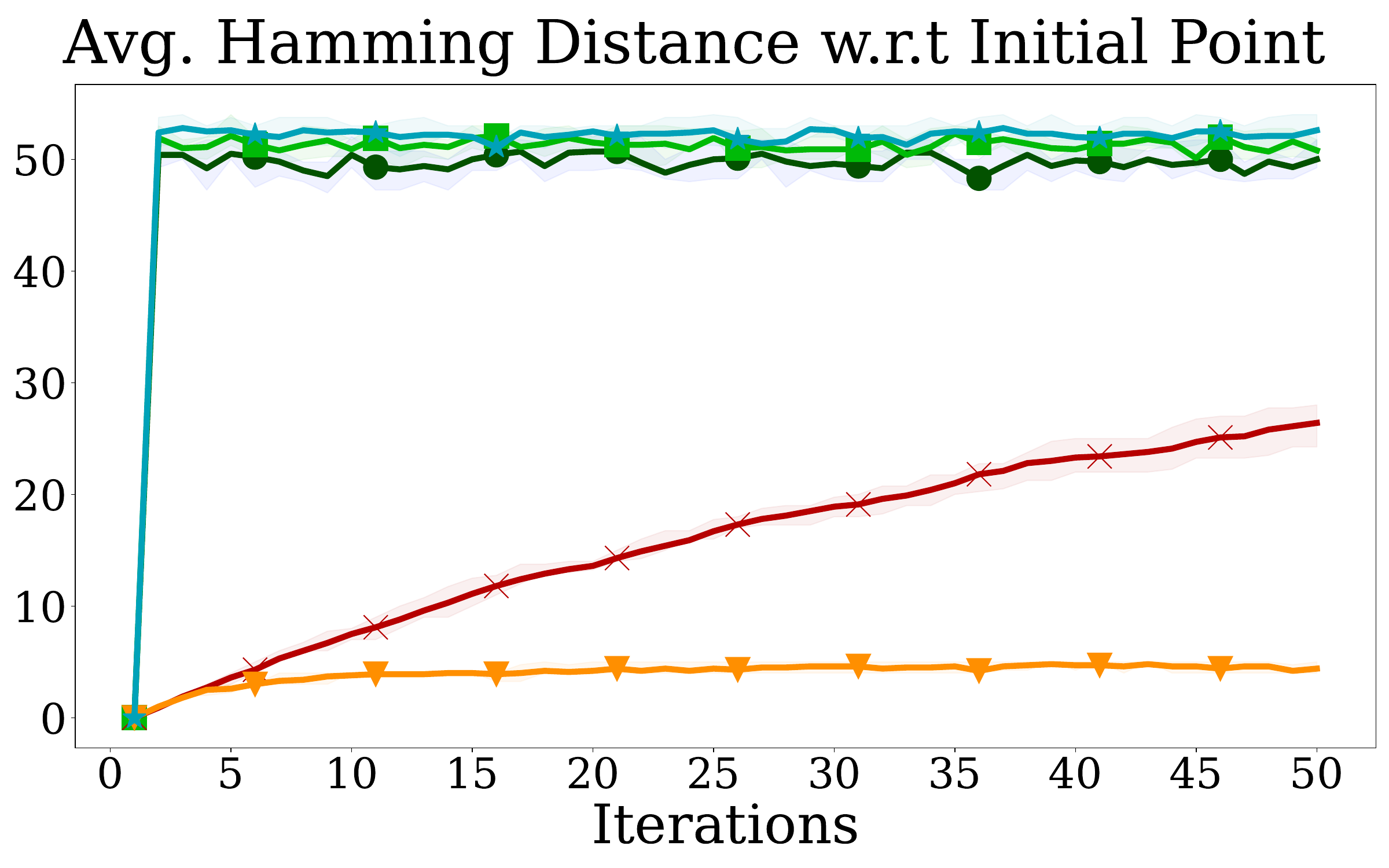}}
    \caption{\emph{GB1(55)}, $n=55$, $10$ reps.}
    \label{fig:avg-perf-plot-hamming-b}
    \end{subfigure}%
    \\
    \begin{subfigure}{1\textwidth}
    \centering
    {\includegraphics[width=1\linewidth,trim=40 20 5 15,clip]{figures/experiment_results/legend_5.pdf}}
    \label{fig:subfig1}
    \end{subfigure}%
\setlength{\abovecaptionskip}{-2pt}
 \caption{
 Performance results for the sampled batch diversity w.r.t past measured via mean Hamming distance between the executed variant and the closest initial point from the training set, under batch size $B=5$. Each point on each line is the average of multiple replications initiated with different training sets having $100$ variants for \emph{GB1(4)} \& \emph{Halogenase} and $1000$ for \emph{GB1(55)} \& \emph{GFP}. Similarly, error bars are interquartile ranges averaged over replications.
 In all experiments, \textsc{GameOpt} explores significantly faster than the baseline methods.}
\label{fig:avg-perf-plot-hamming}
\end{figure*}

The results depicted in Figure~\ref{fig:avg-perf-plot-hamming} underscore that \textsc{GameOpt} explores at a faster rate compared to baselines except for \textsc{Random}. Notably, even in the initial iterations, \textsc{GameOpt} demonstrates the capability to discover points beyond the trust region of its GP prior. 
Furthermore, it consistently upholds the sampled batch diversity compared to previously executed strategies. As also illustrated in Figure~\ref{fig:avg-perf-plot-hamming-2}, \textsc{GameOpt} explores effectively at the beginning and gradually converges to a region conducive to exploitation. 
This enhanced exploration across the search space contributes to its outperforming performance in identifying high fitness-valued protein sequences.

On the other hand, the exploration strategy employed by \textsc{Random} relies on the generation of $B$ best responses through random selection, a method that does not consistently ensure a diverse sampled batch in the input space.  Furthermore, \textsc{IBR-Fitness} shows a moderate sampled batch diversity concerning the past, attributed to its more exploitative nature—specifically, the sampling of $B$ best responses based on true log fitness values in comparison to other baseline methods. 
While \textsc{PR} manages to maintain a diverse sampled batch concerning the past in the context of \emph{GB1(4)}, its performance falters when applied to other settings. 
Additionally, the sampling process of \textsc{PR} involves computing the expected UCB across all potential strategy combinations of players, making its performance, and consequently its sampled batch diversity, highly reliant on an accurate estimate of this expectation. 

\subsection{Comparison with GP-UCB} 
\label{app:experiment-results-gp-ucb}

\begin{figure*}[h]
    \centering  
    \begin{subfigure}{0.33\textwidth}
    \centering
    {\includegraphics[width=1\linewidth,trim=10 10 0 10,clip]{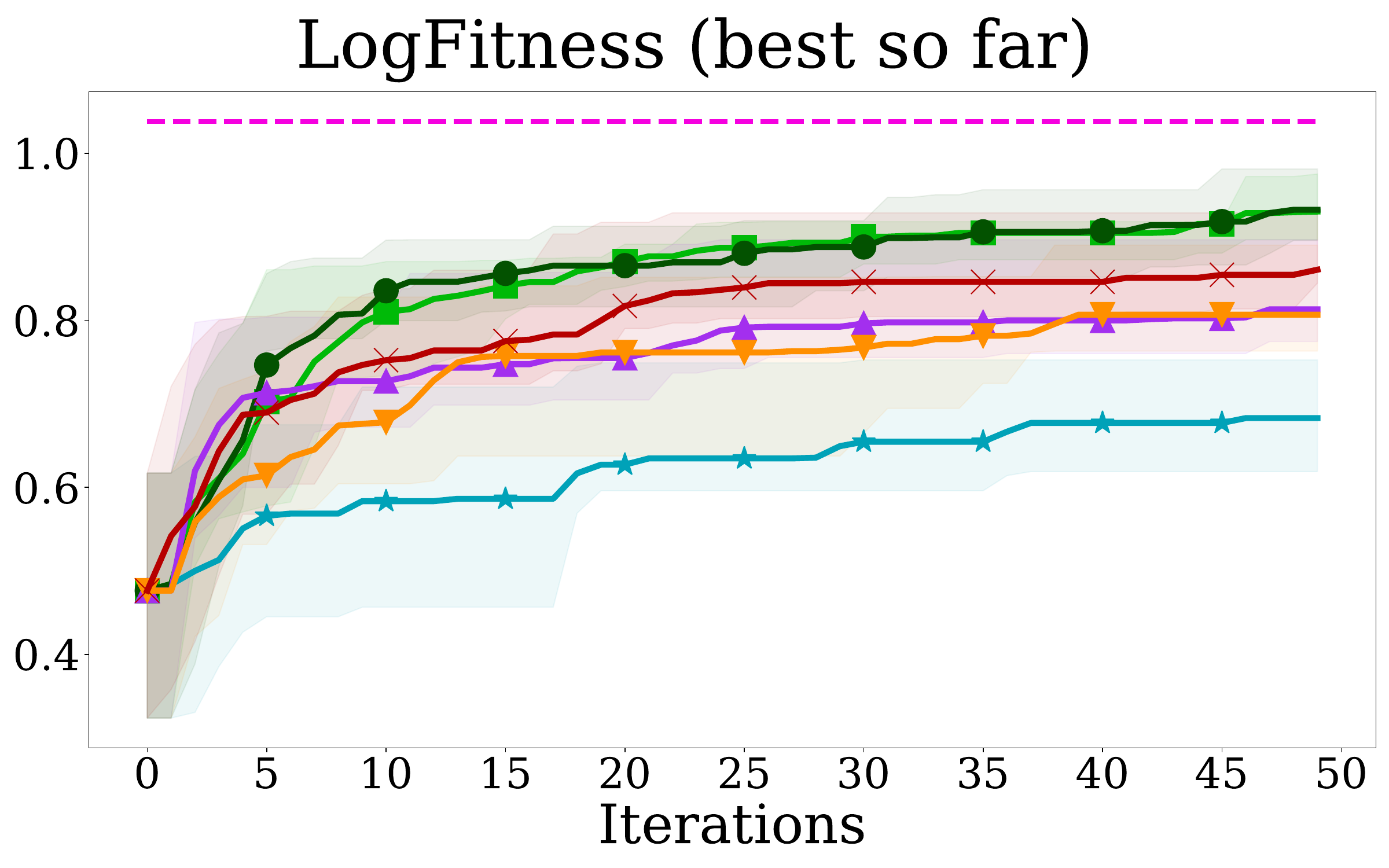}}
    \caption{$100$ training points.}
    \label{fig:avg-perf-plot-a}
    \end{subfigure}%
    \begin{subfigure}{0.33\textwidth}
    \centering
    {\includegraphics[width=1\linewidth,trim=10 10 0 10,clip]{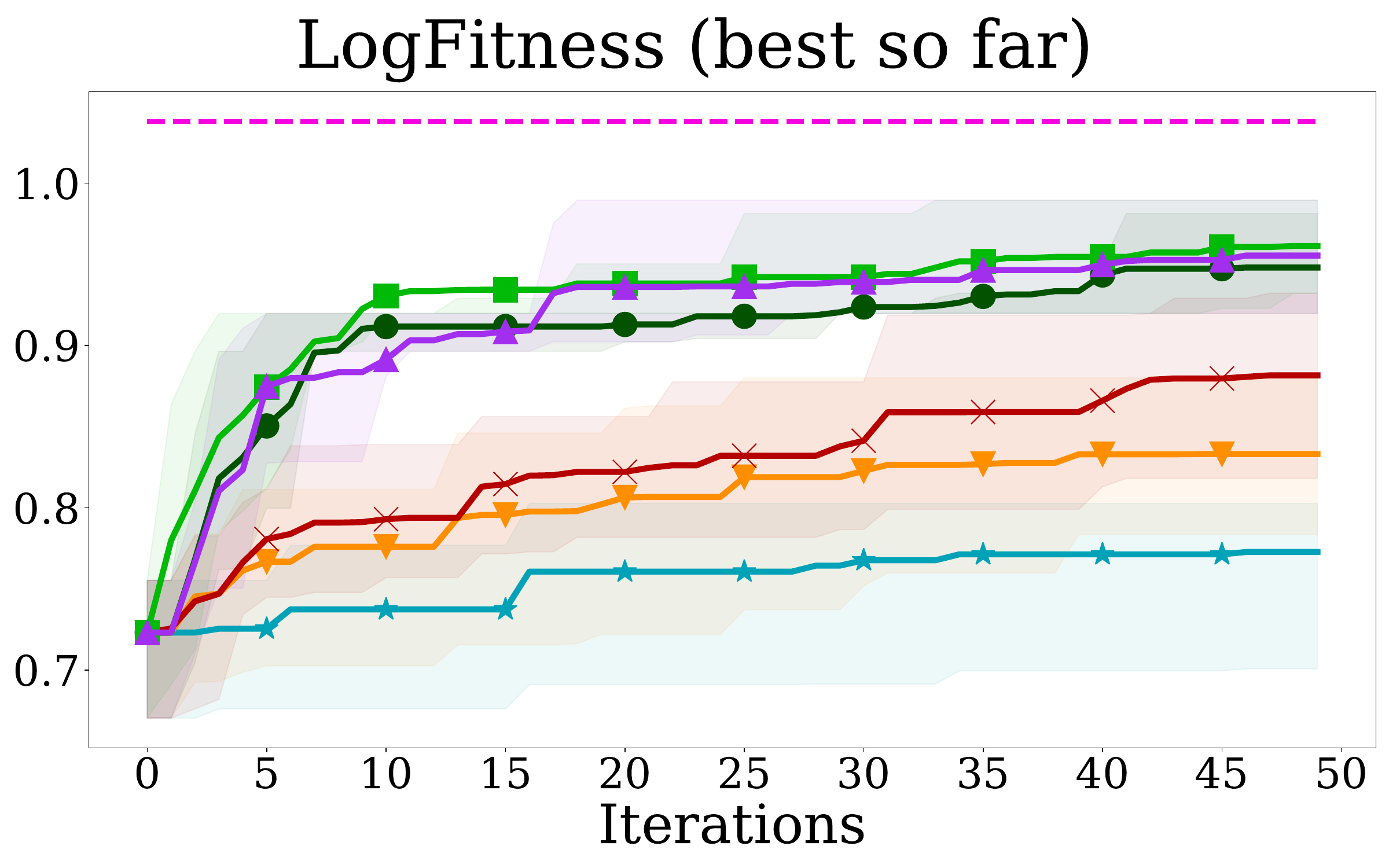}}
    \caption{$500$ training points.}
    \label{fig:avg-perf-plot-b}
    \end{subfigure}%
    \\
    \begin{subfigure}{1\textwidth}
    \centering
    {\includegraphics[width=1\linewidth,trim=40 20 5 15,clip]{figures/experiment_results/legend_5.pdf}}
    \label{fig:subfig1}
    \end{subfigure}%
 \setlength{\abovecaptionskip}{-5pt}
 \caption{The effect of training set size on the performance under setting \emph{GB1(4)}, $n=4$, $18$ reps. \textsc{GP-UCB} mitigates saturating behavior when leveraging a more informative initial GP surrogate model. In contrast, \textsc{GameOpt} showcases resilience in overcoming the limitations associated with a GP model trained with a limited amount of data.
 }
\label{fig:gp-ucb-comparison}
\end{figure*}

We further analyzed the saturating behavior exhibited by \textsc{GP-UCB} in the \emph{GB1(4)} setting. As depicted in Figure~\ref{fig:gp-ucb-comparison}, our investigation focused on the influence of the initial GP surrogate model, considering different training set sizes, specifically with $100$ and $500$ training points. 

Our findings underscore that the efficacy of \textsc{GP-UCB} heavily relies on the quality of the initial GP surrogate model. 
Particularly, an initial GP surrogate trained with only $100$ data points proves insufficient for \textsc{GP-UCB} to effectively identify high-log fitness protein sequences.
Given that \textsc{GP-UCB} optimizes the UCB globally and selects the $B$ best points in each iteration, the limited informativeness of sampled batch points under this GP surrogate constrains the algorithm. 
Therefore, \textsc{GP-UCB} ends up converging to a point where further improvement is impeded. 
In contrast, employing a potentially more informative GP model with $500$ training points enables \textsc{GP-UCB} to perform comparably to \textsc{GameOpt}.
Our proposed approach, however, exhibits robustness by overcoming the constraints associated with a model initialized with limited data. Through computing evaluation points as the equilibria of cooperative game-playing, it consistently gathers diverse and informative batches to guide the GP, thereby enhancing sample efficiency.

\subsection{Comparison with Discrete Local Search methods}
\label{app:comp-w-dls}

We further compared \textsc{GameOpt} against Discrete Local Search baselines \textsc{DLS} and \textsc{DLS\_Best} \citep{botorch} that perform a discrete local search by exploring k-Hamming distance neighborhood. While \textsc{DLS} employs random initialization, \textsc{DLS\_Best} starts the local search from the best-discovered sequence. 
In our experiments, we set their neighborhood size as 2-Hamming distance and let the baselines greedily select $B$ sequences at each iteration. For a fair comparison against \textsc{DLS\_Best}, we also run the best-discovered sequence initialization version of our approach, called \textsc{GameOpt-Ibr\_Best}. 

We remark that \textsc{DLS} and \textsc{DLS\_Best} can be seen as \textit{constrained} versions of \textsc{GameOpt} that require a prior definition of the neighborhood (thus, unlike our approach, they require a notion of distance too). Moreover, being \emph{centralized}, they are subject to a higher computational complexity which grows \emph{exponentially} with the neighborhood size.
For a sequence of length $n$ (players) with $\mid \mathcal{X}^{(i)}\mid=d$ many amino acid choices, \textsc{GameOpt} reduces the intractable optimization of acquisition function (i.e., $O(d^n)$) to $O(nd)$ complexity. Instead, \textsc{DLS} \& \textsc{DLS\_Best} search over $k$-distance neighborhoods yielding $O(C(n,k)d^k)$ complexity, where $C$ denotes the combination operation and $k$ is the Hamming distance.
We observe \textsc{GameOpt} performs comparably to \textsc{DLS} baselines (see Figures~\ref{fig:avg-perf-plot-rebuttal}, \ref{fig:avg-perf-plot-hamming-2-rebuttal}, \ref{fig:avg-perf-plot-hamming-rebuttal}, and additional analyses below), but requiring a significantly lower compute cost, as shown in \Cref{app:runtimes}.

\clearpage
\subsection{Computational costs}
\label{app:runtimes}

We report the total compute amount (measured in hours) for the experiments with $T=50$ BO iterations in Table~\ref{tab:total-compute}. 
We conduct the experiments on an internal compute cluster equipped with NVIDIA A100 80 GB tensor-core GPUs.
For smaller domains (with short protein sequences), we were able to execute replications concurrently without encountering memory errors. However, in configurations necessitating the generation of embeddings for numerous evaluation points with longer protein sequences in each iteration, parallel execution was not feasible, thus necessitating a per-replication computation time report.
Specifically, we present the total compute amount for \textit{all} replications across domains \emph{Halogenase}, \emph{GB1(4)}, and \emph{GFP}; whereas we report \textit{per replication} compute amount accompanied by its standard deviation for the two largest \emph{GB1(55)} domains.

Referring to Table~\ref{tab:total-compute}, \textsc{GameOpt} variations provide \textit{tractable} acquisition function optimization. Their computational demand predominantly hinges on the number of sequences they evaluate per BO iteration. 
This is also influenced by the number of game rounds--which we intentionally set to a higher value to guarantee convergence to game equilibrium (see Table~\ref{tab:params-and-hyperparams} for the exact values). Nevertheless, our observations reveal that game equilibrium is often reached well before the rounds are completed. Consequently, the computational times reported for \textsc{GameOpt} can be considered to be within the \textit{worst-case scenario}.

\begin{table*}[h]
\caption{ Total compute amount (in hours) required for experiments with $T=50$ BO iterations. 
Each entry with $\pm$ represents the average \textit{per-replication} computation time, accompanied by their standard deviation, whereas the others show the total compute time \textit{for all} replications.
We run $18$ replications for \emph{Halogenase}, \emph{GB1(4)}, and $10$ replications for the other domains. 
In all domains, particularly for the larger search spaces, \textsc{GameOpt} provides tractable acquisition function optimization.
 }
\label{tab:total-compute}
\centering
\resizebox{1\linewidth}{!}{%
\begin{tabular}{l|c|c|c|c|c|c}
\specialrule{1.5pt}{1pt}{1pt} 
\multirow{2}{*}{\textbf{Method}} & \multicolumn{6}{c}{\textbf{Domain}} \\
\cline{2-7}
 &  Halogenase & GB1(4) & GFP, $n=6$ & GFP, $n=8$ & GB1(55), $n=10$ & GB1(55), $n=55$ \\
\specialrule{1.5pt}{1pt}{1pt} 
\textsc{GameOpt-Ibr} & $\boldsymbol{0.19}$ & $\boldsymbol{0.30}$ & $\boldsymbol{1.02}$ & $\boldsymbol{2.74}$ & $\boldsymbol{1.61 \pm 0.46}$
& $\boldsymbol{11.25 \pm 3.19}$  \\
\textsc{GameOpt-Hedge} & $2.44$ & $1.95$ & $2.05 \pm 0.02$ & $2.60 \pm 0.09$ & $3.85 \pm 0.39$ & $20.13 \pm 1.04$ \\
\textsc{GP-UCB} & $0.63$ & $29.56$ &- & - & - & -  \\
\textsc{DLS} & $0.29$ & $1.10$ & $2.90 \pm 0.33$ & $4.94 \pm 0.24$ &  
 $4.10 \pm 2.36$ & $102.86 \pm 27.42$ \\
 \specialrule{1.5pt}{1pt}{1pt} 
\end{tabular}
}
\end{table*}
\subsection{Additional analyses}
\label{app:comp-w-dls-additional-analyses}
We performed further analyses to assess the effectiveness of our \textsc{GameOpt} framework against baselines in terms of: $(1)$ fraction of global optima discovered, $(2)$  fraction of solutions found above a fitness threshold, $(3)$  cumulative maximum for sequences proposed and $(4)$ mean pairwise Hamming distance between proposed sequences.

We evaluated the fraction of global optima discovered under the \emph{GB1(4)} setting, as depicted in Table~\ref{tab:fraction-global-optima}. For this fully combinatorial dataset, we identify the global optimizer sequence by exhaustive search. Our findings revealed that \textsc{GameOpt-IBR} successfully identified the global optimum sequence in $33.33\%$ of cases out of 18 replications, highlighting its superior convergence against baselines.

Additionally, given the difficulty of identifying global optima in non-fully combinatorial datasets, we examined the fraction of optima found above a fitness threshold, $f_\tau = 0.8$. Table~\ref{tab:fraction-optima-thr} demonstrates that across all problem domains (excluding best cumbersome initialization versions), the \textsc{GameOpt} framework consistently samples batches containing a higher number of optimal sequences compared to baseline methods. Its effectiveness is highlighted more as the search space size gets higher. In the most complex domain, \emph{GB1(55)} with $55$ players, its best cumbersome initialization version performs comparably to \textsc{DLS\_Best}. However, it is essential to also acknowledge the comparison w.r.t the computational complexity associated with that configuration as discussed in Appendix~\ref{app:runtimes}. 

\begin{figure*}[ht]
    \centering  
    \begin{subfigure}{0.33\textwidth}
    \centering
    {\includegraphics[width=1\linewidth,trim=10 10 0 10,clip]{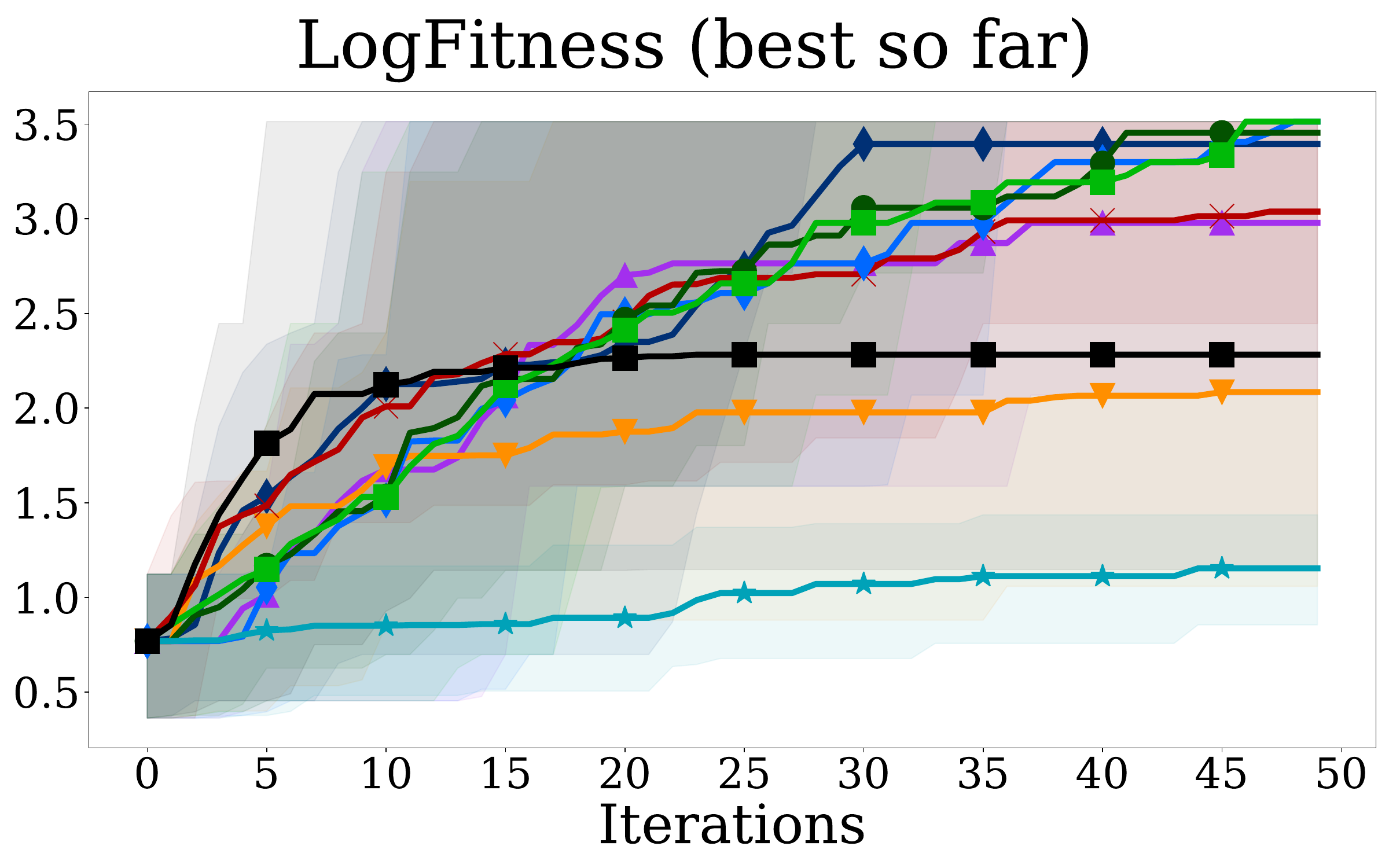}}  
    \caption{\emph{Halogenase}, $n=3$, $18$ reps.}
    \label{fig:avg-perf-plot-d}
    \end{subfigure}%
    \begin{subfigure}{0.33\textwidth}
    \centering
    {\includegraphics[width=1\linewidth,trim=10 10 0 10,clip]{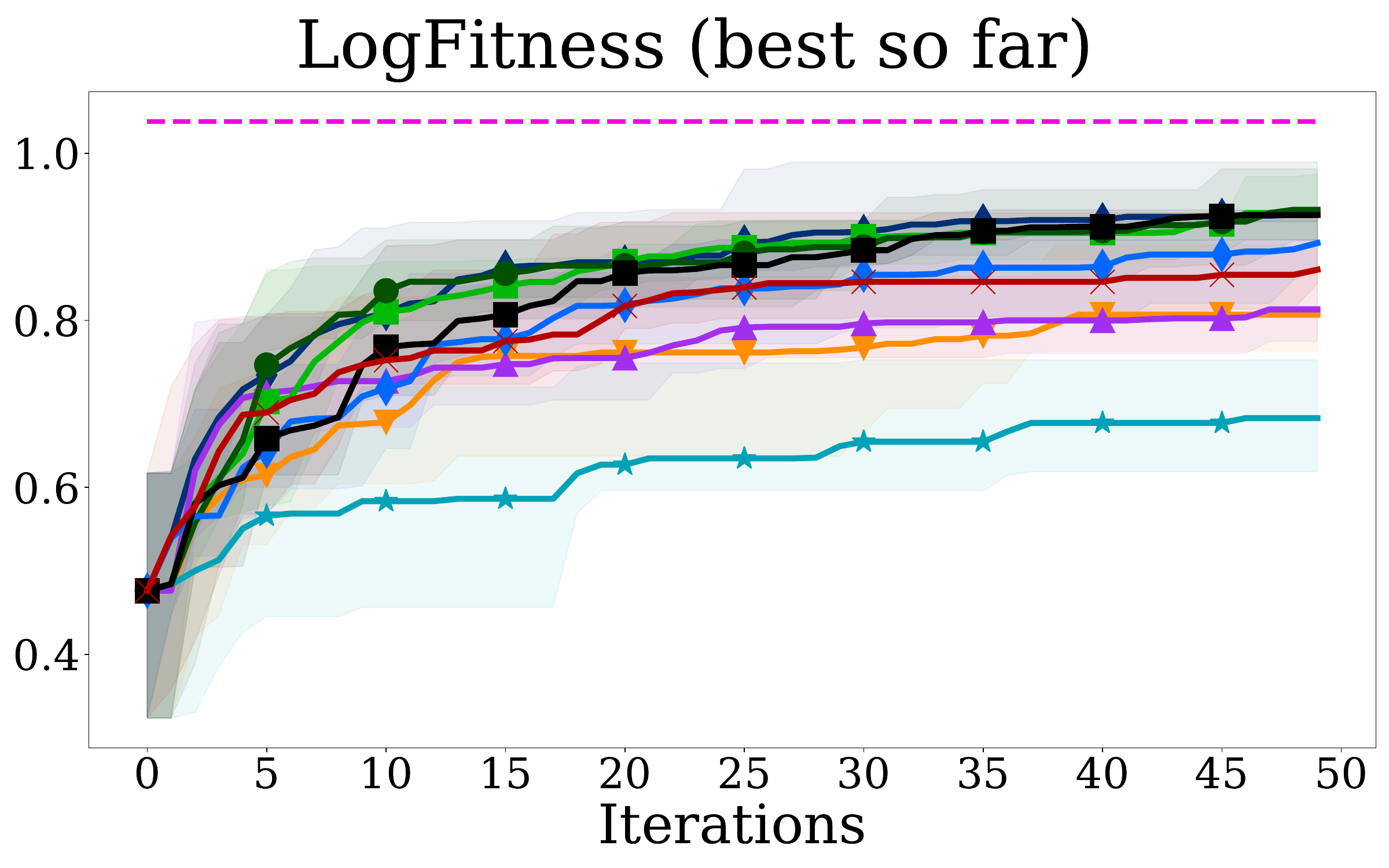}}
    \caption{\emph{GB1(4)}, $n=4$, $18$ reps.}
    \label{fig:avg-perf-plot-a}
    \end{subfigure}%
    \begin{subfigure}{0.33\textwidth}
    \centering
    {\includegraphics[width=1\linewidth,trim=10 10 0 10,clip]{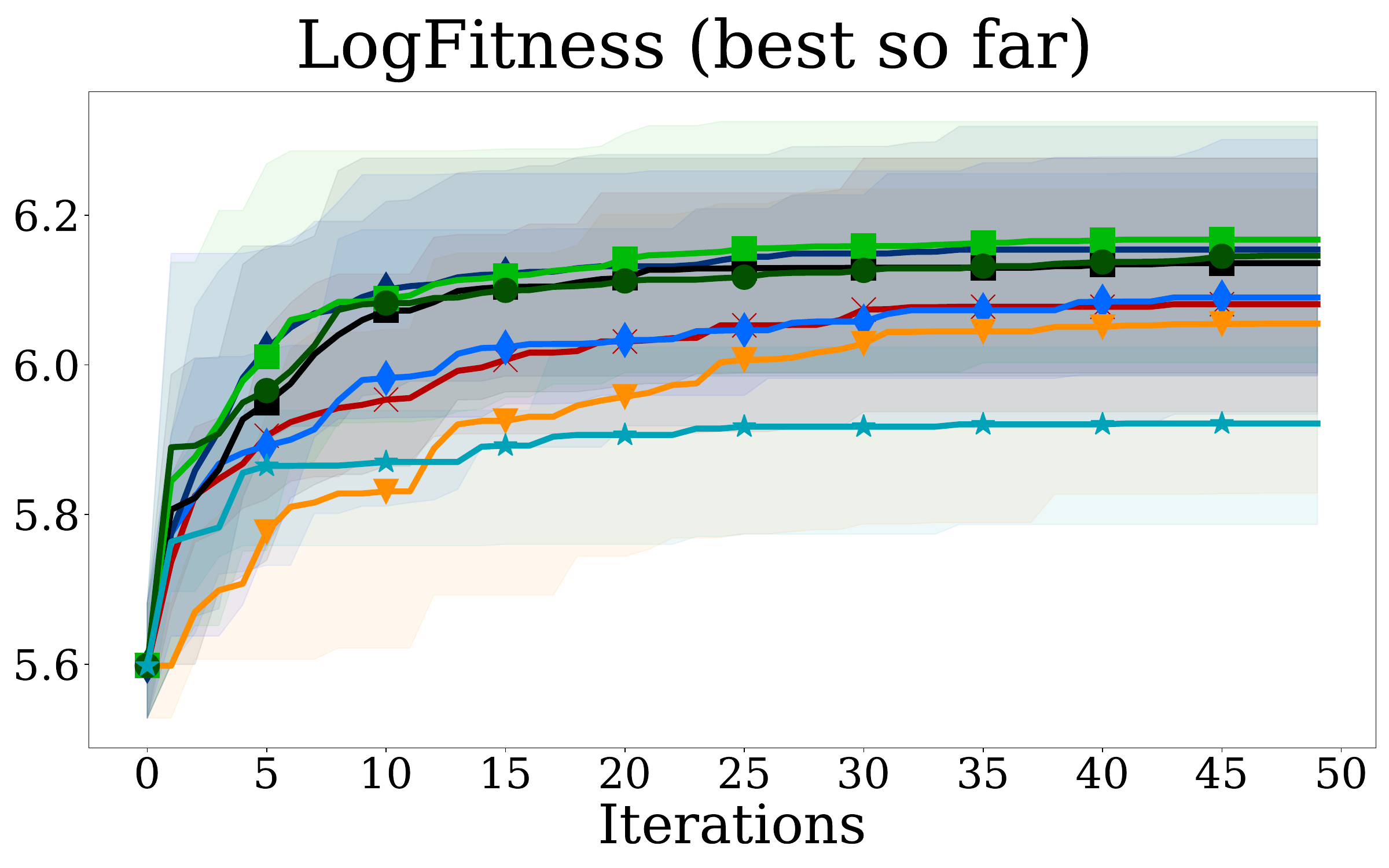}}
    \caption{\emph{GFP}, $n=6$, $10$ reps.}
    \label{fig:avg-perf-plot-h}
    \end{subfigure}%
    \\
    \vspace{1em}
    \begin{subfigure}{0.33\textwidth}
    {\includegraphics[width=1\linewidth,trim=10 10 0 10,clip]{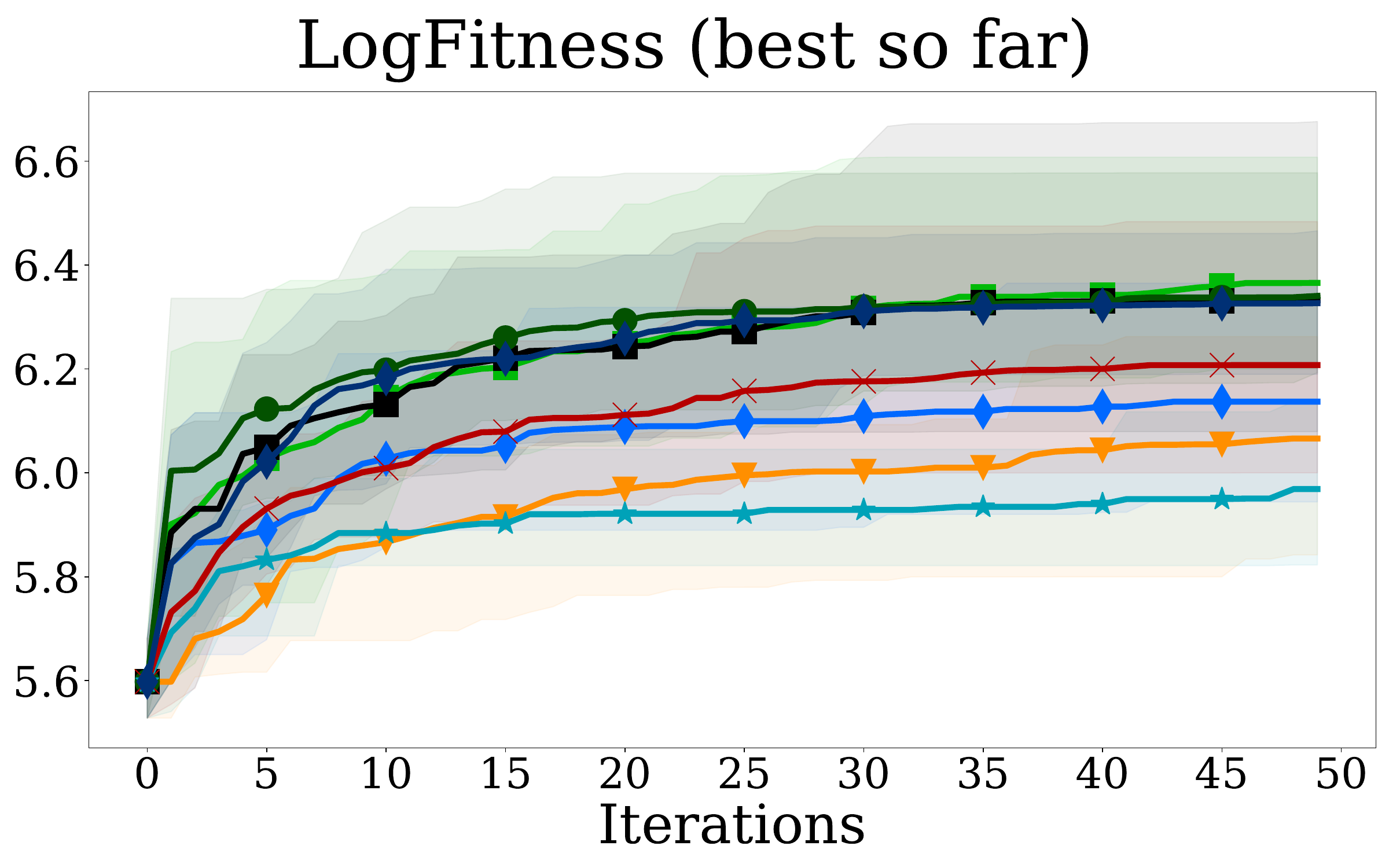}}
    \caption{\emph{GFP}, $n=8$, $10$ reps.}
    \label{fig:avg-perf-plot-e}
    \end{subfigure}%
        \begin{subfigure}{0.33\textwidth}
    \centering
    {\includegraphics[width=1\linewidth,trim=10 10 0 10,clip]{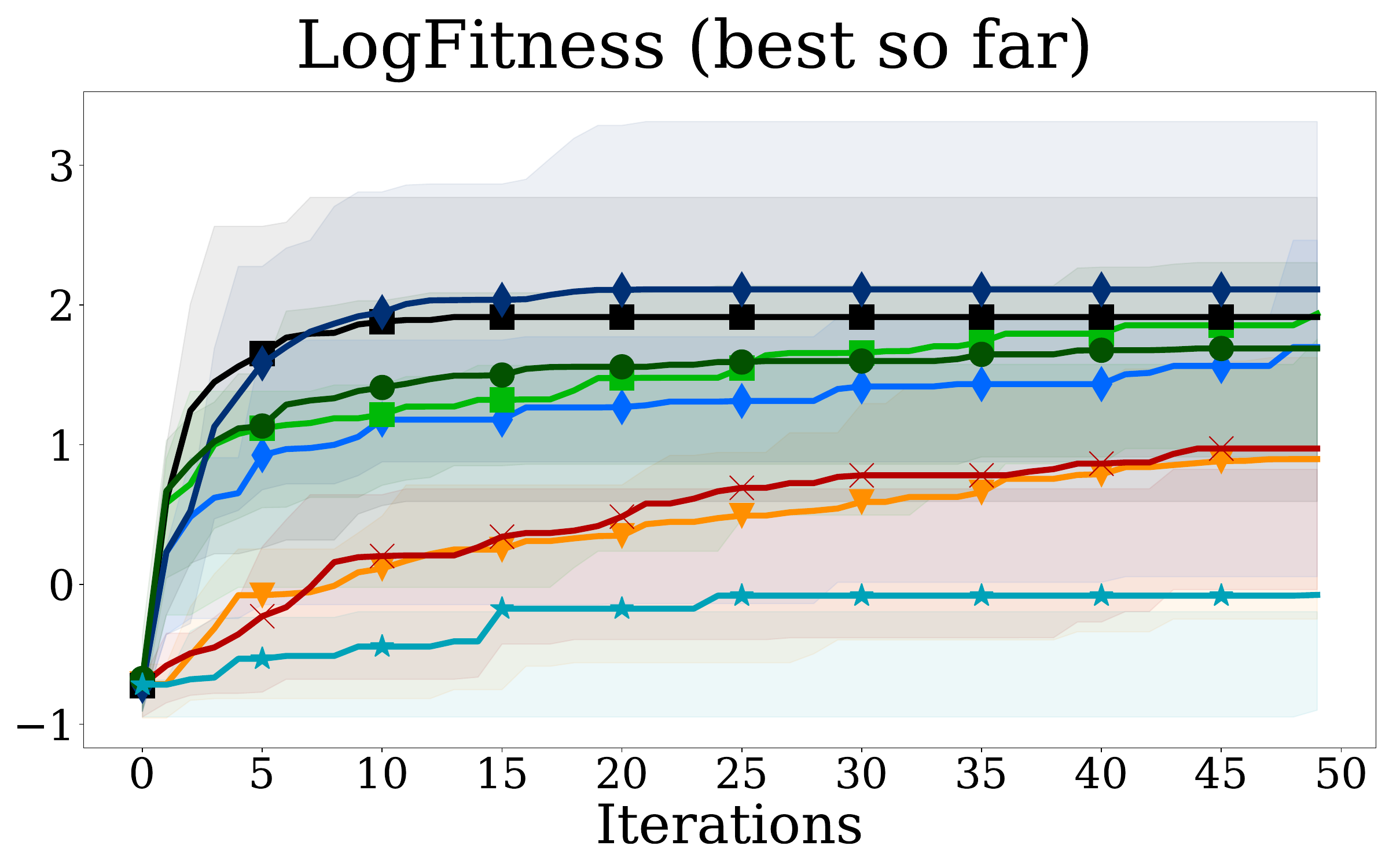}}
    \caption{\emph{GB1(55)}, $n=10$, $10$ reps.}
    \label{fig:avg-perf-plot-c}
    \end{subfigure}%
    \begin{subfigure}{0.33\textwidth}
    \centering
    {\includegraphics[width=1\linewidth,trim=10 10 0 10,clip]{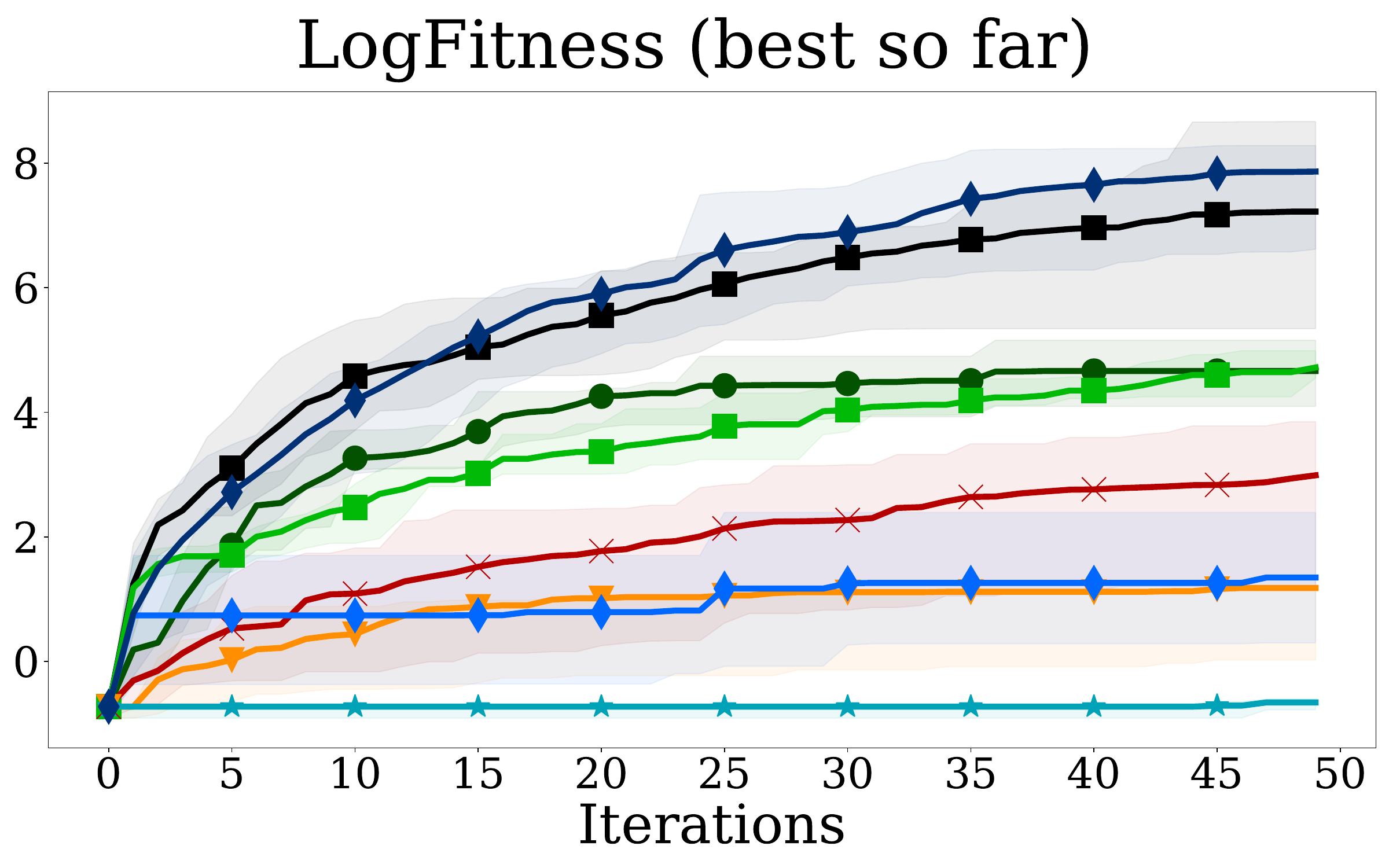}}
    \caption{\emph{GB1(55)}, $n=55$, $10$ reps.}
    \label{fig:avg-perf-plot-b}
    \end{subfigure}%
    \\
    \vspace{0.5em}
    \begin{subfigure}{1\textwidth}
    \centering
    {\includegraphics[width=1\linewidth,trim=20 5 5 10,clip]{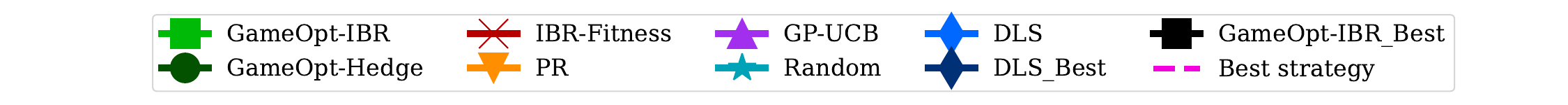}}
    \label{fig:subfig1}
    \end{subfigure}%
 \setlength{\abovecaptionskip}{-2pt}
 \caption{Convergence speed of methods in terms of log fitness value of the best-so-far protein throughout BO iterations, under batch size $B =5$. Each point is the average of multiple replications initiated with different training sets having $100$ protein variants for \emph{Halogenase} and \emph{GB1(4)}, and $1000$ protein variants for the rest of the problem domains. Similarly, error bars are interquartile ranges averaged over replications. }
 
\label{fig:avg-perf-plot-rebuttal}
\end{figure*}

Regarding the cumulative maximum for proposed sequences (see Table~\ref{tab:cumulative-maximum}), our framework demonstrates notable performance by consistently proposing protein variants with higher fitness values. Moreover, its superior convergence speed, illustrated in Figure~\ref{fig:avg-perf-plot-rebuttal}, underscores its effectiveness against baselines including \textsc{DLS}. The discrete local search with best initialization, 
\textsc{DLS\_Best}, converges relatively slower, particularly in small domains, yet performs comparably against \textsc{GameOpt\_Best} in finding high log fitness valued variants. However, it does not provide tractable optimization as detailed in Appendix~\ref{app:runtimes}. 

\begin{figure*}[ht!]
    \centering  
    \begin{subfigure}{0.33\textwidth}
    \centering
    {\includegraphics[width=1\linewidth,trim=0 10 0 0,clip]{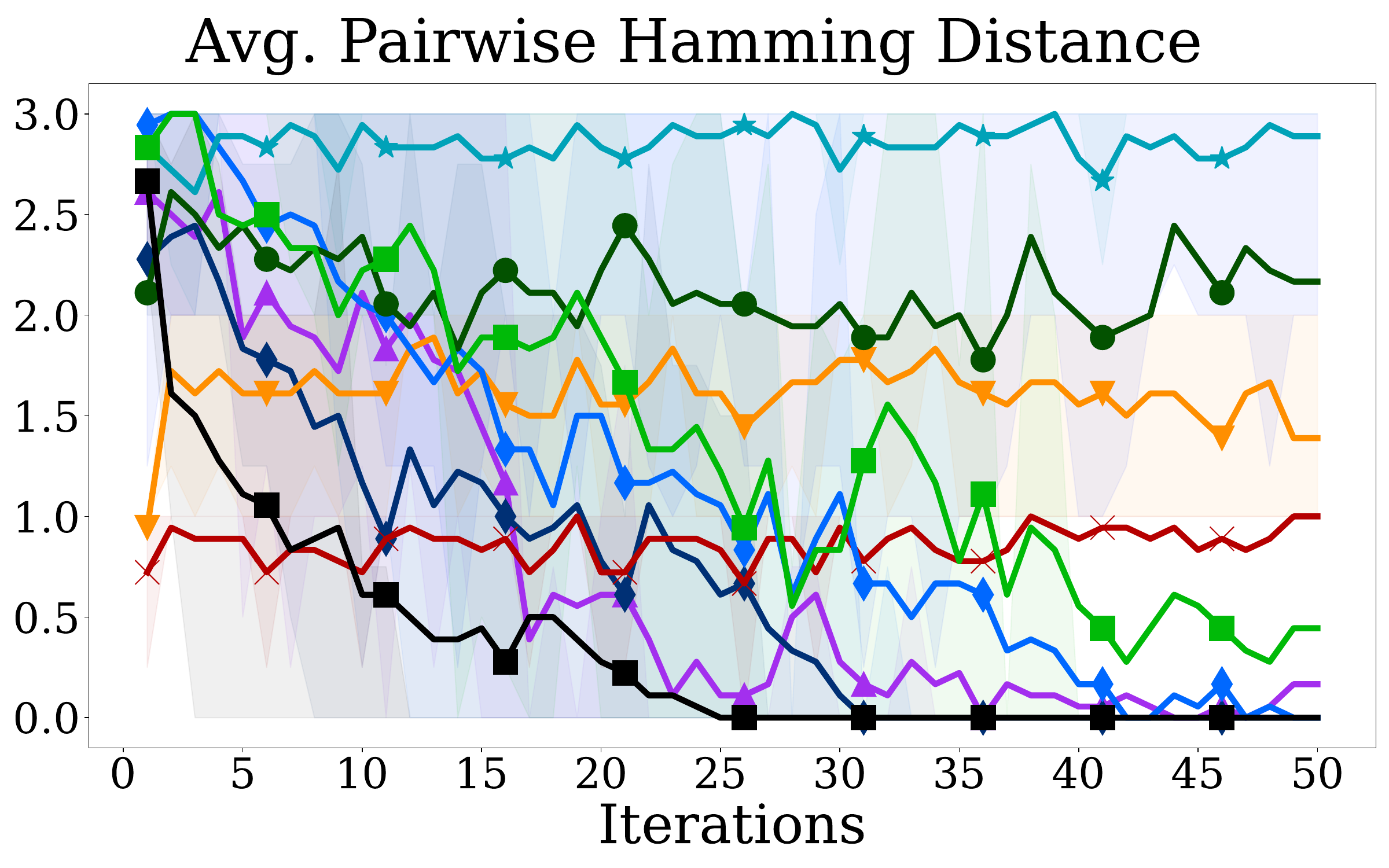}}
    \caption{\emph{Halogenase}, $n=3$, $18$ reps.}
    \label{fig:avg-perf-plot-hamming-c}
    \end{subfigure}%
    \begin{subfigure}{0.33\textwidth}\centering
    {\includegraphics[width=1\linewidth,trim=0 10 0 0,clip]{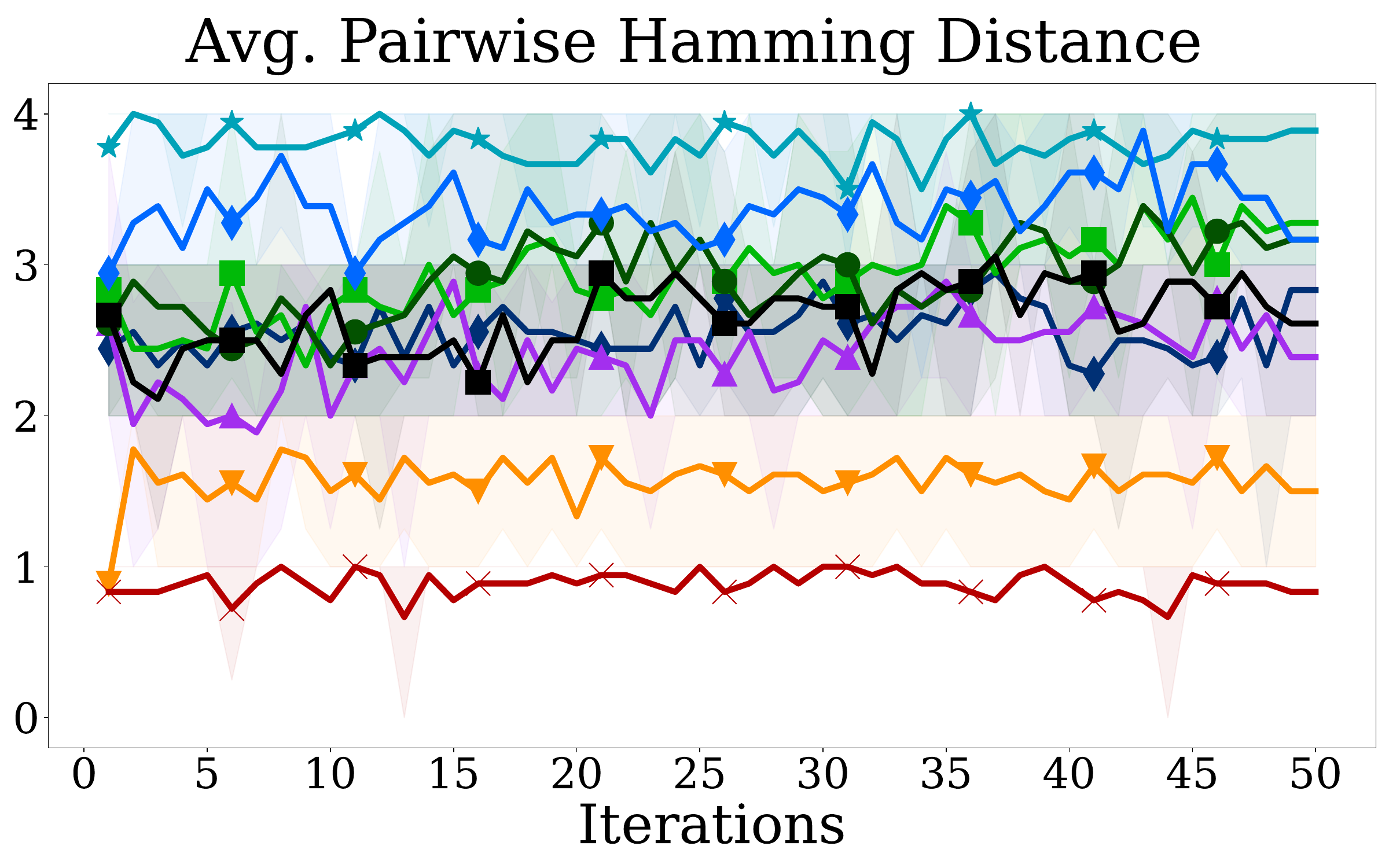}}
    \caption{\emph{GB1(4)}, $n=4$, $18$ reps.}
    \label{fig:avg-perf-plot-hamming-e}
    \end{subfigure}%
    \begin{subfigure}{0.33\textwidth}
    \centering
    {\includegraphics[width=1\linewidth,trim=0 10 0 0,clip]{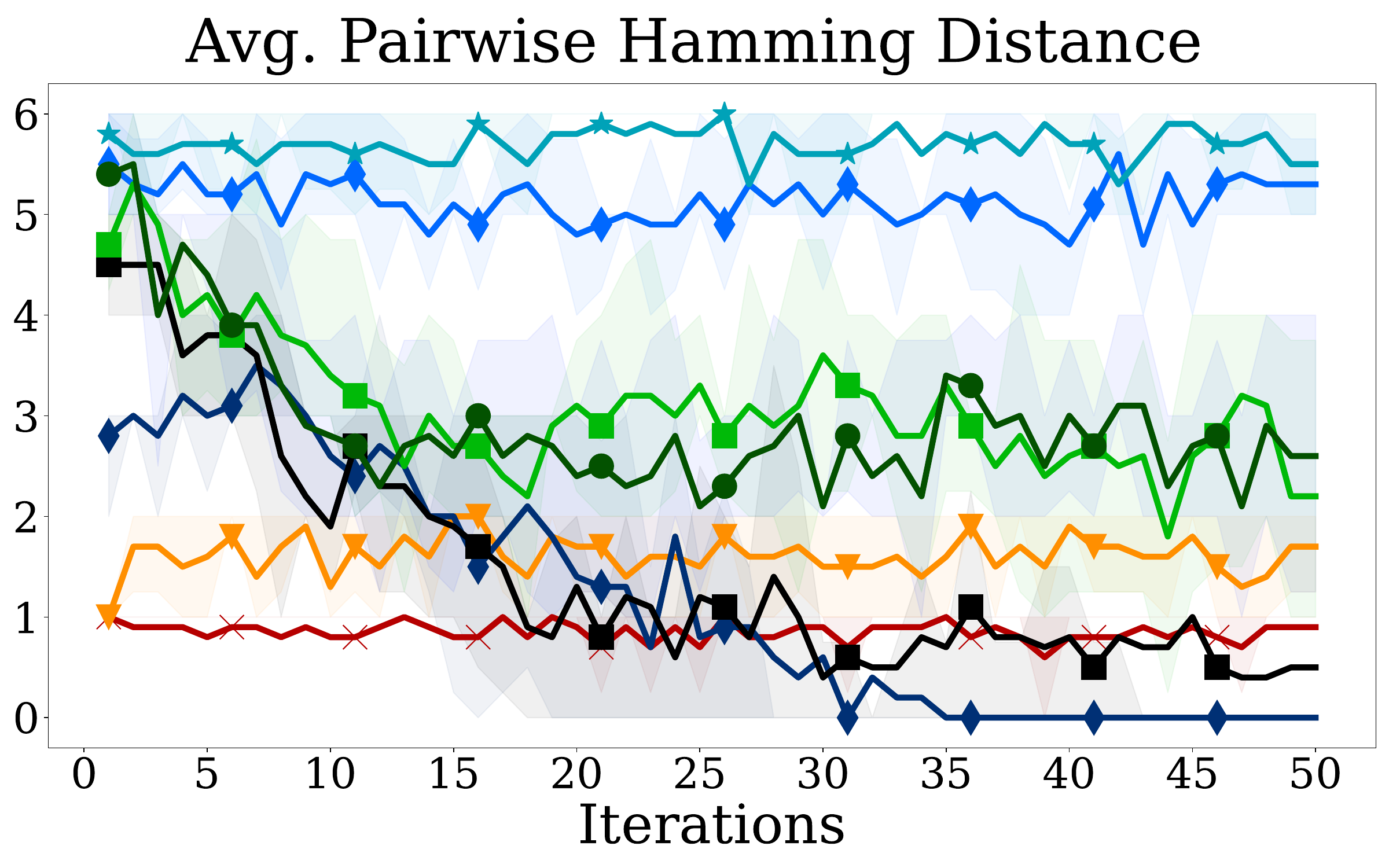}}
    \caption{\emph{GFP}, $n=6$, $10$ reps.}
    \label{fig:avg-perf-plot-hamming-h}
    \end{subfigure}%
    \\
    \begin{subfigure}{0.33\textwidth}
    \centering
    {\includegraphics[width=1\linewidth,trim=0 10 0 0,clip]{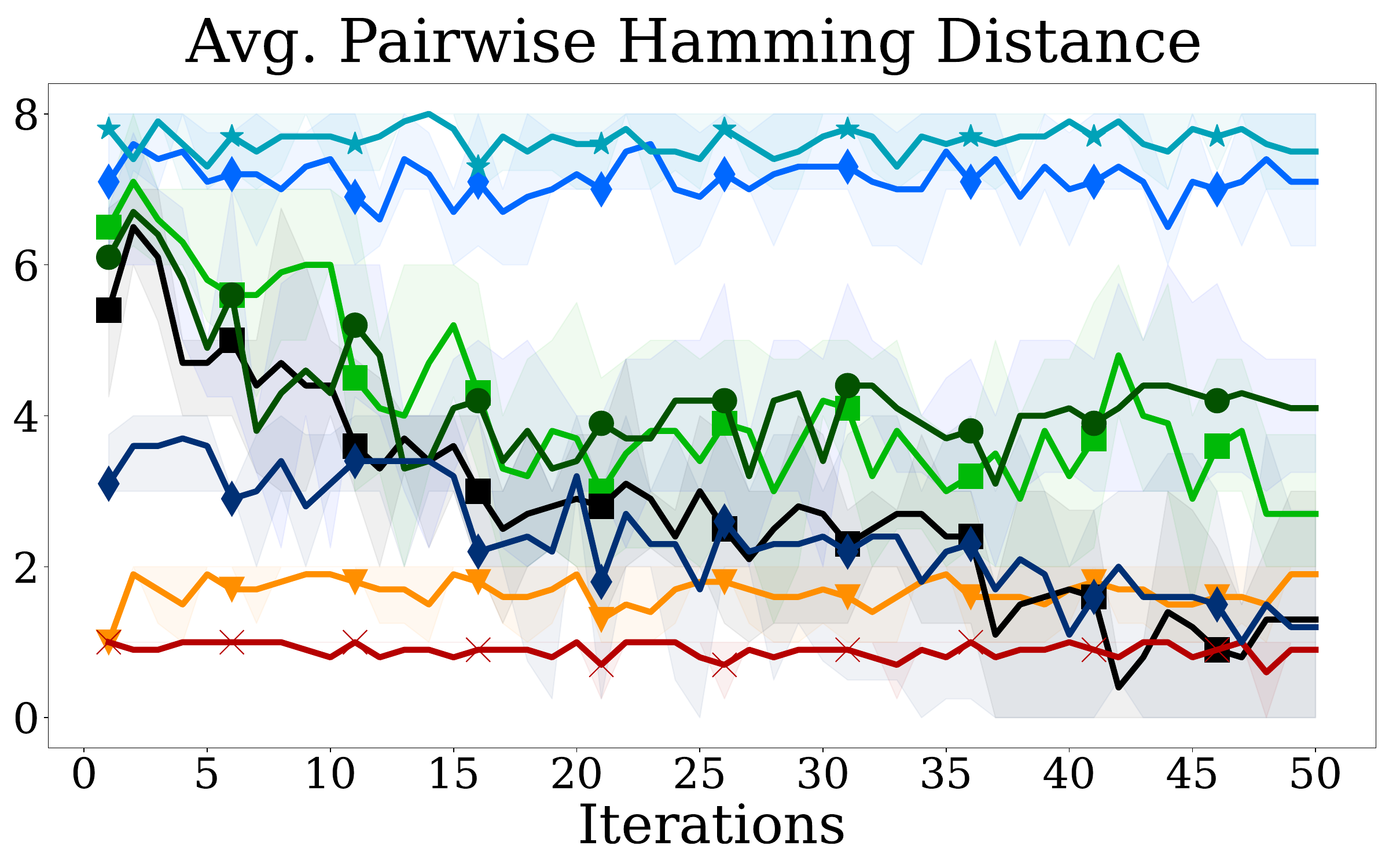}}
    \caption{\emph{GFP}, $n=8$, $10$ reps.}
    \label{fig:avg-perf-plot-hamming-d}
    \end{subfigure}%
        \begin{subfigure}{0.33\textwidth}
    \centering
    {\includegraphics[width=1\linewidth,trim=0 10 0 0,clip]{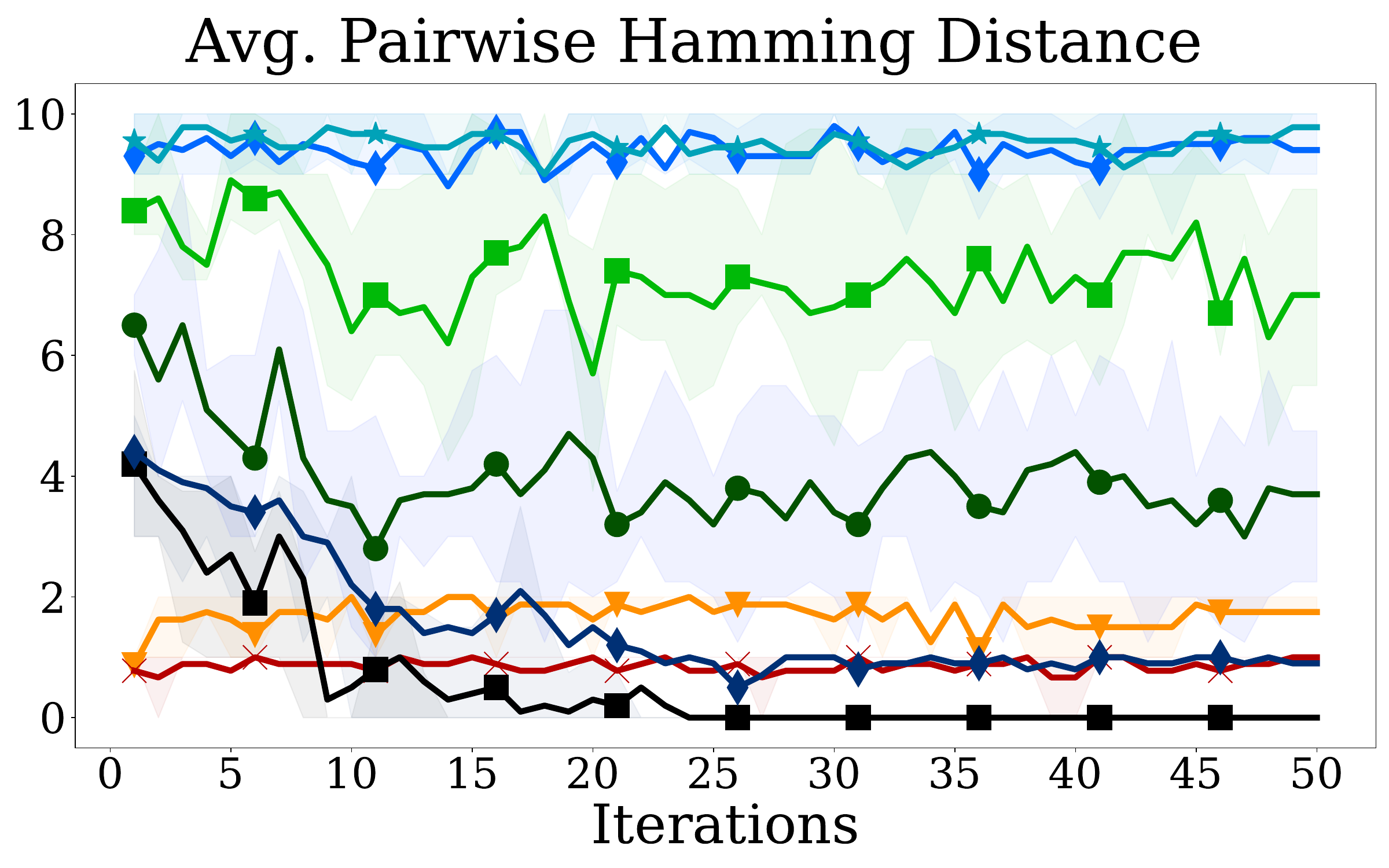}}
    \caption{\emph{GB1(55)}, $n=10$, $10$ reps.}
    \label{fig:avg-perf-plot-hamming-g}
    \end{subfigure}%
    \begin{subfigure}{0.33\textwidth}
    \centering
    {\includegraphics[width=1\linewidth,trim=0 10 0 0,clip]{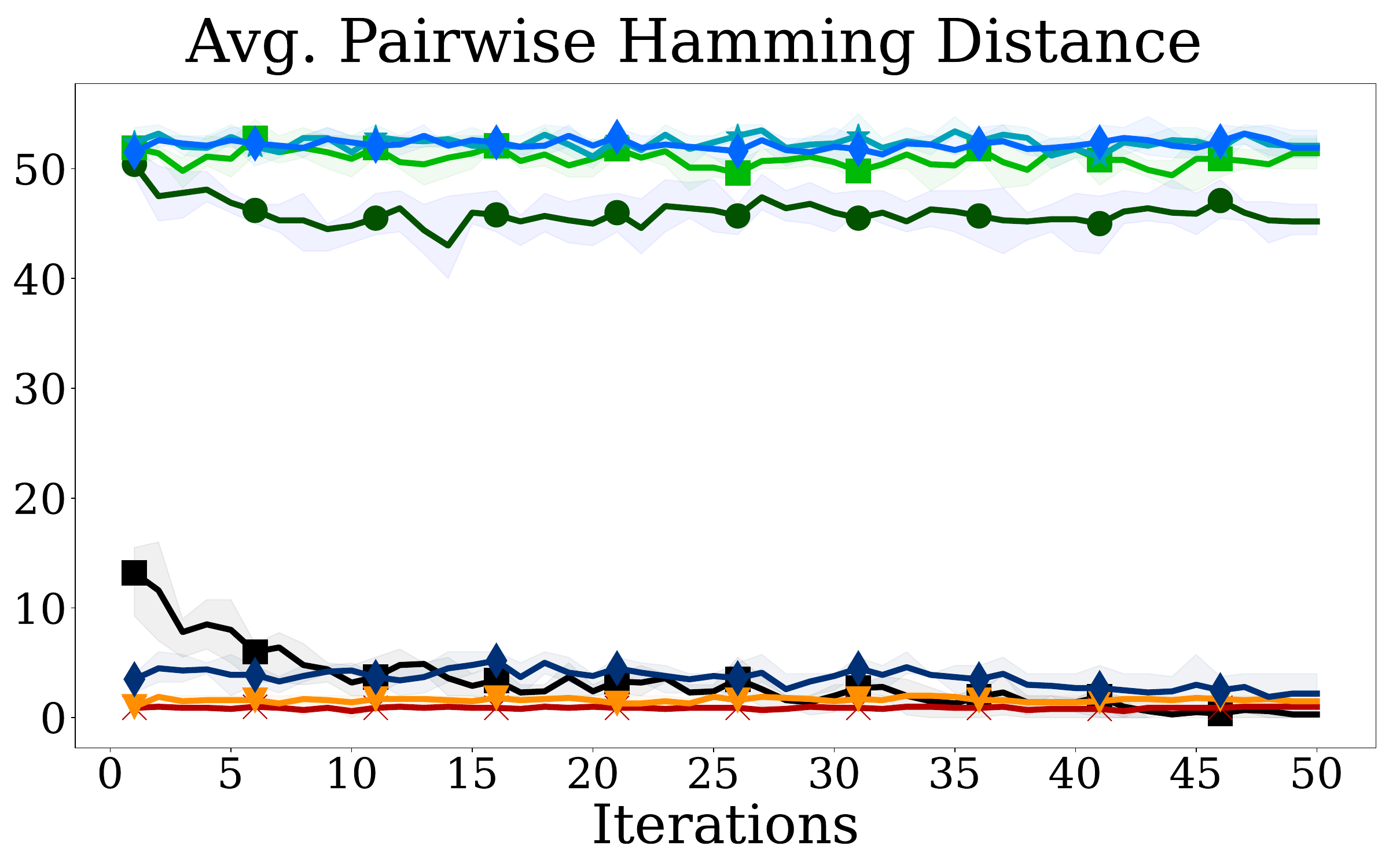}}
    \caption{\emph{GB1(55)}, $n=55$, $10$ reps.}
    \label{fig:avg-perf-plot-hamming-f}
    \end{subfigure}%
    \\
    
    \vspace{0.5em}
    
    \begin{subfigure}{1\textwidth}
    \centering
    {\includegraphics[width=1\linewidth,trim=20 5 5 10,clip]{figures/experiment_results/legend_rebuttal_best_2.pdf}}
    \label{fig:subfig1}
    \end{subfigure}%
\setlength{\abovecaptionskip}{-5pt}
 \caption{Sampled batch diversity w.r.t past measured via mean Hamming distance between the executed variant and the proposed variant from the previous iteration (pairwise distance), under batch size $B=5$. Each point on each line is the average of multiple replications initiated with different training sets having $100$ variants for \emph{GB1(4)} \& \emph{Halogenase} and $1000$ for \emph{GB1(55)} \& \emph{GFP}. Similarly, error bars are interquartile ranges averaged over replications.
 In all experiments \textsc{GameOpt} consistently samples a rather diverse batch of evaluation points w.r.t. past proposed variants. This enhanced exploration of the search space contributes to its strong performance compared to the baseline methods.}
\label{fig:avg-perf-plot-hamming-2-rebuttal}
\vspace{-0.15cm}
\end{figure*}

Furthermore, we analyzed the performance of methods considering the mean pairwise Hamming distance between the sequences proposed at each BO iteration (see Table~\ref{tab:mean-hamming-distance} and Figure~\ref{fig:avg-perf-plot-hamming-2-rebuttal}) which is an indicator for sampled batch diversity. The results indicate that \textsc{GameOpt} explores moderately, however, it balances exploration and exploitation to prevent over-exploring seen in \textsc{Random} and \textsc{DLS}, as well as over-exploiting observed in \textsc{GameOpt\_Best}, \textsc{DLS\_Best} and \textsc{PR}.

\begin{figure*}[ht!]
    \centering  
    \begin{subfigure}{0.33\textwidth}
    \centering
    {\includegraphics[width=1\linewidth,trim=0 10 0 0,clip]{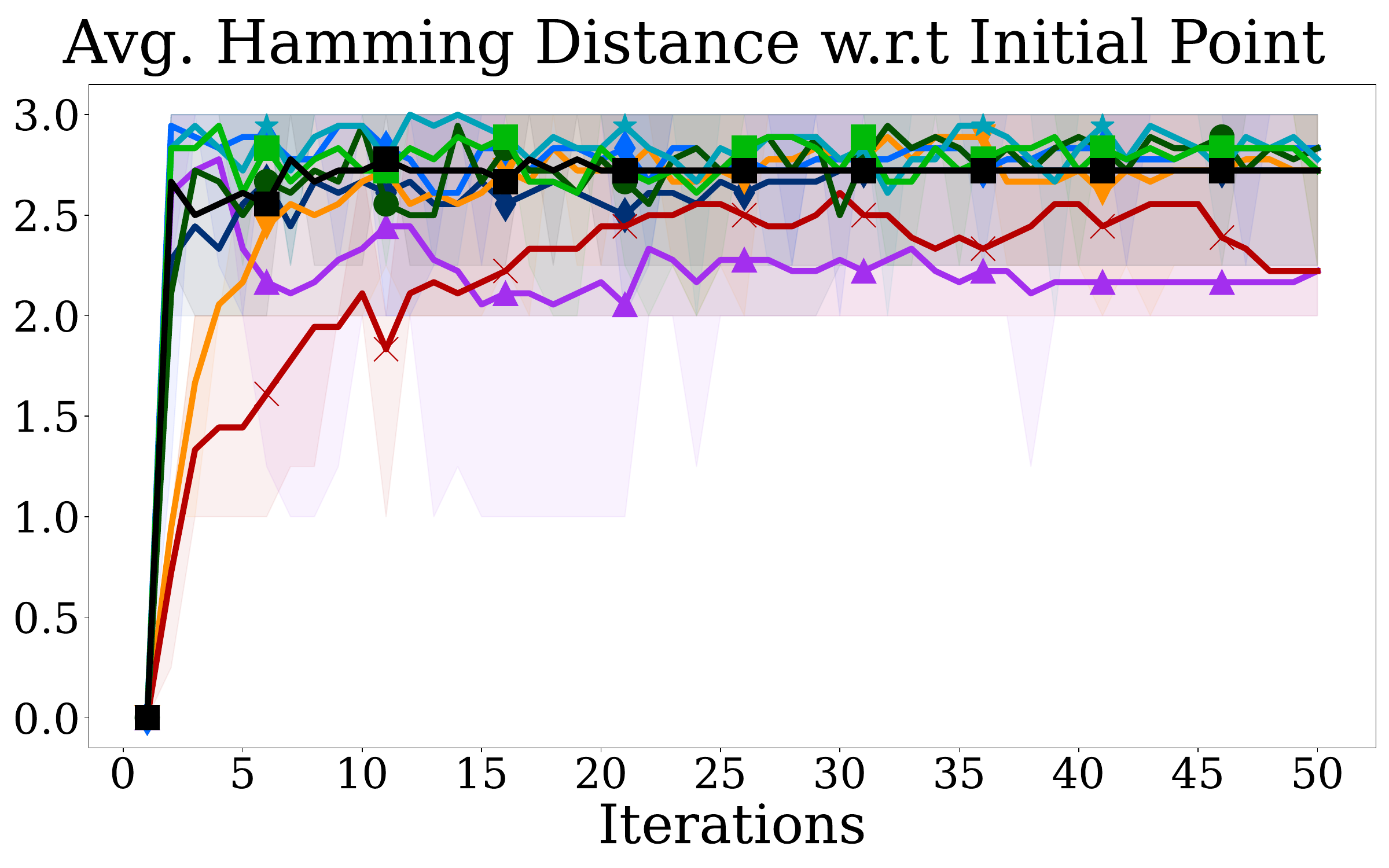}}
    \caption{\emph{Halogenase}, $n=3$, $18$ reps.}
    \label{fig:avg-perf-plot-hamming-d}
    \end{subfigure}%
    \begin{subfigure}{0.33\textwidth}
    \centering
    {\includegraphics[width=1\linewidth,trim=0 10 0 0,clip]{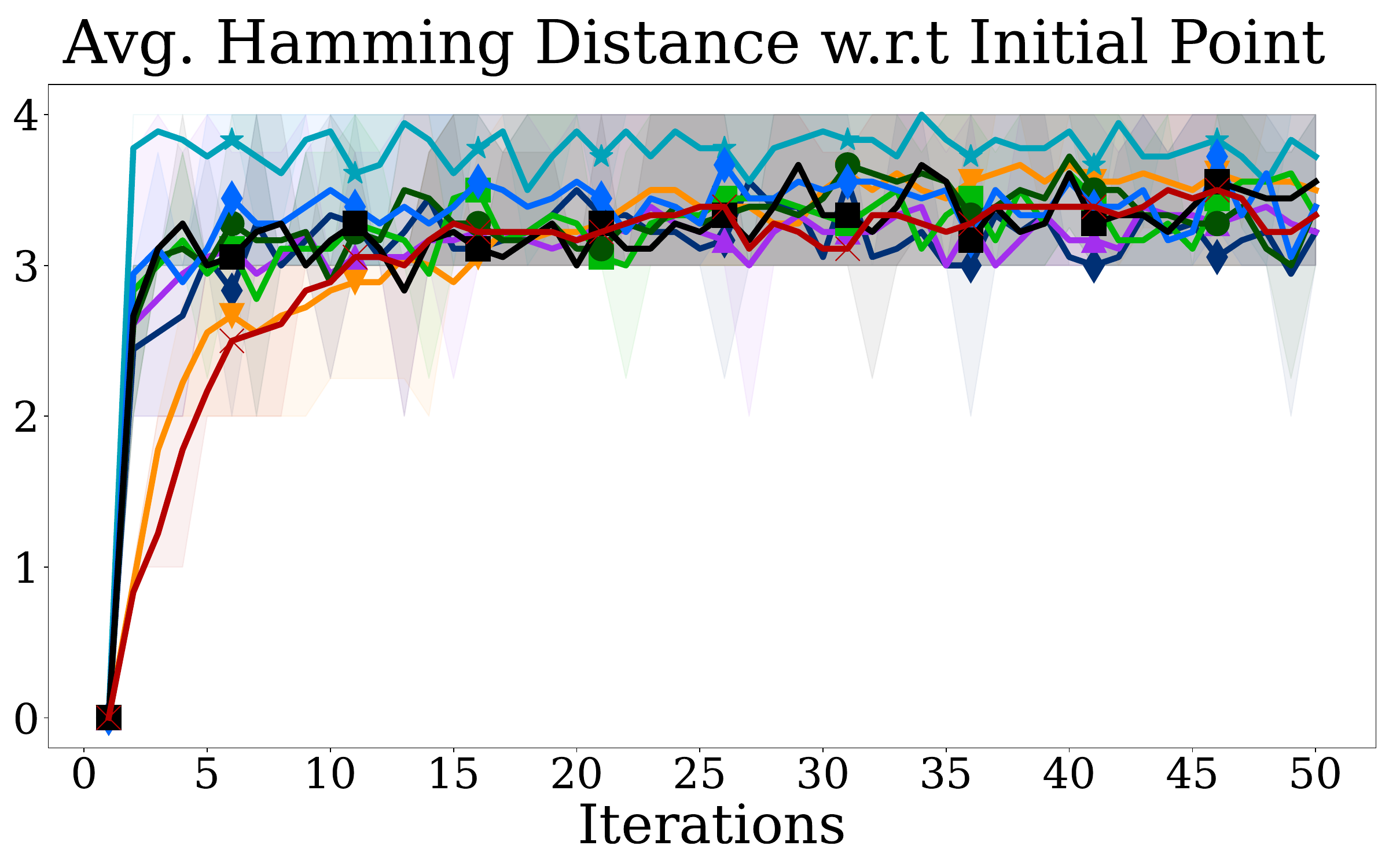}}
    \caption{\emph{GB1(4)}, $n=4$, $18$ reps.}
    \label{fig:avg-perf-plot-hamming-a}
    \end{subfigure}%
    \begin{subfigure}{0.33\textwidth}
    \centering
    {\includegraphics[width=1\linewidth,trim=0 10 0 0,clip]{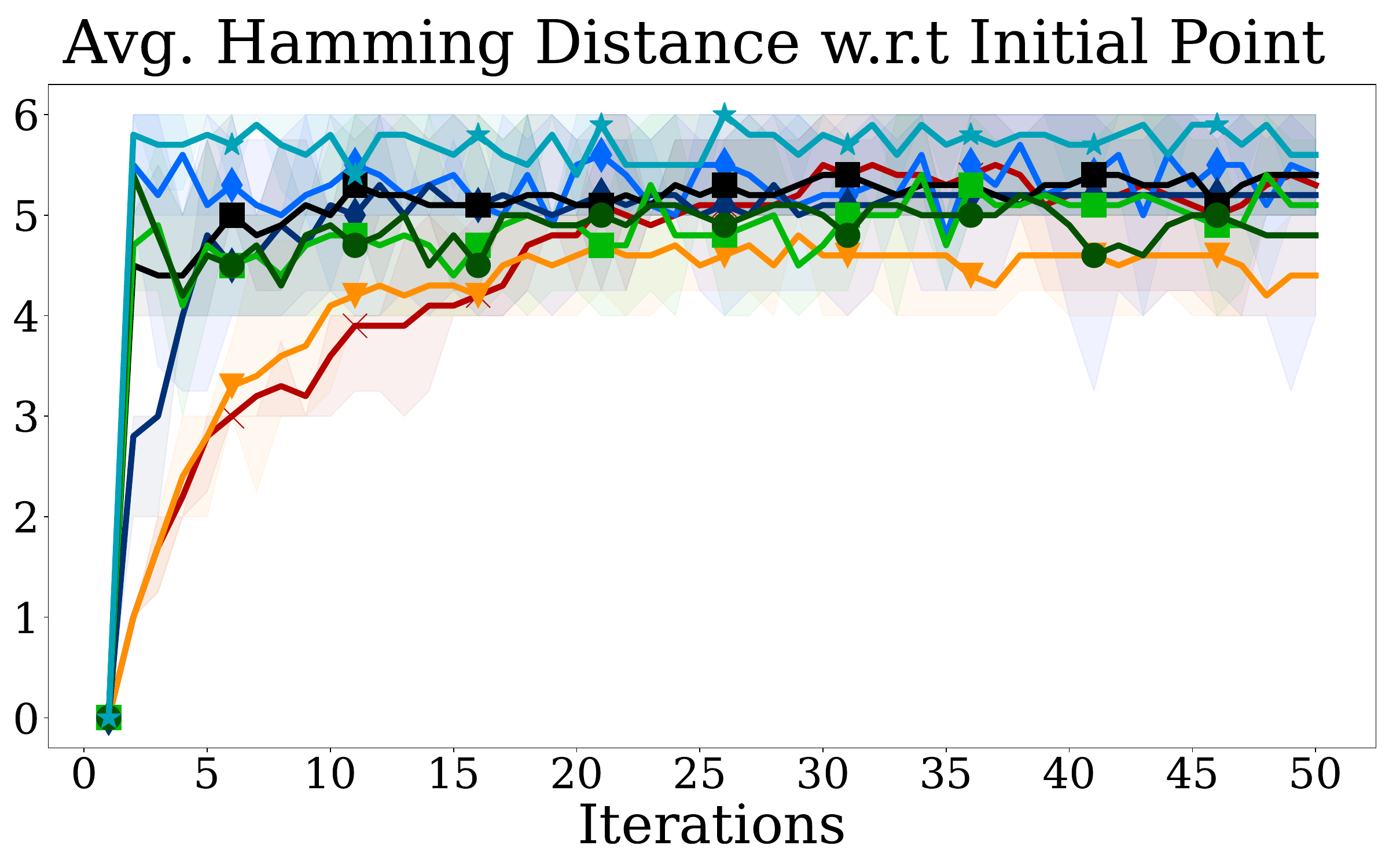}}
    \caption{\emph{GFP}, $n=6$, $10$ reps.}
    \label{fig:avg-perf-plot-hamming-h}
    \end{subfigure}%
    \\
    \begin{subfigure}{0.33\textwidth}
    \centering
    {\includegraphics[width=1\linewidth,trim=0 10 0 0,clip]{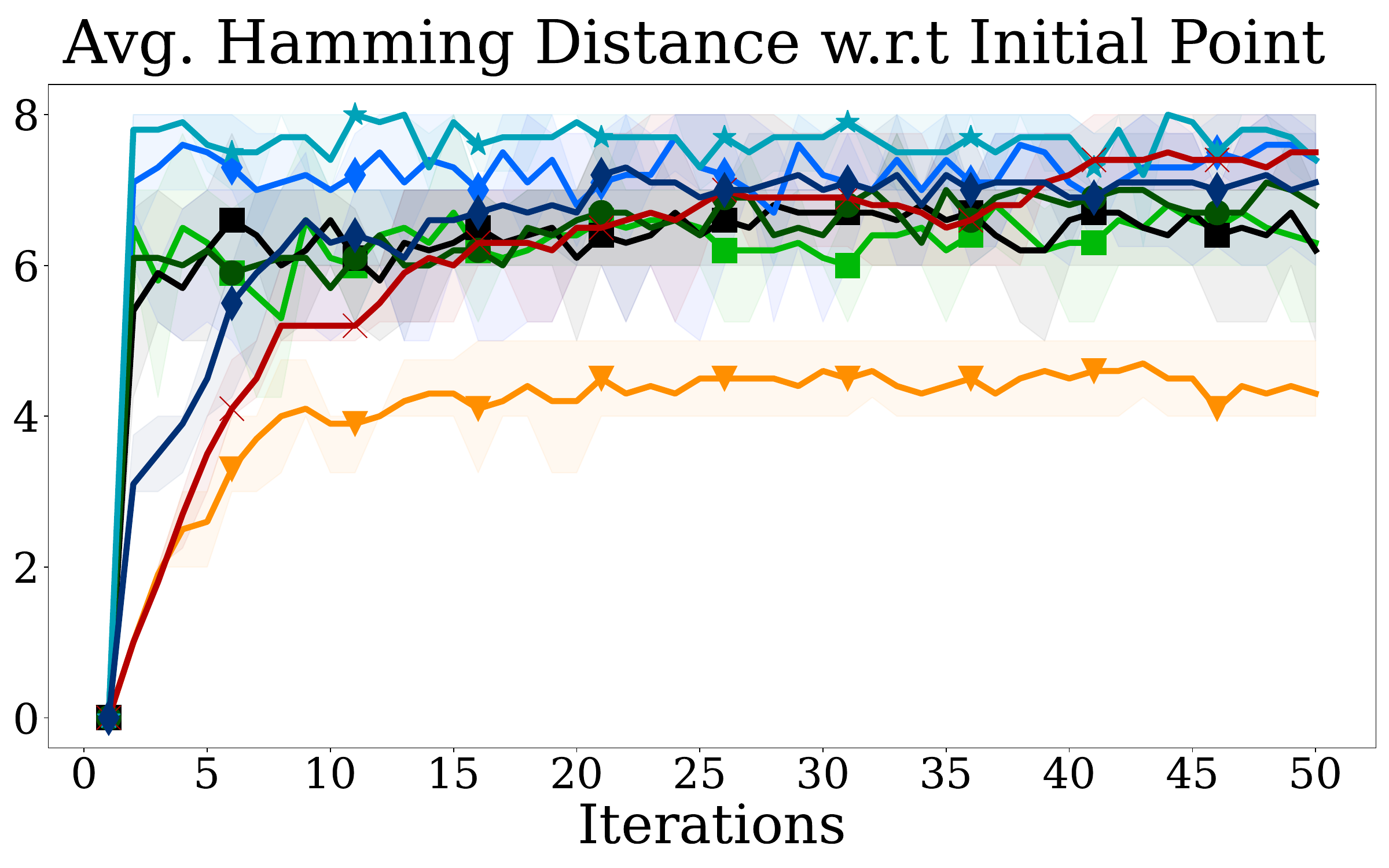}}
    \caption{\emph{GFP}, $n=8$, $10$ reps.}
    \label{fig:avg-perf-plot-hamming-e}
    \end{subfigure}%
        \begin{subfigure}{0.33\textwidth}
    \centering
    {\includegraphics[width=1\linewidth,trim=0 10 0 0,clip]{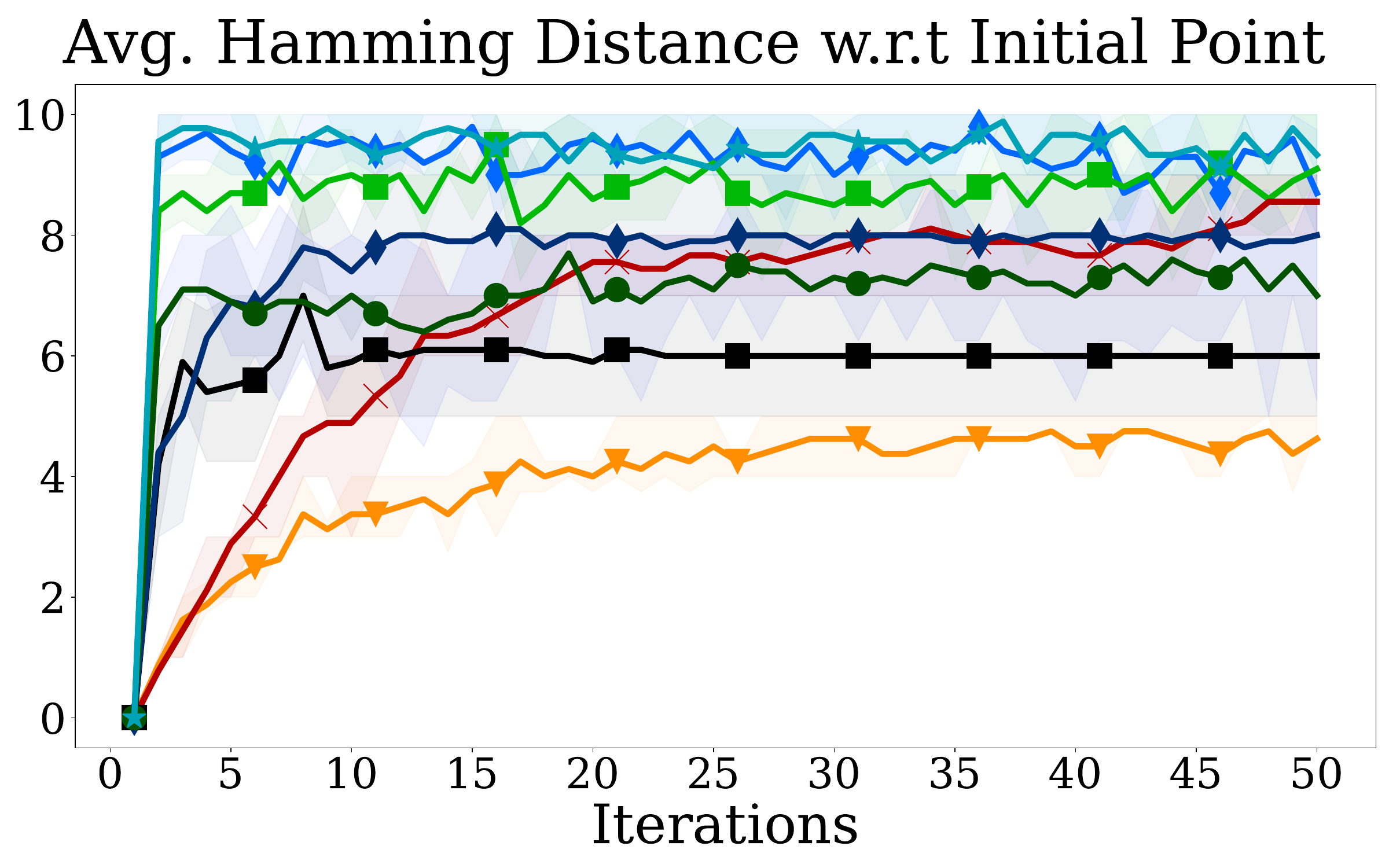}}
    \caption{\emph{GB1(55)}, $n=10$, $10$ reps.}
    \label{fig:avg-perf-plot-hamming-c}
    \end{subfigure}%
        \begin{subfigure}{0.33\textwidth}
    \centering
    {\includegraphics[width=1\linewidth,trim=0 10 0 0,clip]{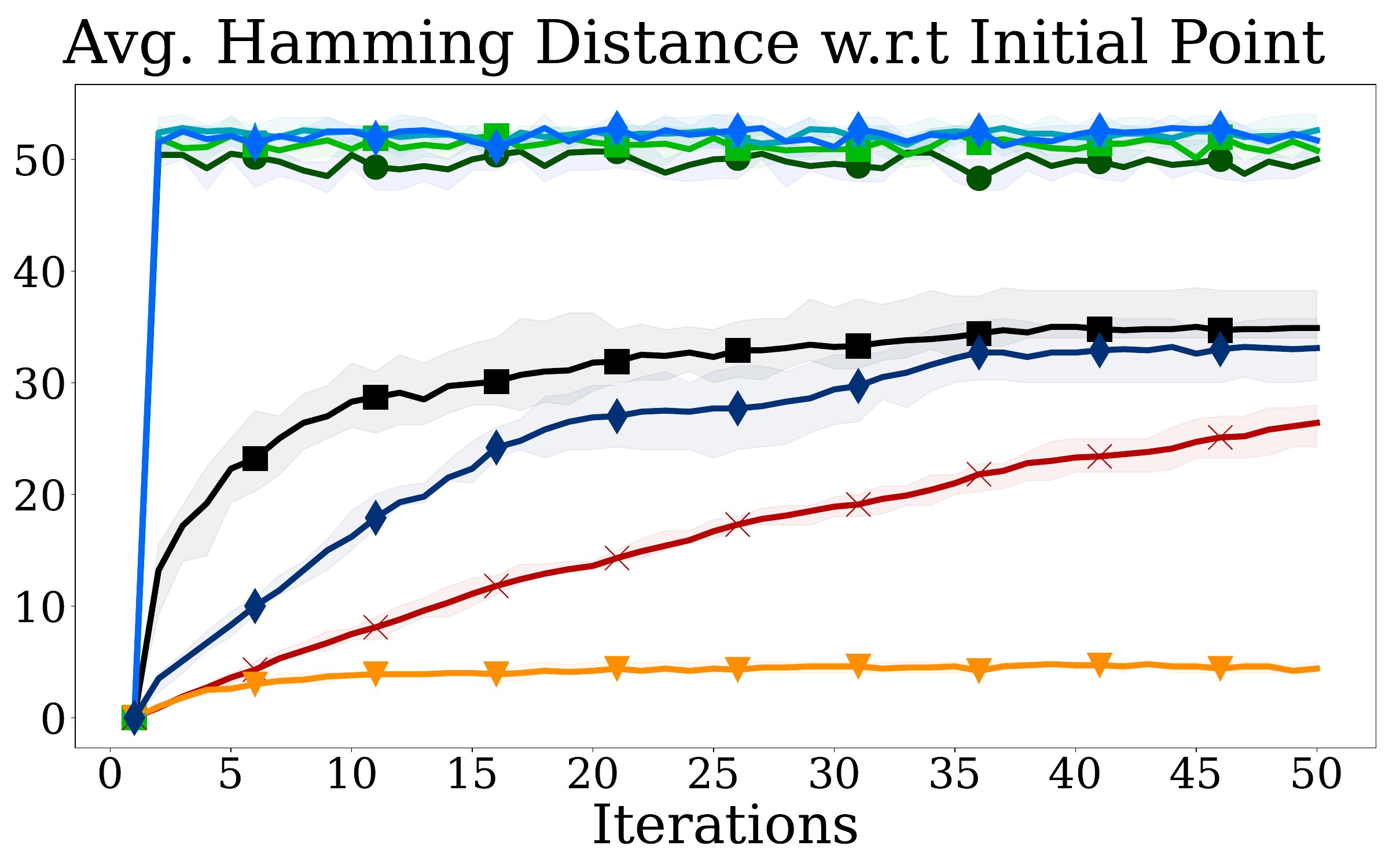}}
    \caption{\emph{GB1(55)}, $n=55$, $10$ reps.}
    \label{fig:avg-perf-plot-hamming-b}
    \end{subfigure}%
    \\
    \begin{subfigure}{1\textwidth}
    \centering
    {\includegraphics[width=1\linewidth,trim=20 5 5 10,clip]{figures/experiment_results/legend_rebuttal_best_2.pdf}}
    \label{fig:subfig1}
    \end{subfigure}%
\setlength{\abovecaptionskip}{-2pt}
 \caption{
 Performance results for the sampled batch diversity w.r.t past measured via mean Hamming distance between the executed variant and the closest initial point from the training set, under batch size $B=5$. Each point on each line is the average of multiple replications initiated with different training sets having $100$ variants for \emph{GB1(4)} \& \emph{Halogenase} and $1000$ for \emph{GB1(55)} \& \emph{GFP}. Similarly, error bars are interquartile ranges averaged over replications.
 In all experiments, \textsc{GameOpt} explores  faster than the baseline methods, except for \textsc{Random} and \textsc{DLS}.}
\label{fig:avg-perf-plot-hamming-rebuttal}
\end{figure*}

Lastly, the mean Hamming distance between the proposed variant and the closest initial point from the training set (see Figure~\ref{fig:avg-perf-plot-hamming-rebuttal}) showcases that \textsc{GameOpt} explores the solution space faster than the baseline methods, except for \textsc{Random} and \textsc{DLS}.

\begin{table}[h]
\caption{Fraction of global optima found for the \emph{GB1(4)} dataset. Each entry is the average of $18$ replications. \textsc{GameOpt} variations are able to sample global optimum sequence (AHCA) more frequently compared to other baselines. Entries of the outperforming methods are denoted in bold. The results show that \textsc{GameOpt-IBR} converges to the best strategy more frequently compared to the baselines.}
\centering
\resizebox{0.6\linewidth}{!}{
\begin{tabular}{l|c}
\specialrule{1.5pt}{1pt}{1pt} 
\textbf{Method} &  \% \textbf{best strategy (AHCA) found} \\
\specialrule{1.5pt}{1pt}{1pt} 
\textsc{GameOpt-Ibr} & $\boldsymbol{33.33}$ \\
\textsc{GameOpt-Hedge} & $16.67$ \\
\textsc{IBR-Fitness} & $16.67$ \\
\textsc{PR} & $0.00$ \\
\textsc{GP-UCB} & $0.00$\\
\textsc{Random} & $0.00$ \\
\textsc{DLS} & $0.00$ \\
\textsc{DLS\_Best} & $11.11$ \\
\textsc{GameOpt-Ibr\_Best} & $11.11$ \\
\specialrule{1.5pt}{1pt}{1pt} 
\end{tabular}
}
\label{tab:fraction-global-optima}
\end{table}

\begin{table}[h]
\caption{Percentage of computed candidates that are above a threshold of $f_{\tau}=0.8* $(maximum fitness value). Each entry is the average of $18$ replications for \emph{Halogenase} and \emph{GB1(4)} settings, and $10$ replications for the others. Entries of the outperforming methods are denoted in bold. Results indicate the effectiveness of \textsc{GameOpt} variations on sampling more optima than the baseline methods.}
\centering
\resizebox{1\linewidth}{!}{
\begin{tabular}{l|c|c|c|c|c|c}
\specialrule{1.5pt}{1pt}{1pt} 
\multirow{2}{*}{\textbf{Method}}       & \multicolumn{6}{c}{\textbf{Domain}} \\
\cline{2-7}
&  Halogenase & GB1(4) & GFP, $n=6$ & GFP, $n=8$ & GB1(55), $n=10$ & GB1(55), $n=55$\\
\specialrule{1.5pt}{1pt}{1pt} 
\textsc{GameOpt-Ibr} &  $\boldsymbol{100.00}$ & $\boldsymbol{21.20}$ & $\boldsymbol{28.52}$ & 50.96  & \textbf{11.60} & 1.96 \\
\textsc{GameOpt-Hedge} & 94.44 & 14.62 & 26.68 & $\boldsymbol{61.72}$ & 9.48 & 2.12 \\
\textsc{IBR-Fitness} & 72.22 & 7.46 & 18.96 & 35.32 & 0.36 & 0.00 \\
\textsc{PR} & 38.89 & 2.05 & 15.96 & 15.32 & 0.00 & 0.04 \\
\textsc{GP-UCB} & 72.22 & 4.82 & - & - & - & -\\
\textsc{Random} & 0.00 & 0.29 & 0.76 & 1.60 & 0.00 & 0.00 \\
\textsc{DLS} & 94.44 & 7.46 & 22.60 & 36.48  & 0.16 &  0.00\\
\textsc{DLS\_Best} & 88.89 & 19.88 & 16.80 & 45.92  & 2.64 & \textbf{52.24} \\
\textsc{GameOpt-Ibr\_Best} & 44.44 & 17.40 & 12.32 & 42.64 & 1.68 & 48.64\\
\specialrule{1.5pt}{1pt}{1pt} 
\end{tabular}
}
\label{tab:fraction-optima-thr}
\end{table}

\begin{table}[h]
\caption{Cumulative maximum for sequences proposed at the end of the BO iterations. Each entry is the average of $18$ replications for \emph{Halogenase} and \emph{GB1(4)} settings, and $10$ replications for the others. Entries of the outperforming methods are denoted in bold. \textsc{GameOpt} variations show superior performance in proposing higher fitness-valued protein variants. }
\centering
\resizebox{1\linewidth}{!}{
\begin{tabular}{l|c|c|c|c|c|c}
\specialrule{1.5pt}{1pt}{1pt} 
\multirow{2}{*}{\textbf{Method}}       & \multicolumn{6}{c}{\textbf{Domain}} \\
\cline{2-7}
 &  Halogenase & GB1(4) & GFP, $n=6$ & GFP, $n=8$ & GB1(55), $n=10$ &  GB1(55), $n=55$ \\
\specialrule{1.5pt}{1pt}{1pt} 
\textsc{GameOpt-Ibr} & \textbf{3.51} & 0.93 & \textbf{6.17} & \textbf{6.37} & 1.94 & 4.72 \\
\textsc{GameOpt-Hedge} & 3.45 & \textbf{0.94} & 6.15 &  6.34 & 1.69 & 4.66 \\
\textsc{IBR-Fitness} & 3.10 & 0.86 & 6.08 & 6.21 & 0.97 & 2.99 \\
\textsc{PR} & 2.08 & 0.81 & 6.06 & 6.07 & 0.90 & 1.18 \\
\textsc{GP-UCB} & 2.98 & 0.82 & - & - & - & -\\
\textsc{Random} & 1.15 & 0.68 & 5.92 & 5.97 & -0.08 & -0.66\\
\textsc{DLS} & \textbf{3.51} & 0.89 & 6.09 & 6.14 & 1.70 & 1.35 \\
\textsc{DLS\_Best} & 3.40 & 0.93 & 6.15 & 6.33 & \textbf{2.11} & \textbf{7.87} \\
\textsc{GameOpt-IBR\_Best} & 2.28 & 0.93 & 6.14 & 6.33 & 1.91 &  7.25 \\
\specialrule{1.5pt}{1pt}{1pt} 
\end{tabular}
}
\label{tab:cumulative-maximum}
\end{table}

\begin{table}[h]
\caption{Mean pairwise Hamming distance between the sequences proposed at each iteration. Each entry is the average of $18$ replications for \emph{Halogenase} and \emph{GB1(4)} settings, and $10$ replications for the others. 
In all problem domains, \textsc{Random} baseline consistently samples a rather diverse batch of evaluation points w.r.t. past proposed variants. Although this shows enhanced exploration, one drawback is the lack of exploitation. Hence, as illustrated in Figure~\ref{fig:avg-perf-plot-hamming-2}, \textsc{GameOpt} balances these two successfully and shows a moderate sampled batch diversity.
 }
\centering
\resizebox{1\linewidth}{!}{
\begin{tabular}{l|c|c|c|c|c|c}
\specialrule{1.5pt}{1pt}{1pt} 
\multirow{2}{*}{\textbf{Method}}       & \multicolumn{6}{c}{\textbf{Domain}} \\
\cline{2-7}
 &  Halogenase & GB1(4) & GFP, $n=6$ & GFP, $n=8$ & GB1(55), $n=10$ & GB1(55), $n=55$ \\
\specialrule{1.5pt}{1pt}{1pt} 
\textsc{GameOpt-Ibr} & 1.40 & 2.94 & 3.08 & 4.11 & 7.32 & 50.93\\
\textsc{GameOpt-Hedge} & 2.13 & 2.91 & 2.93 & 4.23 & 3.98 & 45.89 \\
\textsc{IBR-Fitness} & 0.86 & 0.88 & 0.85 & 0.89 & 0.86 & 0.89\\
\textsc{PR} & 1.61 & 1.57 & 1.61 & 1.67 & 1.69 & 1.60\\
\textsc{GP-UCB} & 0.77 & 2.41 & - & - & - & - \\
\textsc{Random} & 2.85 & 3.8 & 5.69 &  7.64 & 9.51 & 52.34\\
\textsc{DLS} & 1.13 & 3.36 & 5.13 &  7.13 & 9.38 & 52.19\\
\textsc{DLS\_Best} & 0.68 & 2.56 & 1.18 &  2.36 & 1.55 & 3.57\\
\textsc{GameOpt-IBR\_Best} & 0.33 & 2.64 & 1.48 & 2.78 & 0.57 & 3.12 \\
\specialrule{1.5pt}{1pt}{1pt} 
\end{tabular}
}
\label{tab:mean-hamming-distance}
\end{table}

\

\clearpage

\subsection{Exploring players' grouping}
\label{app:grouping-players}
Until this juncture, we have exclusively examined scenarios where a \textsc{GameOpt} player is responsible for only a single site within the protein sequence.
In light of the \textit{epistasis} \citep{epistasis} phenomenon in protein design, which underscores how the effect of a mutation on fitness can be influenced by the presence of other mutations within the same protein, we now explore the concept of grouping protein sites together, \textit{i.e.}, having players being responsible for more than one site.  
This is because modeling protein sites independently may yield different fitness outcomes than finding equilibria among groups of several sites. To this end, we conduct a preliminary investigation into whether this phenomenon alters \textsc{GameOpt}'s performance.

We experiment on \emph{GB1(4)} with $\{0,1,2,3\}$ protein sites and $\mathcal{N}=\{1,2\}$ players, considering 3 possible player \& site groupings: $\{(01,23), (02,13), (03,12)\}$. For instance, setting $(01,23)$ means that the first player is responsible for sites $\{0,1\}$ and the other one for $\{2,3\}$. 

\begin{figure*}[ht]
    \centering  
    \begin{subfigure}{0.33\textwidth}
    \centering
    {\includegraphics[width=1\linewidth,trim=0 10 0 0,clip]{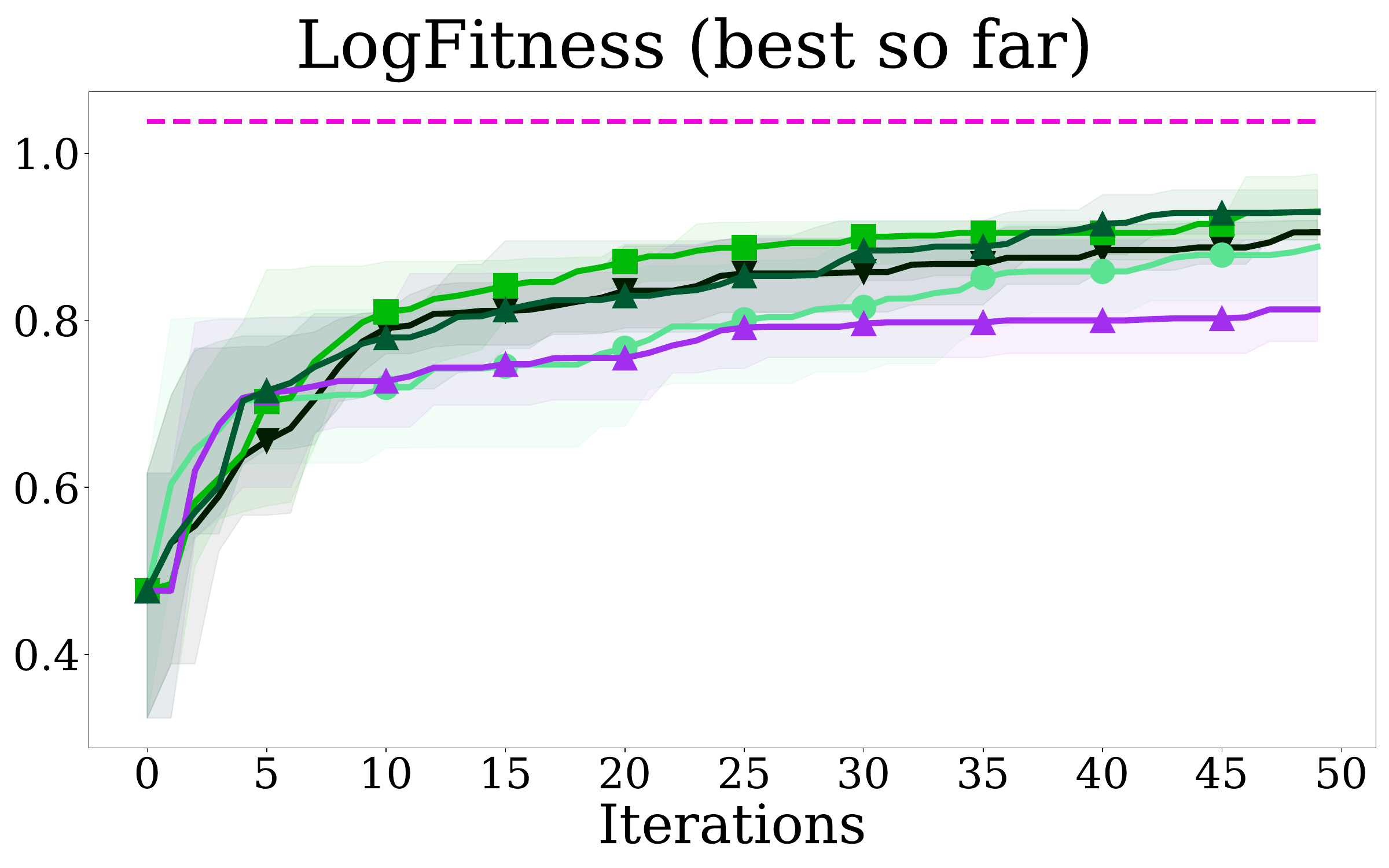}}
    \caption{Convergence speed}
    \label{fig:player-grouping-a}
    \end{subfigure}%
    \begin{subfigure}{0.33\textwidth}
    \centering
    {\includegraphics[width=1\linewidth,trim=0 10 0 0,clip]{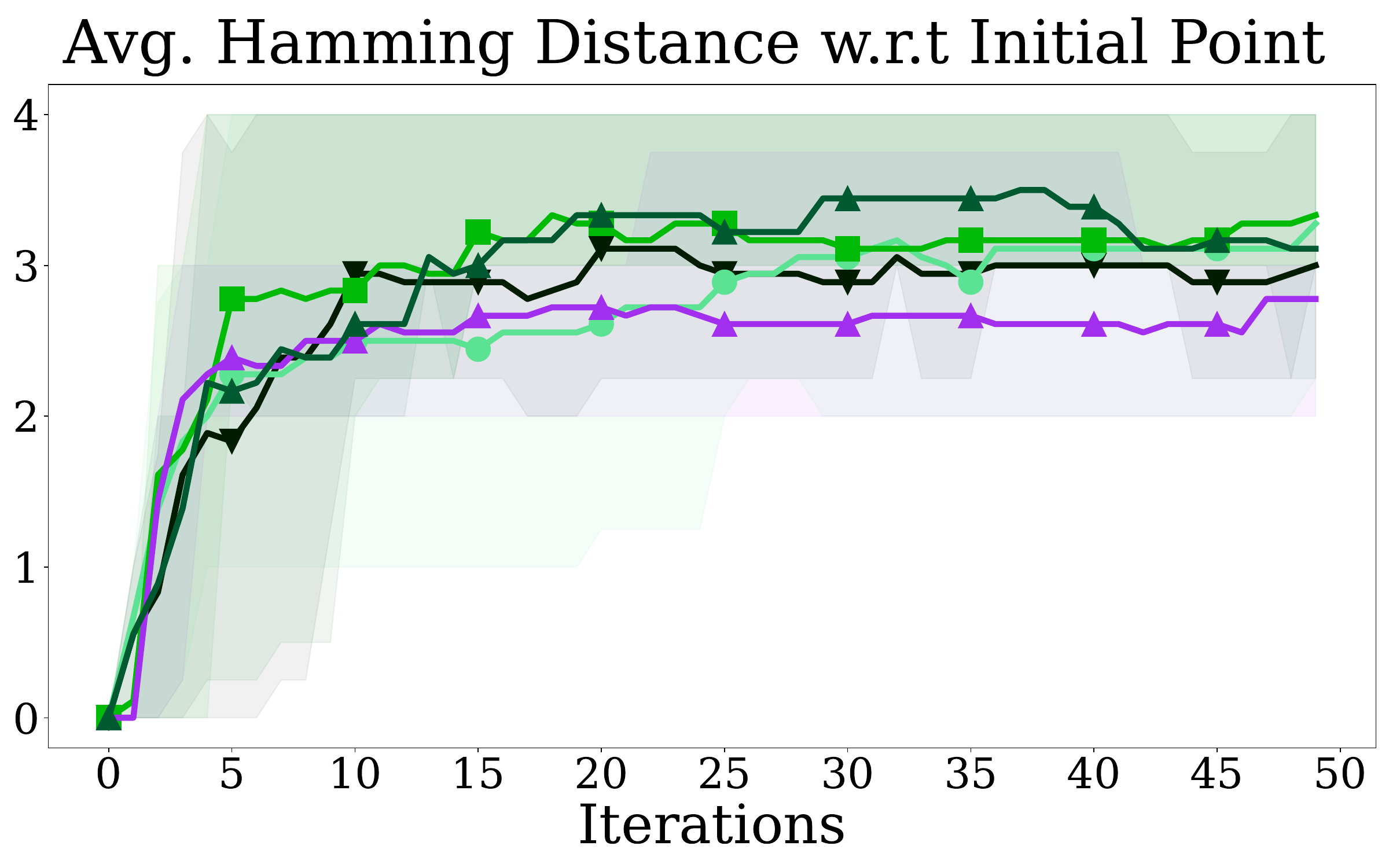}}
    \caption{Sample diversity w.r.t $(1)$}
    \label{fig:player-grouping-b}
    \end{subfigure}%
    \begin{subfigure}{0.33\textwidth}
    \centering
    {\includegraphics[width=1\linewidth,trim=0 10 0 0,clip]{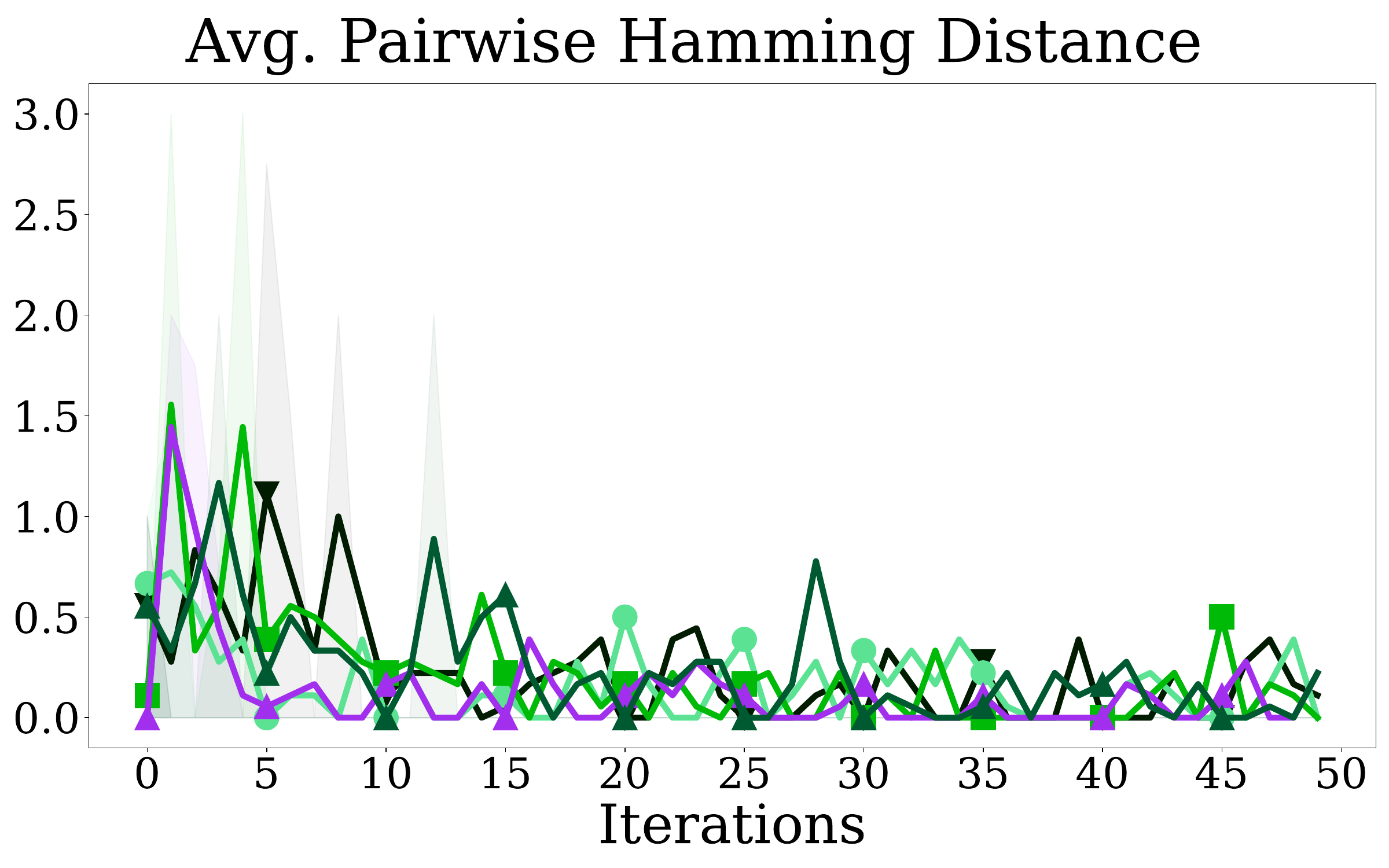}}
    \caption{Sample diversity w.r.t $(2)$}
    \label{fig:player-grouping-c}
    \end{subfigure}%
    \\
    \begin{subfigure}{1\textwidth}
    \centering
    {\includegraphics[width=1\linewidth,trim=30 20 10 15,clip]{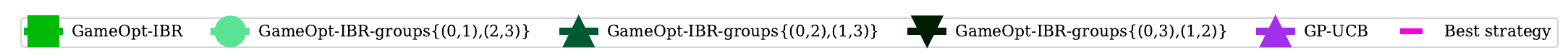}}
    \label{fig:subfig1}
    \end{subfigure}%
\setlength{\abovecaptionskip}{-5pt}
 \caption{\textsc{GameOpt-Ibr} performance for player grouping, under \emph{GB1(4)} setting, $18$ reps. There is no significant performance difference between individual players and player grouping settings under this domain.}
\label{fig:player-grouping-ibr}
\end{figure*}

\begin{figure*}[ht]
    \centering  
    \begin{subfigure}{0.33\textwidth}
    \centering
    {\includegraphics[width=1\linewidth,trim=0 10 0 0,clip]{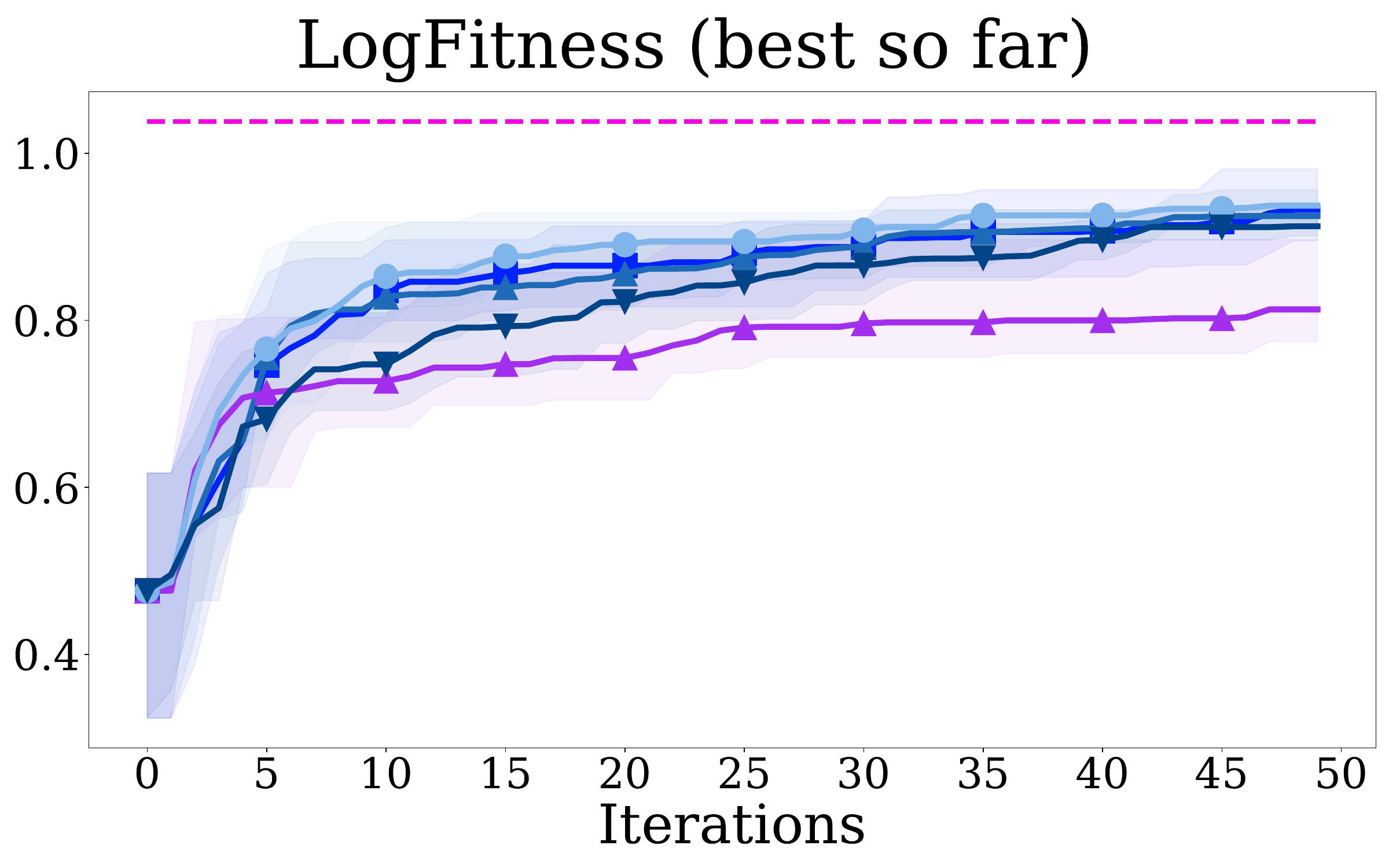}}
    \caption{Convergence speed}
    \label{fig:player-grouping-cce-a}
    \end{subfigure}%
    \begin{subfigure}{0.33\textwidth}
    \centering
    {\includegraphics[width=1\linewidth,trim=0 10 0 0,clip]{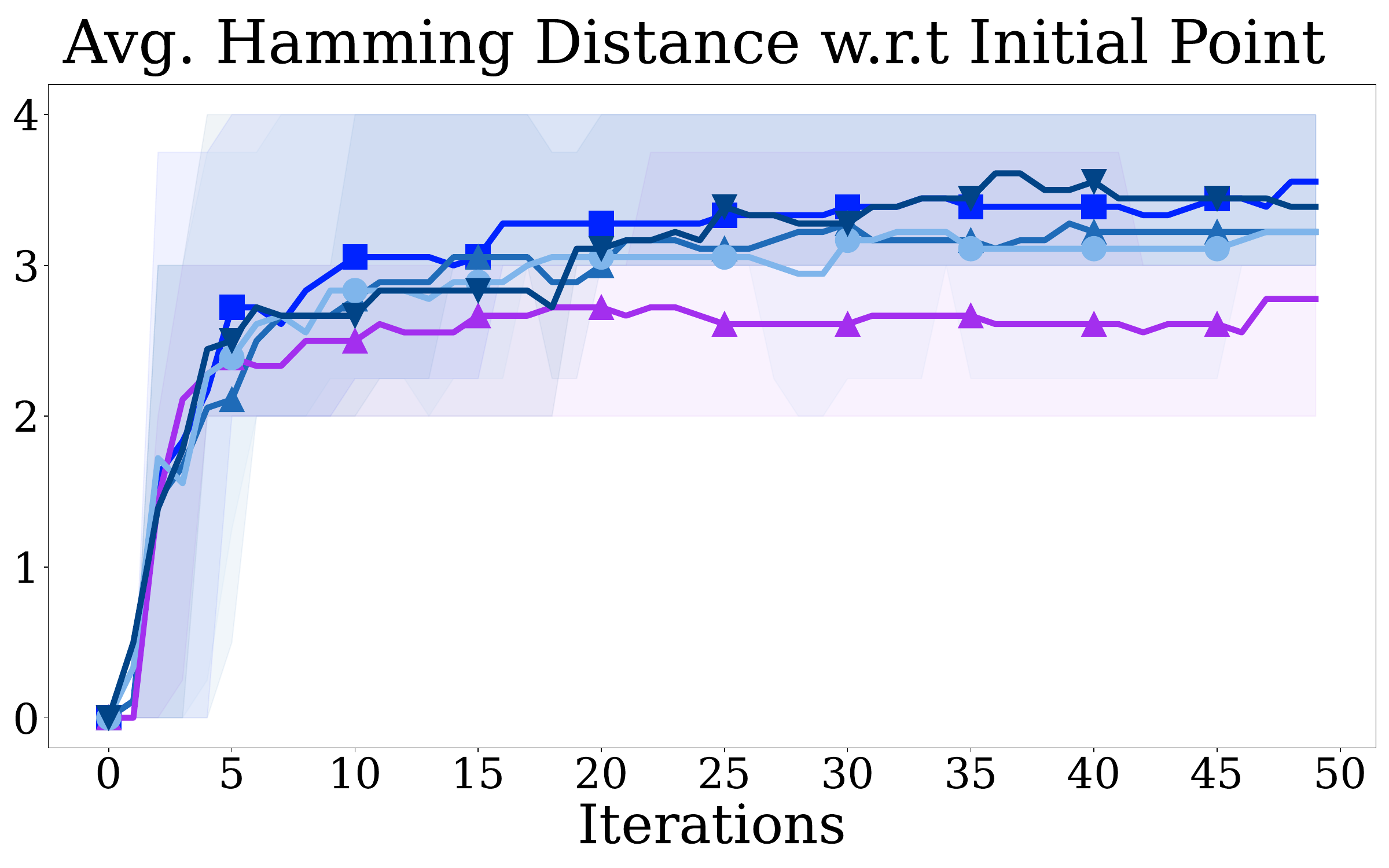}}
    \caption{Sample diversity w.r.t $(1)$}
    \label{fig:player-grouping-cce-b}
    \end{subfigure}%
    \begin{subfigure}{0.33\textwidth}
    \centering
    {\includegraphics[width=1\linewidth,trim=0 10 0 0,clip]{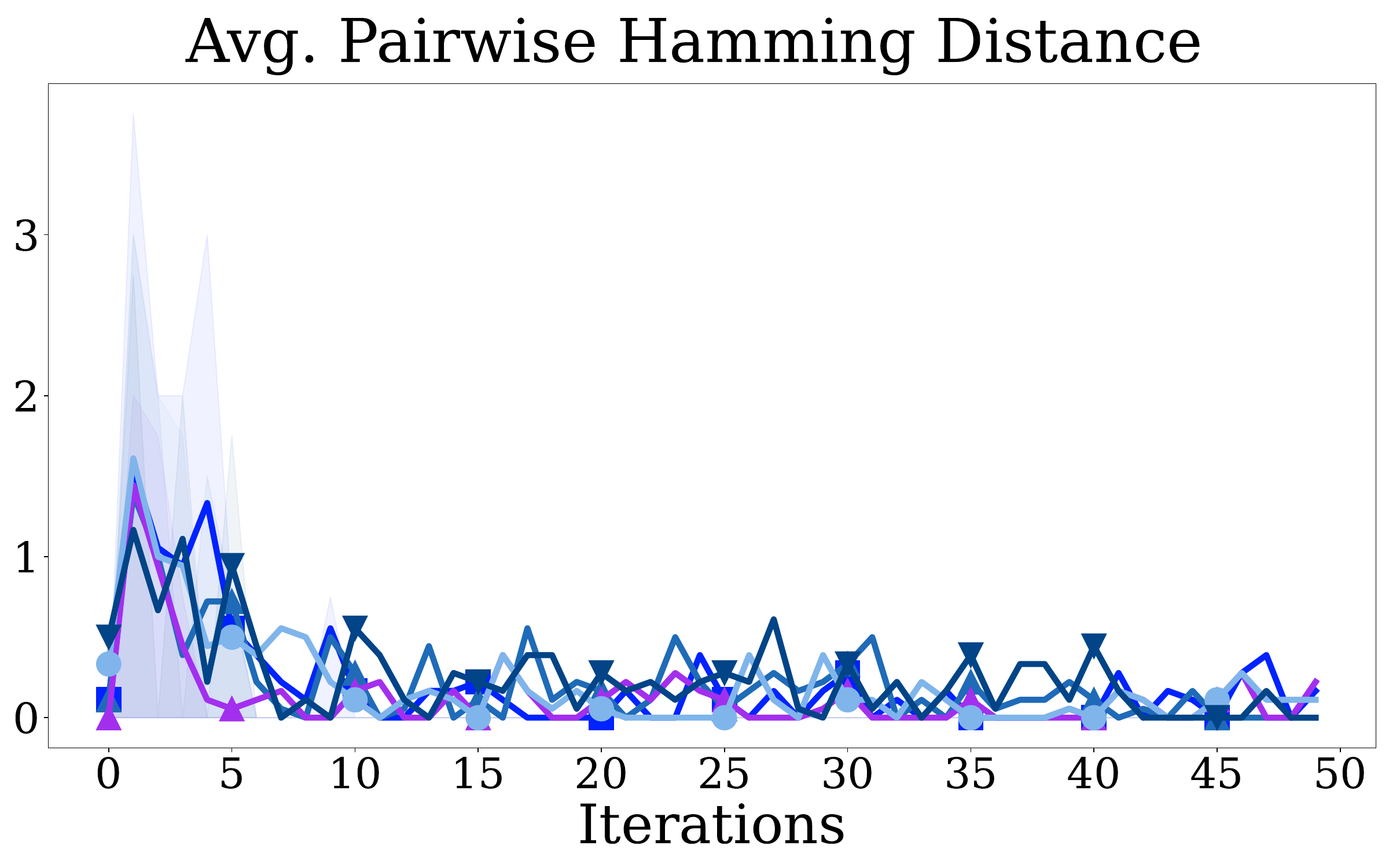}}
    \caption{Sample diversity w.r.t $(2)$}
    \label{fig:player-grouping-cce-c}
    \end{subfigure}%
    \\
    \begin{subfigure}{1\textwidth}
    \centering
    {\includegraphics[width=1\linewidth,trim=30 20 10 15,clip]{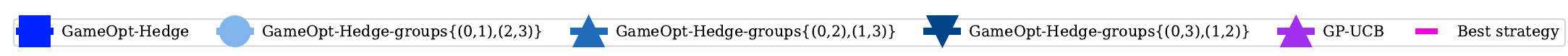}}
    \label{fig:subfig1}
    \end{subfigure}%
\setlength{\abovecaptionskip}{-5pt}
 \caption{\textsc{GameOpt-Hedge} performance for player grouping, under \emph{GB1(4)} setting, $18$ reps. No significant performance difference exists between individual players and player grouping settings under this domain.}
\label{fig:player-grouping-cce}
\end{figure*}

Our evaluations with \textsc{GameOpt-Ibr} and \textsc{GameOpt-Hedge} using the same performance measures (Figures~\ref{fig:player-grouping-ibr} and ~\ref{fig:player-grouping-cce}) showed that there is no significant performance difference between individual players and grouping settings as they all discover the high log fitness valued protein variants at a similar rate while collecting batches of diverse evaluation points. Nevertheless, an in-depth examination of this phenomenon on larger datasets remains a subject for future investigation.

\subsection{Exploring players' with different action set sizes (dimensions)}
\label{app:grouping-players-different-dimensions}

In our main experiments presented in Section~\ref{sec:results}, as well as the grouping of players explored in Appendix~\ref{app:grouping-players}, we exclusively examined scenarios where \textsc{GameOpt} players have an equal number of actions, i.e. $\mid \mathcal{X}^{(i)} \mid$ are same for all $i$. In the context of protein design, this corresponds to the setting where players are responsible for an equal number of sites.
However, \textsc{GameOpt} can also be applied to the setting where players have \textit{different} numbers of actions, i.e. $\mid \mathcal{X}^{(i)} \mid$ differ among players.

To demonstrate the \textit{generalizability} of \textsc{GameOpt} to such settings, we further experiment on settings with player groupings, where each group is responsible for a different number of protein sites. 
Particularly, we consider site groupings: $\{(0, 12), (1,02), (2,01)\}$ for the \emph{Halogenase} problem domain. Whereas we consider groupings: $\{(0, 123), (1, 023), (2, 013), (3,012)\}$ for the \emph{GB1(4)} domain.
For instance, the setting $(0,12)$ means that the first player is responsible for site $\{0\}$ and the other one for $\{1,2\}$, which makes their action size as $\mathcal{X}^{(i=1)}=20, \mathcal{X}^{(i=2)}=20^2$. 

\begin{figure*}[ht]
    \centering  
    \begin{subfigure}{0.35\textwidth}
    \centering
    {\includegraphics[width=1\linewidth,trim=0 10 0 0,clip]{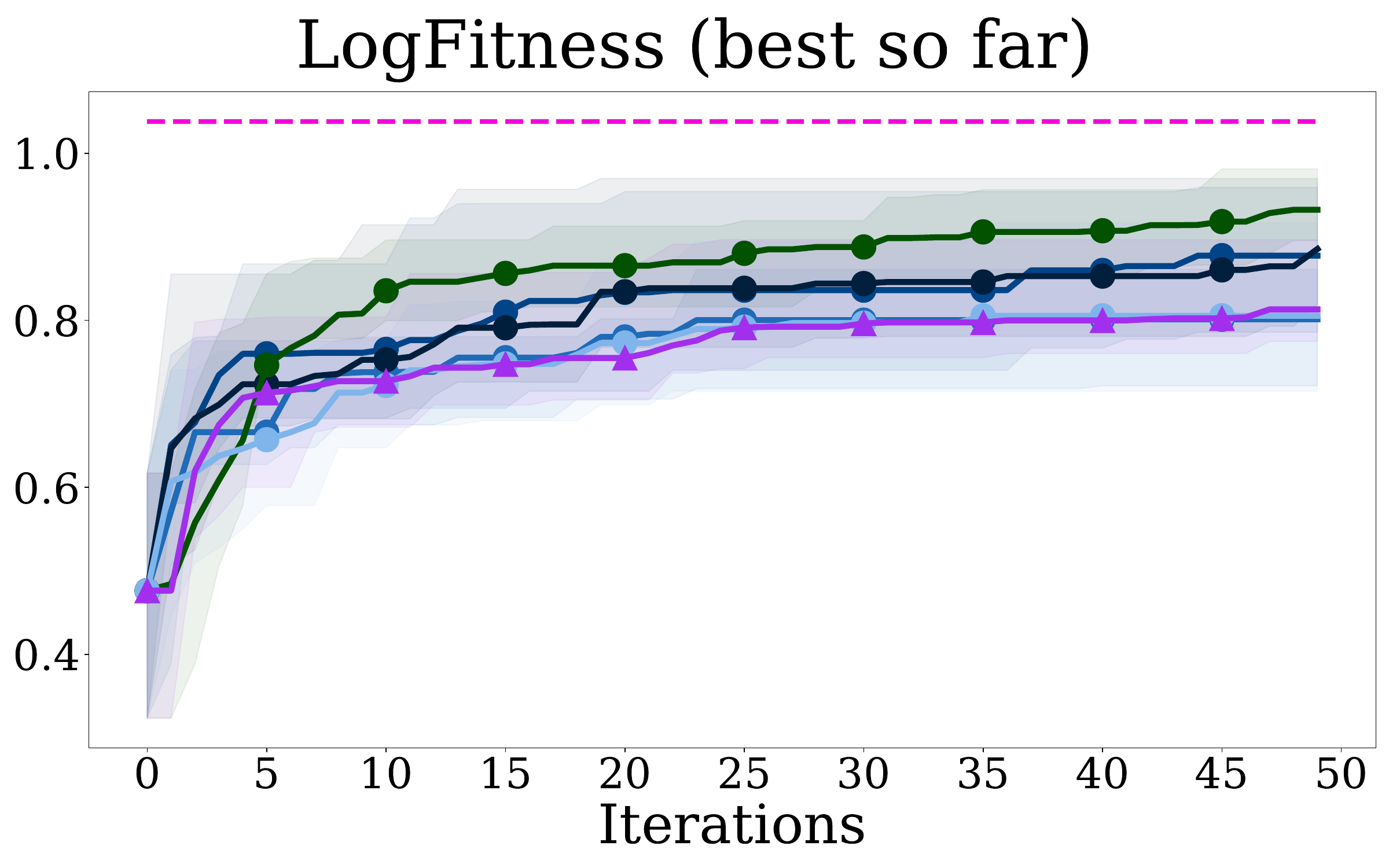}}
    \caption{\textsc{GameOpt-Hedge} on \emph{GB1(4)}.}
    \label{fig:player-grouping-a}
    \end{subfigure}%
        \begin{subfigure}{0.35\textwidth}
    \centering
    {\includegraphics[width=1\linewidth,trim=0 10 0 0,clip]{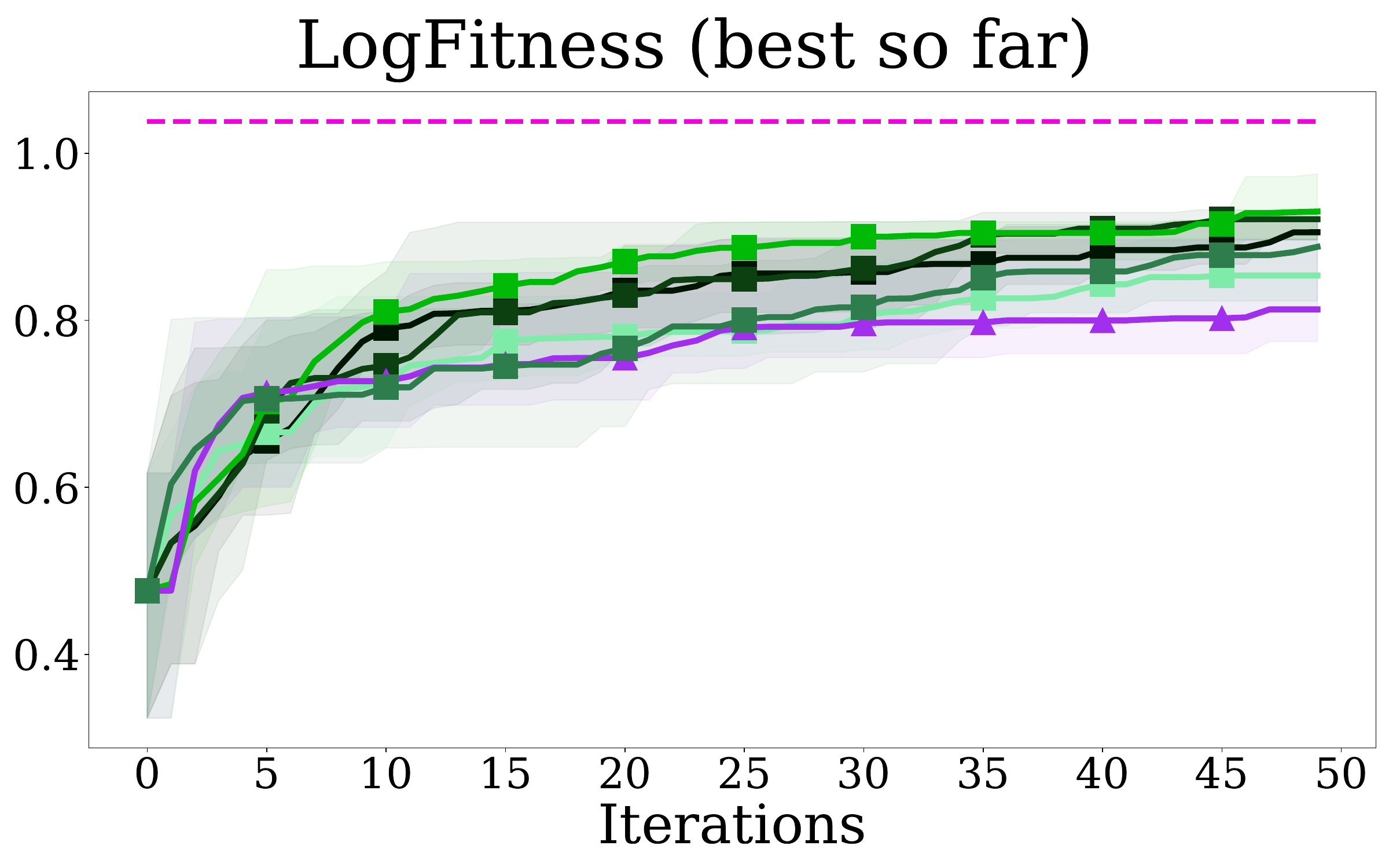}}
    \caption{\textsc{GameOpt-IBR} on \emph{GB1(4)}.}
    \label{fig:player-grouping-a}
    \end{subfigure}%
    \\
    \begin{subfigure}{1\textwidth}
    \centering
    {\includegraphics[width=1\linewidth,trim=30 20 10 15,clip]{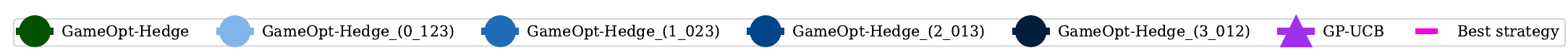}}
    \label{fig:subfig1}
    \end{subfigure}%
    \\
    \begin{subfigure}{1\textwidth}
    \centering
    {\includegraphics[width=1\linewidth,trim=30 20 10 15,clip]{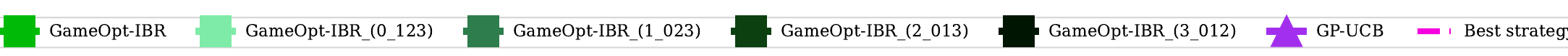}}
    \label{fig:subfig1}
    \end{subfigure}%
    \\
        \begin{subfigure}{0.35\textwidth}
    \centering
    {\includegraphics[width=1\linewidth,trim=0 10 0 0,clip]{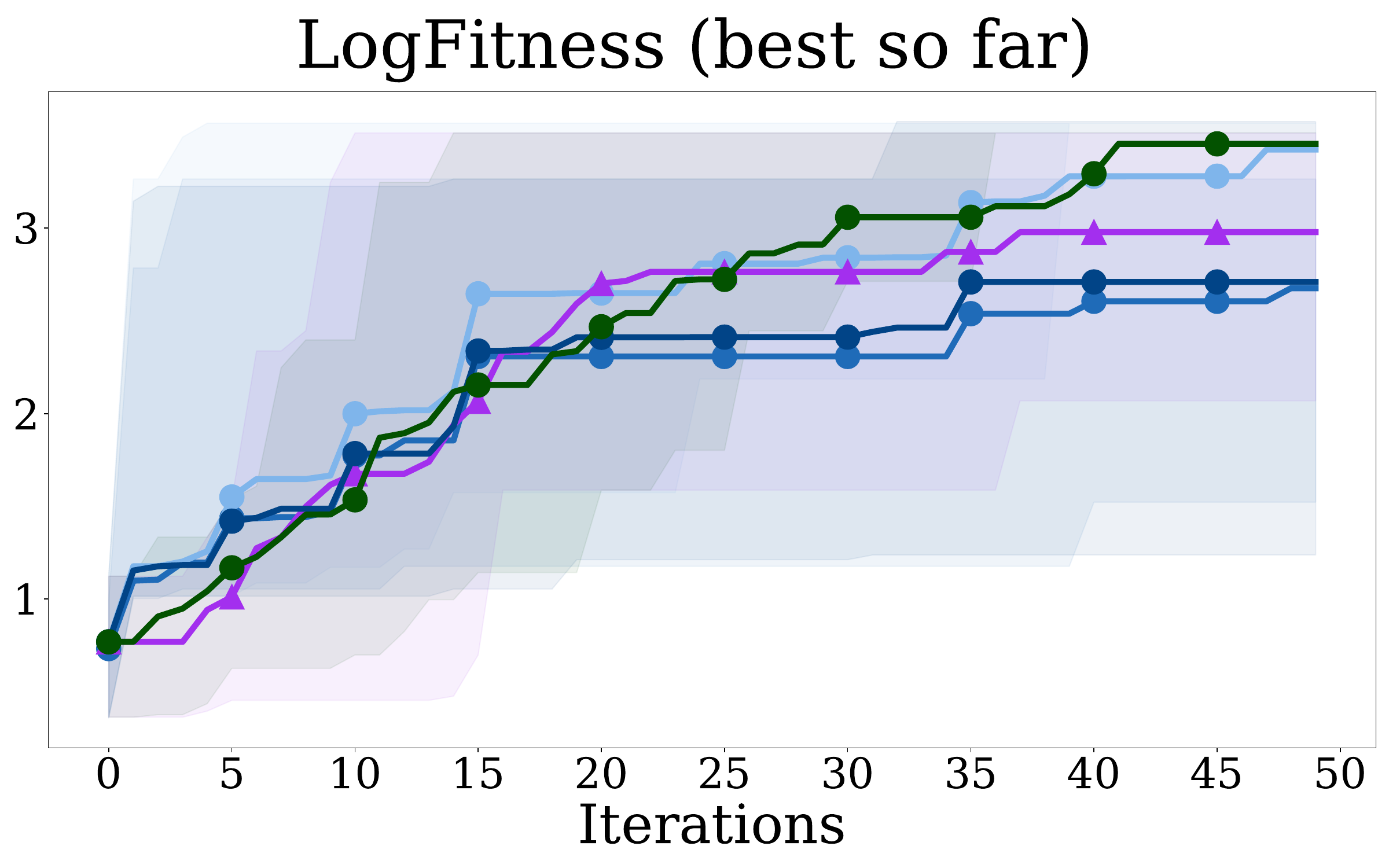}}
    \caption{\textsc{GameOpt-Hedge} on \emph{Halogenase}.}
    \label{fig:player-grouping-a}
    \end{subfigure}%
        \begin{subfigure}{0.35\textwidth}
    \centering
    {\includegraphics[width=1\linewidth,trim=0 10 0 0,clip]{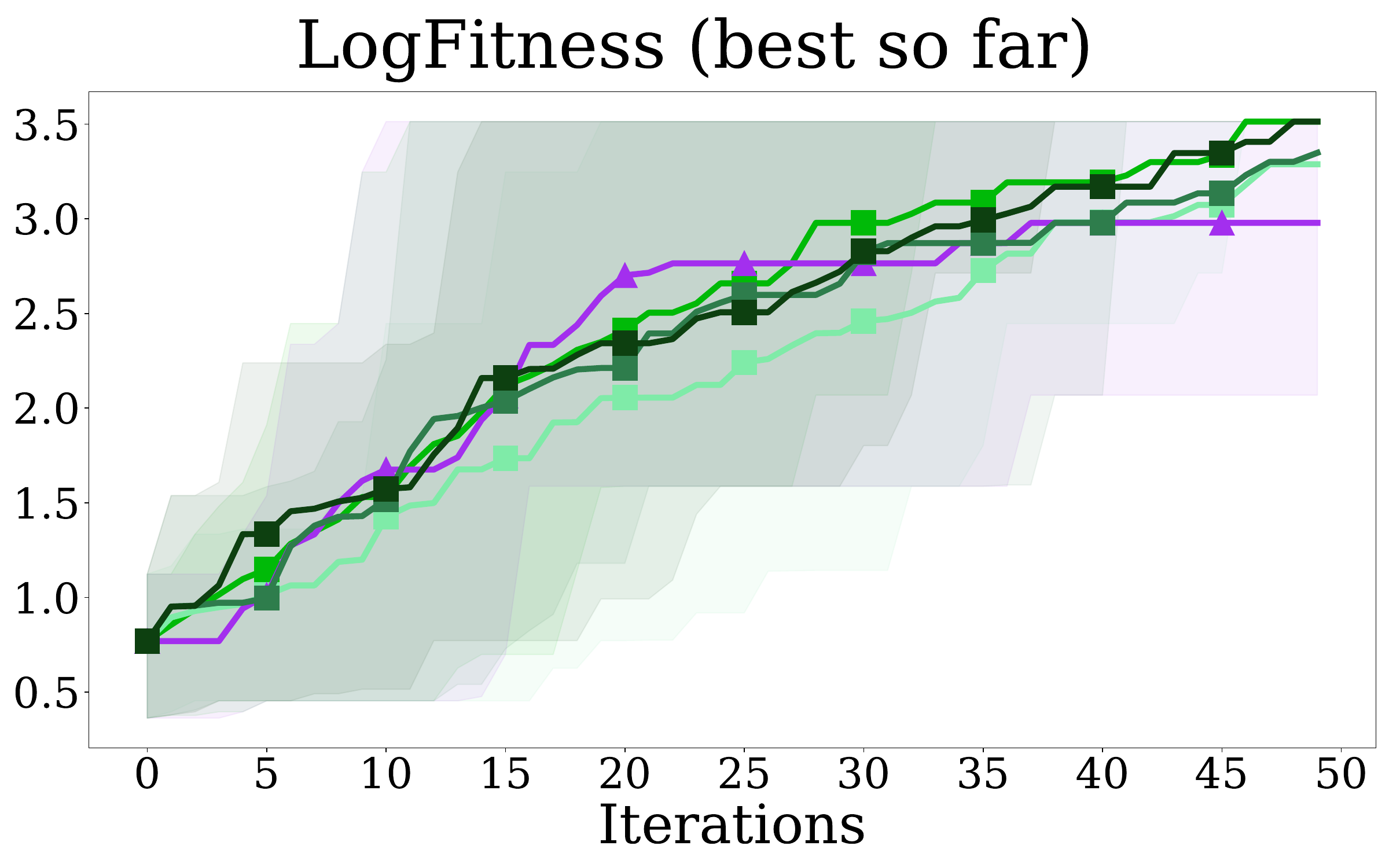}}
    \caption{\textsc{GameOpt-IBR} on \emph{Halogenase}.}
    \label{fig:player-grouping-a}
    \end{subfigure}%
    \\
    \begin{subfigure}{1\textwidth}
    \centering
    {\includegraphics[width=1\linewidth,trim=30 20 10 15,clip]{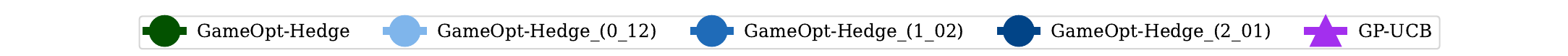}}
    \label{fig:subfig1}
    \end{subfigure}%
    \\
    \begin{subfigure}{1\textwidth}
    \centering
    {\includegraphics[width=1\linewidth,trim=30 20 10 15,clip]{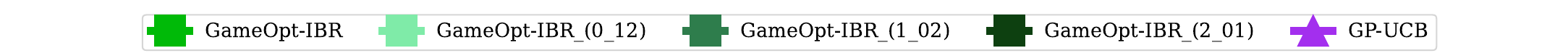}}
    \label{fig:subfig1}
    \end{subfigure}%
\setlength{\abovecaptionskip}{-5pt}
 \caption{\textsc{GameOpt-Hedge} and \textsc{GameOpt-Ibr} performance under player groupings with different action sizes, on \emph{GB1(4)} and \emph{Halogenase} domains, $18$ reps. There is no significant performance difference between individual players and player grouping settings, however, in most of the settings \textsc{GameOpt} groupings outperform the GP-UCB baseline. The results support that \textsc{GameOpt} is also effective under problem settings with varying action (dimension) sizes. }
\label{fig:player-grouping-diff-actionsize}
\end{figure*}

The experiment results presented in Figure~\ref{fig:player-grouping-diff-actionsize} show that there is no significant performance difference between individual players and player grouping settings with varying action set sizes, however, in most of the settings \textsc{GameOpt} groupings outperform GP-UCB baseline. This clearly demonstrates the applicability of \textsc{GameOpt} framework under problem domains with unequal action set sizes.

\subsection{Batch size ablations}
\label{app:batch-size-ablation}
\begin{figure}[!htb]
    \centering  
    \begin{subfigure}{0.33\linewidth}
    \centering
    {\includegraphics[width=1\linewidth,trim=10 10 0 10,clip]{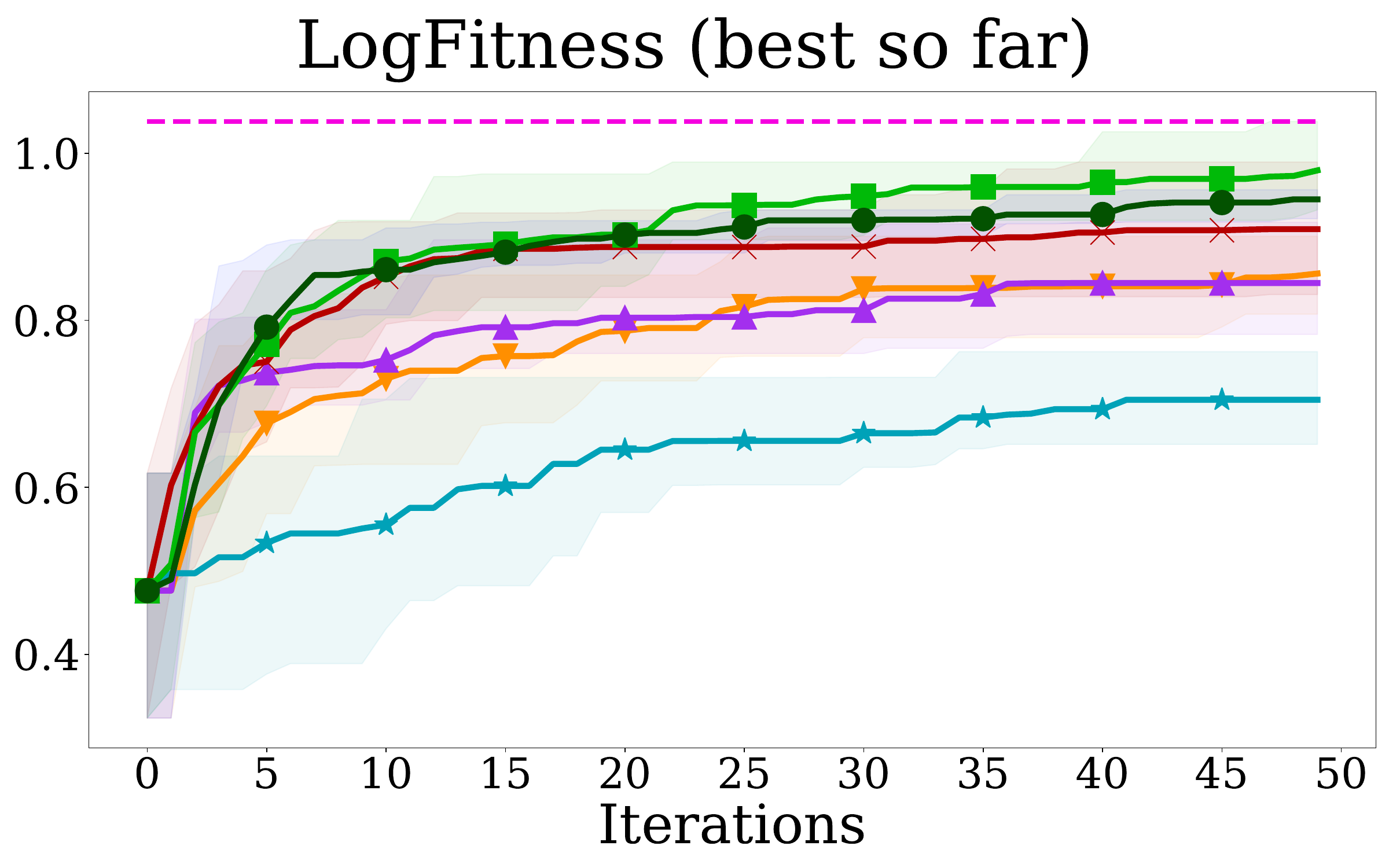}}
    \caption{$B=10$.}
    \label{fig:avg-perf-plot-d}
    \end{subfigure}%
    \begin{subfigure}{0.33\linewidth}
    \centering
    {\includegraphics[width=1\linewidth,trim=10 10 0 10,clip]{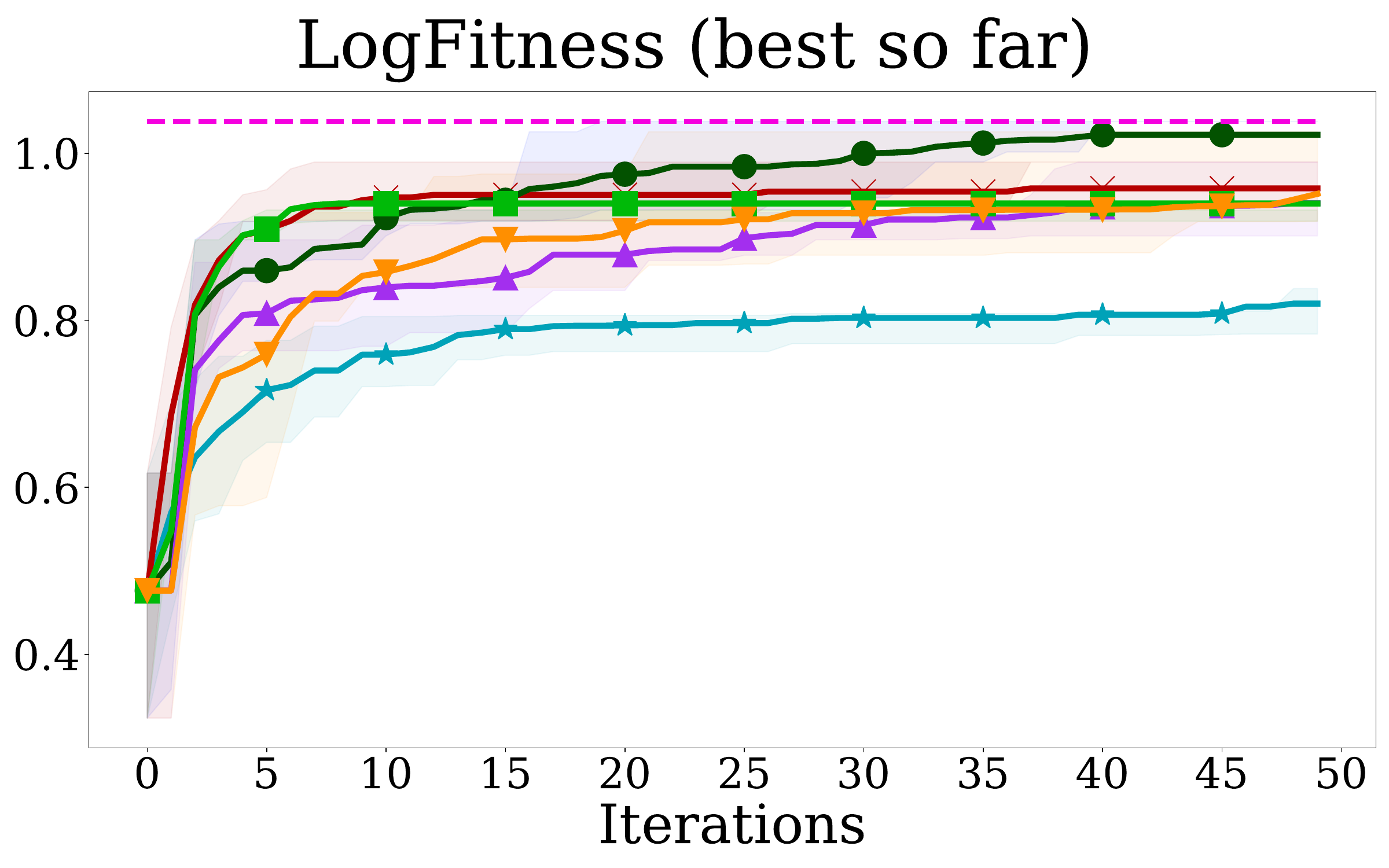}}
    \caption{$B=50$.}
    \label{fig:avg-perf-plot-a}
    \end{subfigure}%
    \\
    \vspace{0.5em}
    \begin{subfigure}{1\linewidth}
    \centering
    {\includegraphics[width=1\linewidth,trim=40 20 5 15,clip]{figures/experiment_results/legend_5.pdf}}
    \label{fig:subfig1}
    \end{subfigure}%
 \setlength{\abovecaptionskip}{-5pt}
 \caption{Performance comparison on the \emph{GB1(4)} dataset using batch sizes $B=10$ and $B=50$.
 }
\label{fig:batch-size-plot}
\end{figure}

We present additional results on the \emph{GB1(4)} dataset using higher batch sizes, $B=10$ and $B=50$. The results in Figure~\ref{fig:batch-size-plot} show that \textsc{GameOpt} variations still outperform baselines by discovering higher log fitness-valued protein sequences at a faster rate due to sampling diverse sets of batches. When batch size increases, \textsc{GameOpt-Hedge} becomes dominant in discovering the best protein sequence. This shows that irrespective of the batch size, \textsc{GameOpt} is effective in large combinatorial BO settings.

\begin{figure}[!htb]
    \centering  
    \begin{subfigure}{0.33\linewidth}
    \centering
    {\includegraphics[width=1\linewidth,trim=10 10 0 10,clip]{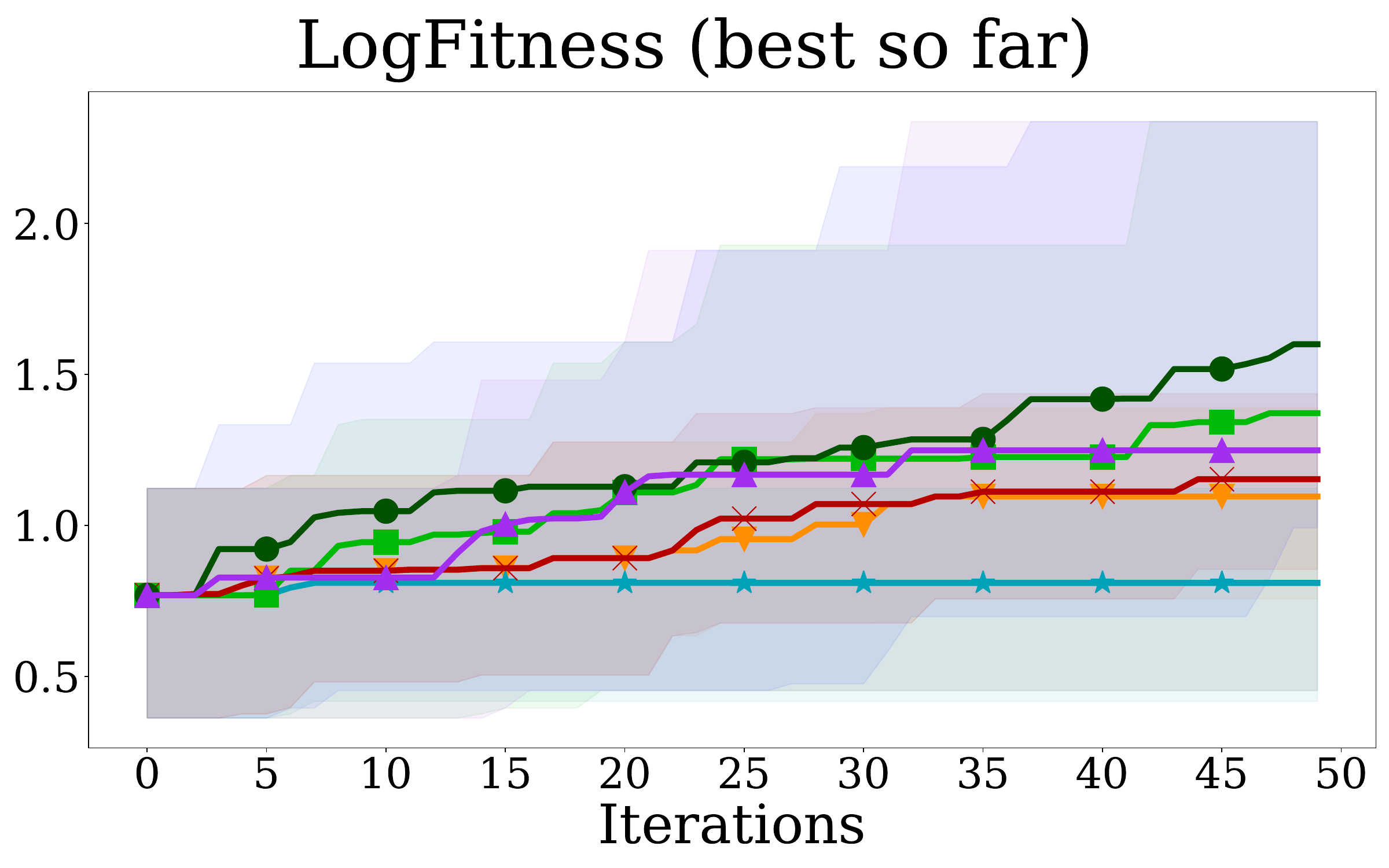}}
    \caption{\emph{Halogenase}, $B=1$.}
    \label{fig:avg-perf-plot-d}
    \end{subfigure}%
    \begin{subfigure}{0.33\linewidth}
    \centering
    {\includegraphics[width=1\linewidth,trim=10 10 0 10,clip]{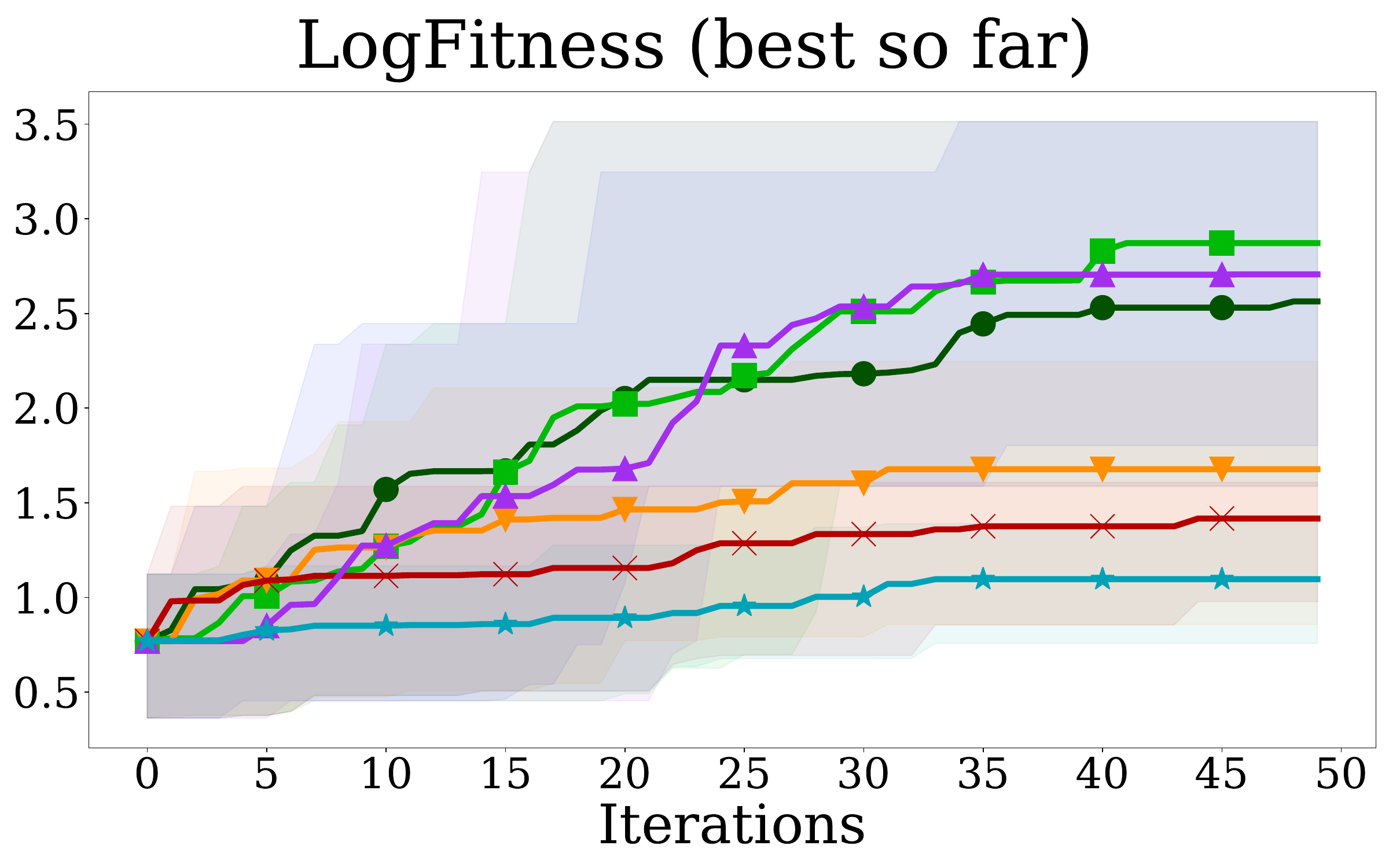}}
    \caption{\emph{Halogenase}, $B=3$.}
    \label{fig:avg-perf-plot-a}
    \end{subfigure}%
    \begin{subfigure}{0.33\linewidth}
    \centering
    {\includegraphics[width=1\linewidth,trim=10 10 0 10,clip]{figures/experiment_results/halogenase_batch_5_final_solid.pdf}}
    \caption{\emph{Halogenase}, $B=5$.}
    \label{fig:avg-perf-plot-a}
    \end{subfigure}%
    \\
    \begin{subfigure}{0.33\linewidth}
    \centering
    {\includegraphics[width=1\linewidth,trim=10 10 0 10,clip]{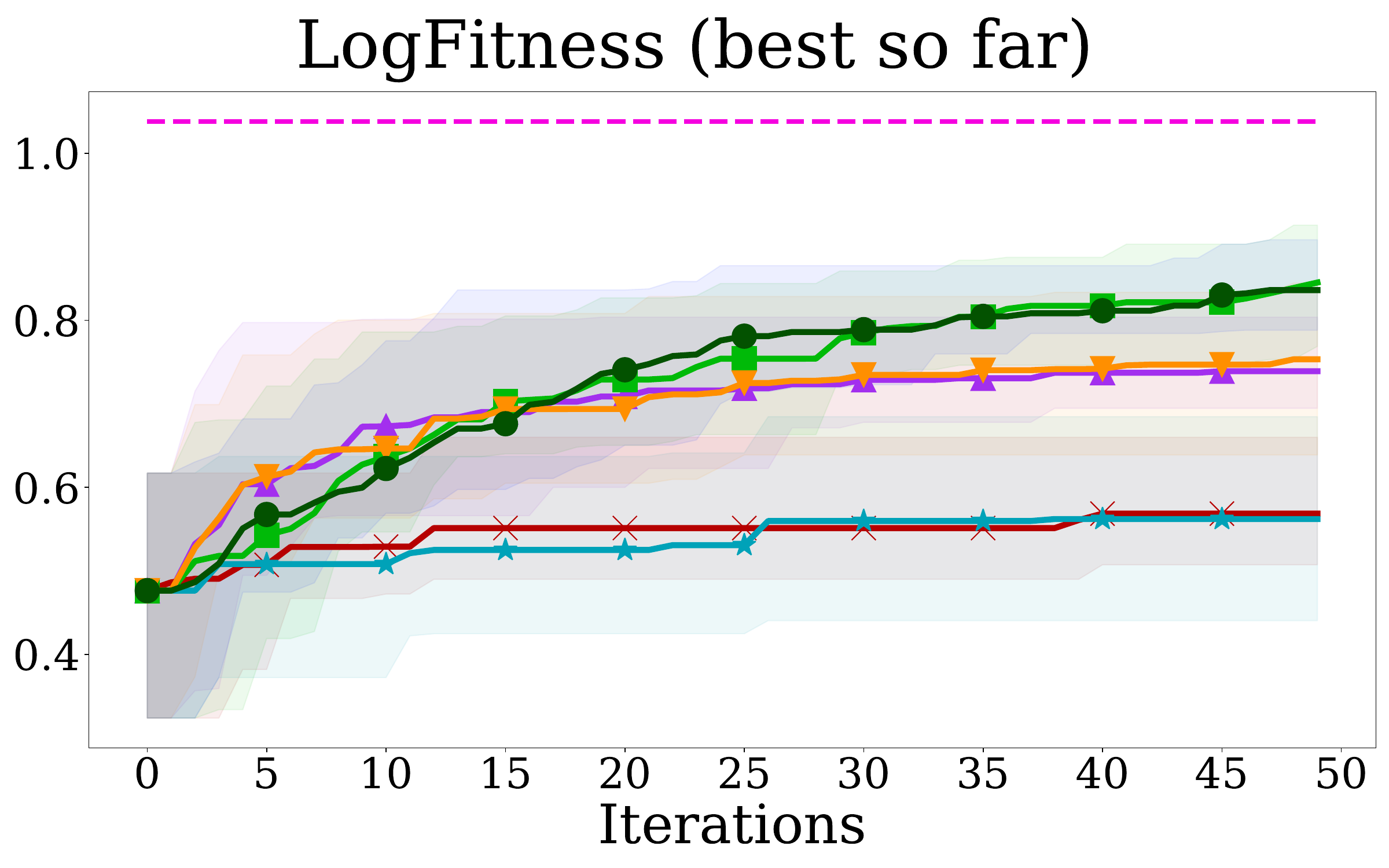}}
    \caption{\emph{GB1(4)}, $B=1$.}
    \label{fig:avg-perf-plot-d}
    \end{subfigure}%
    \begin{subfigure}{0.33\linewidth}
    \centering
    {\includegraphics[width=1\linewidth,trim=10 10 0 10,clip]{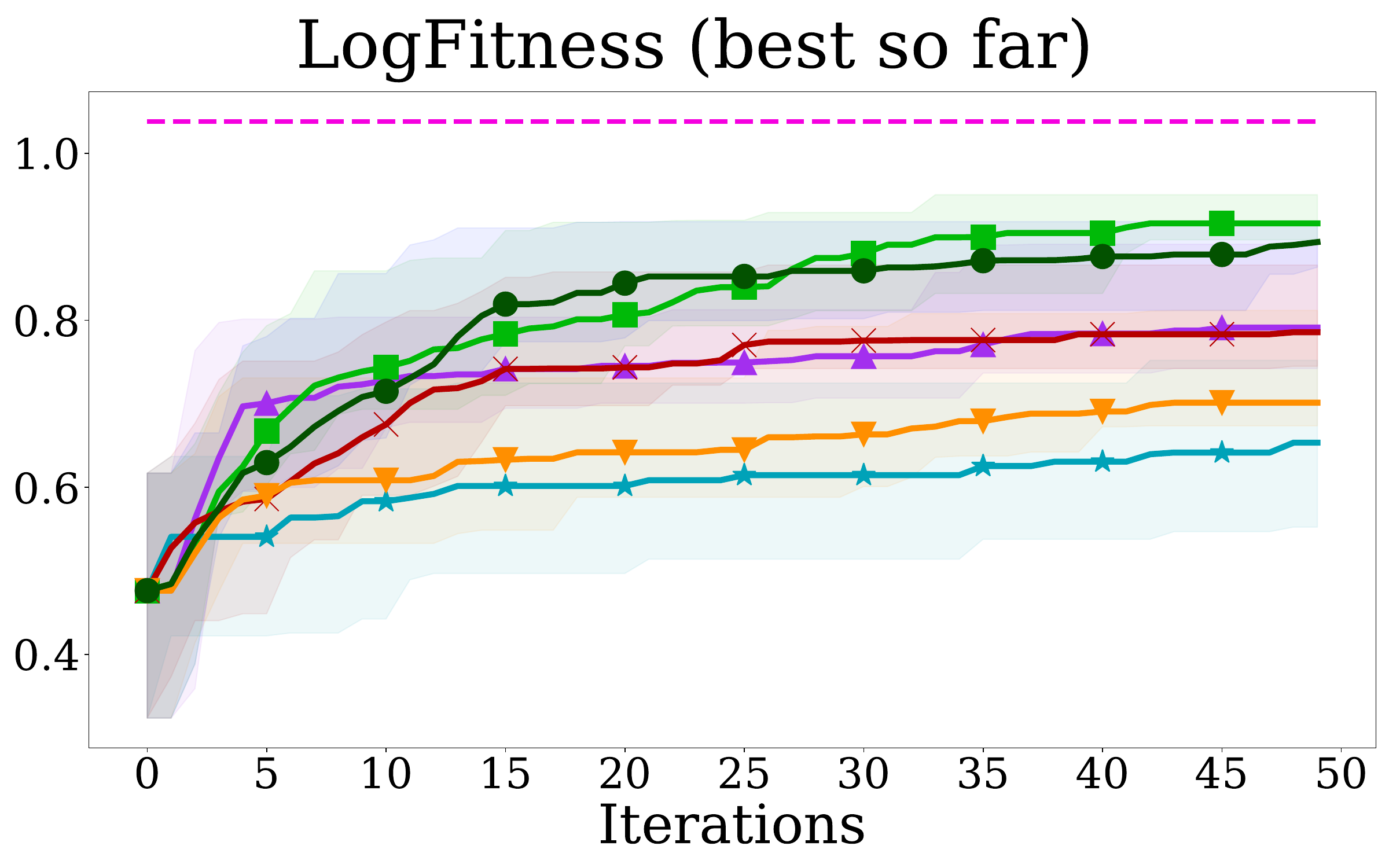}}
    \caption{\emph{GB1(4)}, $B=3$.}
    \label{fig:avg-perf-plot-a}
    \end{subfigure}%
    \begin{subfigure}{0.33\linewidth}
    \centering
    {\includegraphics[width=1\linewidth,trim=10 10 0 10,clip]{figures/experiment_results/gb1_4_batch_5_final_solid.pdf}}
    \caption{\emph{GB1(4)}, $B=5$.}
    \label{fig:avg-perf-plot-a}
    \end{subfigure}%
    \\
    \vspace{0.5em}
    \begin{subfigure}{1\linewidth}
    \centering
    {\includegraphics[width=1\linewidth,trim=40 20 5 15,clip]{figures/experiment_results/legend_5.pdf}}
    \label{fig:subfig1}
    \end{subfigure}%
 \setlength{\abovecaptionskip}{-5pt}
 \caption{Performance comparison on the \emph{Halogenase} and \emph{GB1(4)} datasets using batch sizes $B=\{1,3,5\}$. The results show that \textsc{GameOpt} outperforms baselines even under more restrictive batch size settings. Although initially surpassed by GP-UCB under batch size $B=1$, \textsc{GameOpt}'s better exploration compared to the GP-UCB's exploitative behavior avoids ending up at points with lower fitness values.}
 
\label{fig:batch-size-plot-2}
\end{figure}

We further present additional analysis using more restrictive batch sizes, $B=1$ and $B=3$ on the \emph{GB1(4)} and \emph{Halogenase} datasets. 
The results in Figure~\ref{fig:batch-size-plot-2} demonstrate that even under a more restricted setting on both problem domains, \textsc{GameOpt} variations achieve superior performance compared to the baselines. 
Although initially surpassed by GP-UCB under batch size $B=1$, \textsc{GameOpt}'s better exploration compared to the GP-UCB's exploitative behavior avoids ending up at points with lower fitness values.
As the batch size increases, \textsc{GameOpt}'s optimistic game approach benefits from parallelism and collects diverse local optima, hence, \textsc{GameOpt} performs significantly better against baselines.

\subsection{Learning rate ($\eta$) ablations for \textsc{GameOpt-Hedge}}
\label{app:lr-ablation-hedge}
\begin{figure}[!htb]
    \centering  
    \begin{subfigure}{0.33\linewidth}
    \centering
    {\includegraphics[width=1\linewidth,trim=10 10 0 10,clip]{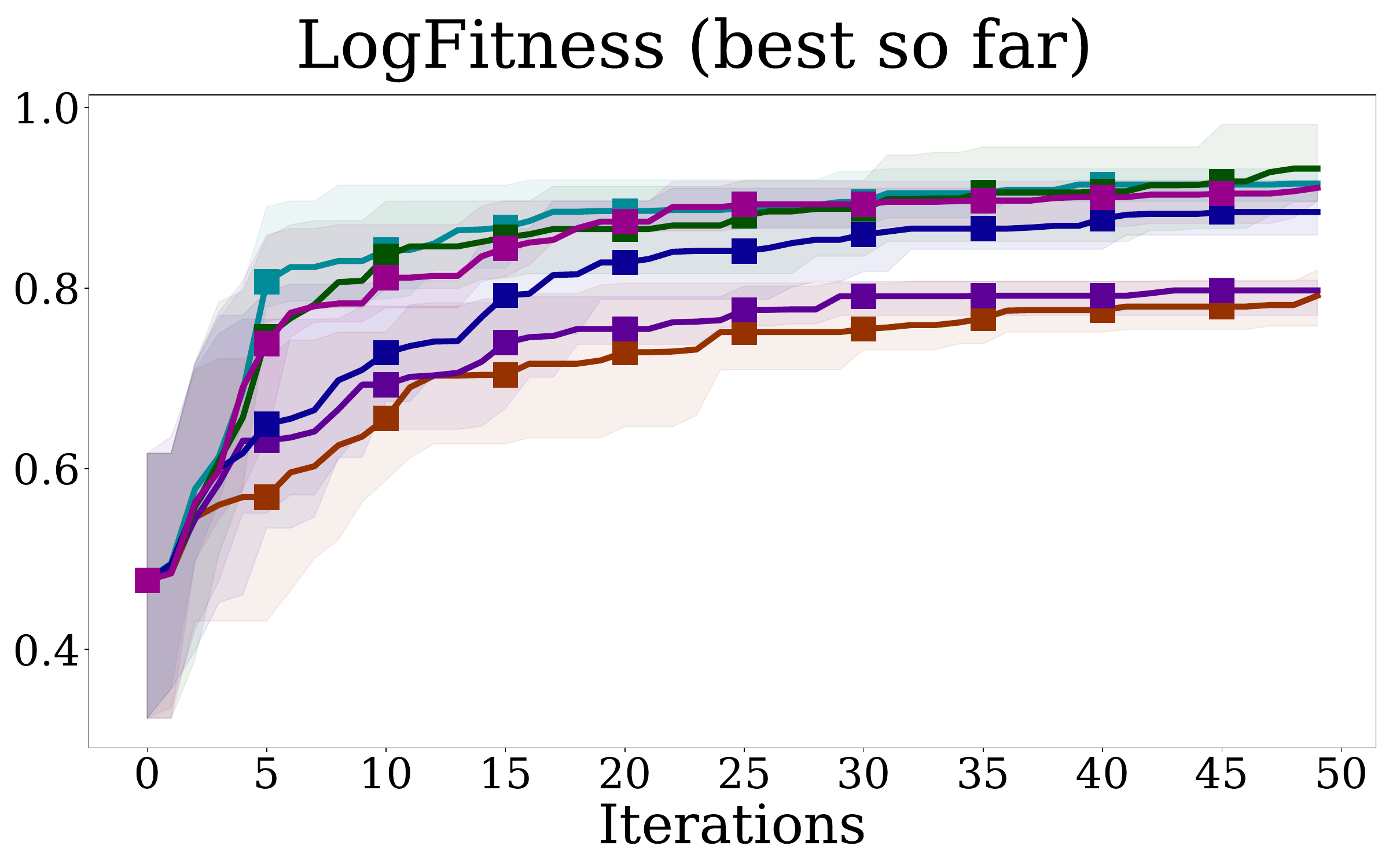}}
    \caption{LogFitness (BSF).}
    \label{fig:avg-perf-plot-a}
    \end{subfigure}%
    \begin{subfigure}{0.33\linewidth}
    \centering
    {\includegraphics[width=1\linewidth,trim=10 10 0 10,clip]{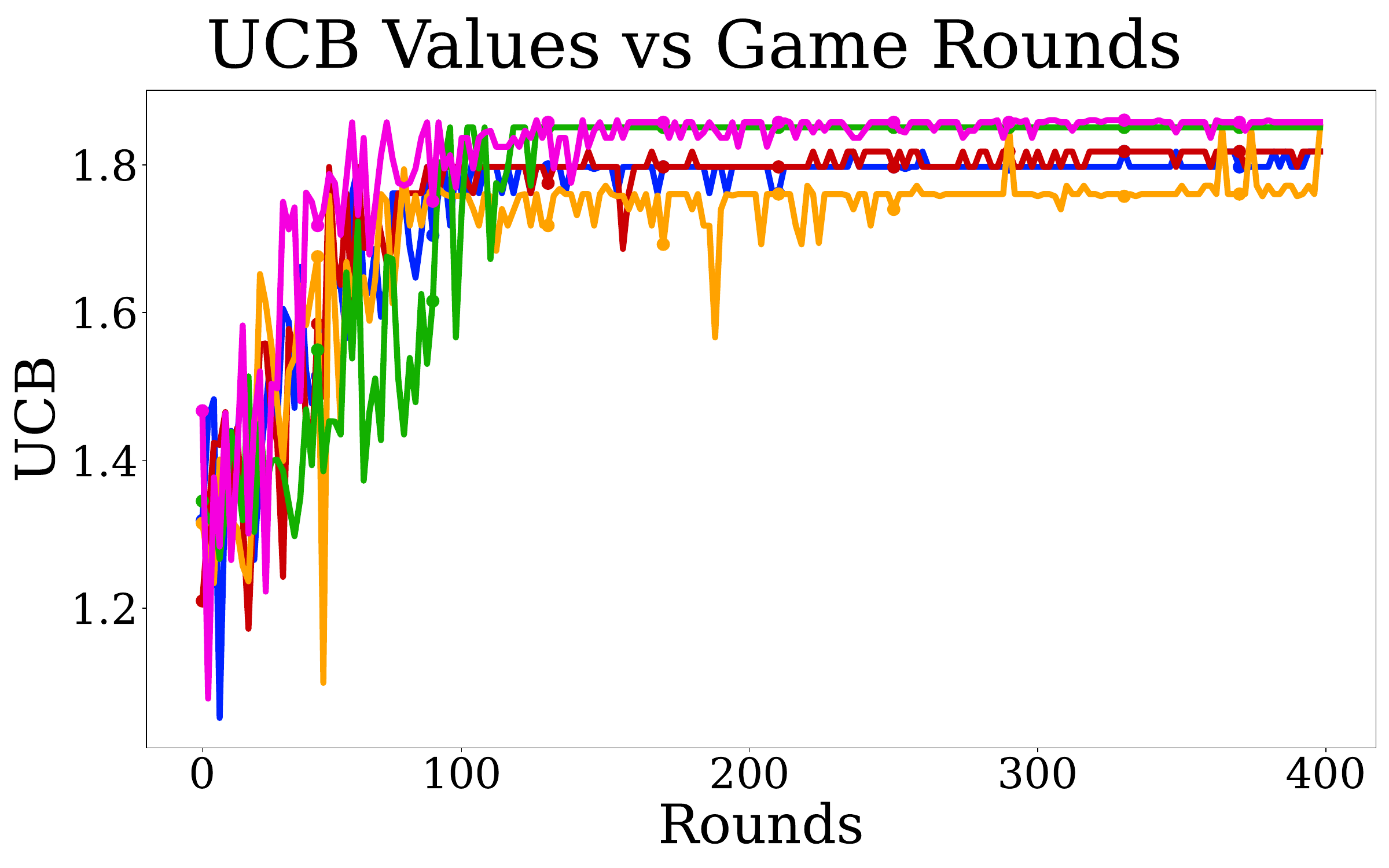}}
    \caption{Convergence of games.}
    \label{fig:convergence-of-games}
    \end{subfigure}%
    \\
    \vspace{0.5em}
    \begin{subfigure}{1\linewidth}
    \centering
    {\includegraphics[width=1\linewidth,trim=70 20 45 15,clip]{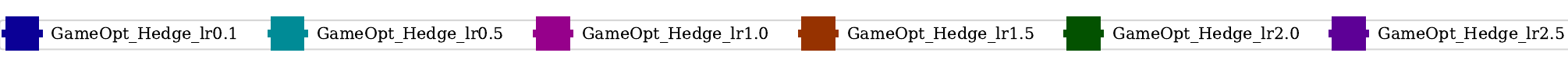}}
    \label{fig:subfig1}
    \end{subfigure}%
 \setlength{\abovecaptionskip}{-5pt}
 \caption{Performance of \textsc{GameOpt-Hedge} under different learning rate ($\eta$) values. The final value is set to ensure equilibrium convergence within a given number of rounds. As an example, we show in (\ref{fig:convergence-of-games}) the convergence of different games to equilibria when $\eta=2.0$.
 }
\label{fig:hedge-lr-plot}
\end{figure}

To select the learning rate ($\eta$) hyperparameter for the \textsc{GameOpt-Hedge} algorithm, we performed a hyperparameter sweep and tuned it accordingly.
Figure~\ref{fig:hedge-lr-plot} illustrates this process, showcasing the impact of different $\eta$ values on the game convergence. Based on this, we selected the optimal performing value.
In particular, Figure~\ref{fig:convergence-of-games} shows how varying $\eta$ influences the convergence to different equilibria at a BO iteration $t$. 
The results demonstrate that the chosen $\eta$ facilitates equilibrium convergence within a finite number of rounds, ensuring practical game convergence. \looseness=-1

\subsection{UCB values of the sampled batches}
\label{app:ucb-values-of-sampled-batches}
As discussed in Sections~\ref{sec:intro} and \ref{sec:further-limitations}, the novel insight and the main cause of benefit of the \textsc{GameOpt} framework lies in its ability to provide efficient (\textit{tractable}) acquisition function optimization by adopting a game-theoretical perspective.

\begin{figure}[!htb]
    \centering  
    \begin{subfigure}{0.33\linewidth}
    \centering
    {\includegraphics[width=1\linewidth,trim=10 10 0 10,clip]{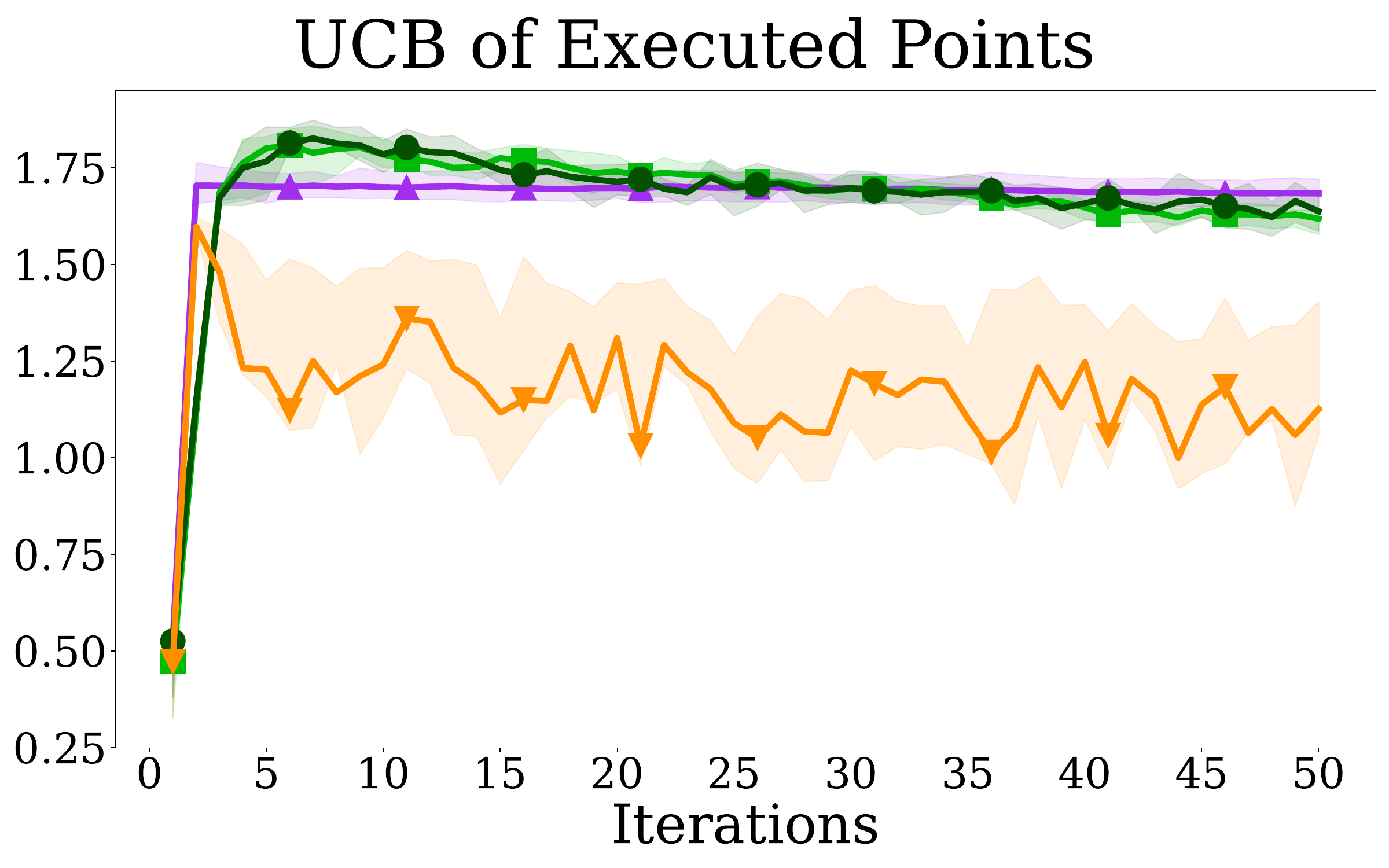}}
    \caption{Executed points.}
    \label{fig:avg-perf-plot-a}
    \end{subfigure}%
    \begin{subfigure}{0.33\linewidth}
    \centering
    {\includegraphics[width=1\linewidth,trim=10 10 0 10,clip]{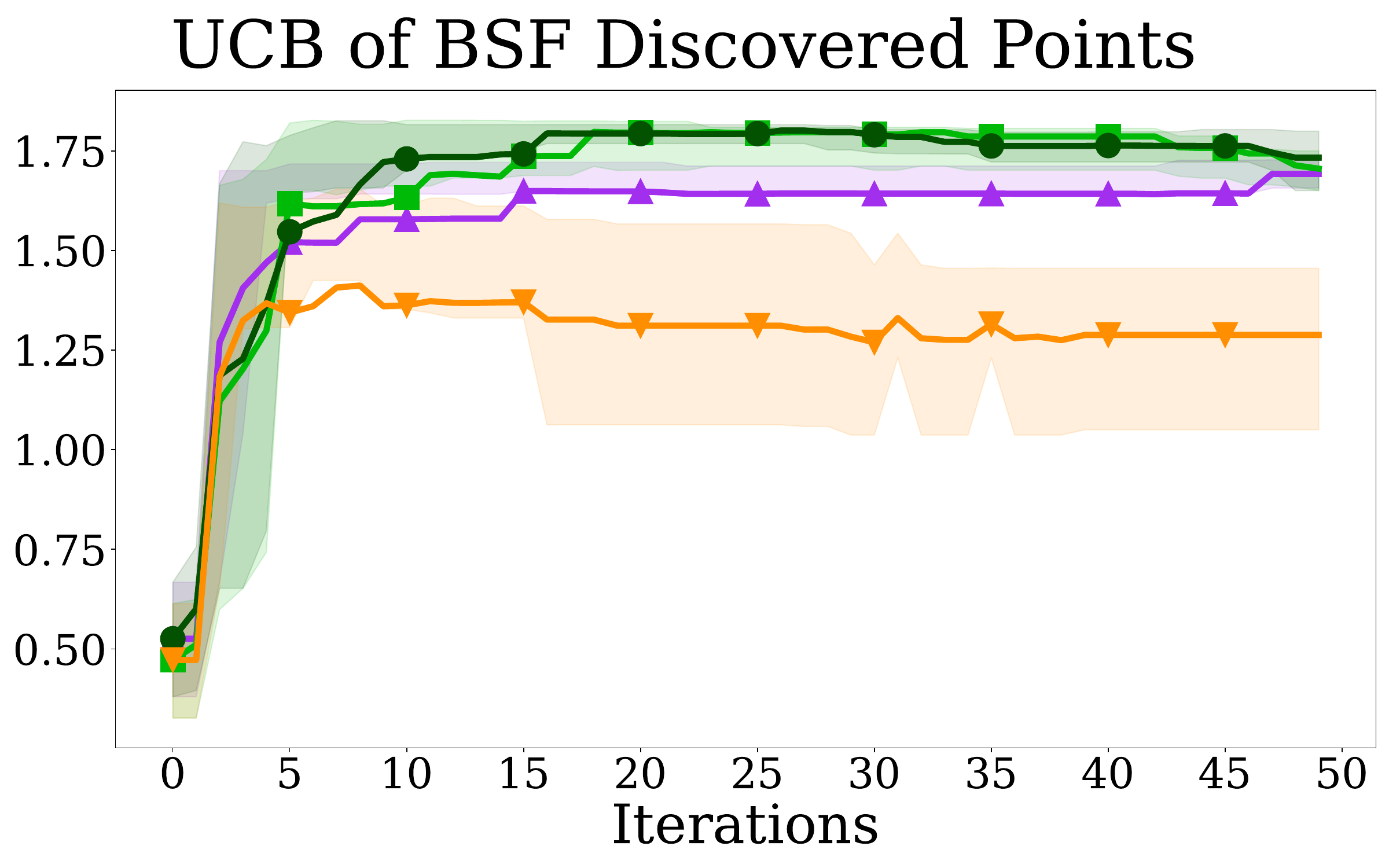}}
    \caption{BSF logfitness valued point.}
    \label{fig:avg-perf-plot-d}
    \end{subfigure}%
    \\
    \vspace{0.5em}
    \begin{subfigure}{0.7\linewidth}
    \centering
    {\includegraphics[width=1\linewidth,trim=0 20 450 15,clip]{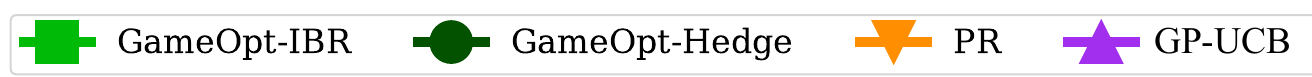}}
    \label{fig:subfig1}
    \end{subfigure}%
 \setlength{\abovecaptionskip}{-5pt}
 \caption{UCB value of the (a): executed point (max UCB), and (b): best-so-far (BSF) logfitness valued point over iterations under \emph{GB1(4)} domain. Although initially \textsc{GameOpt} variations executed points with smaller UCB values compared to GP-UCB, its better exploration helps to identify higher UCB valued points quickly. 
 }
\label{fig:ucb-values-of-sampled-batch}
\end{figure}

When comparing the methods based on the UCB values over BO iterations in the \emph{GB1(4)} domain, as demonstrated in Figure~\ref{fig:ucb-values-of-sampled-batch}, we observe that \textsc{GameOpt} initially selects points with lower UCB values compared to GP-UCB. However, in later iterations, \textsc{GameOpt} identifies points with higher UCB values. This improvement is driven by the framework's leverage of equilibrium points, leading to superior exploration of the search space and ultimately more efficient optimization.

\subsection{Expected Improvement acquisition function}
\label{app:ei-acquisition}
\begin{figure}[!h]
    \centering  
    \begin{subfigure}{0.33\linewidth}
    \centering
    {\includegraphics[width=1\linewidth,trim=10 10 0 10,clip]{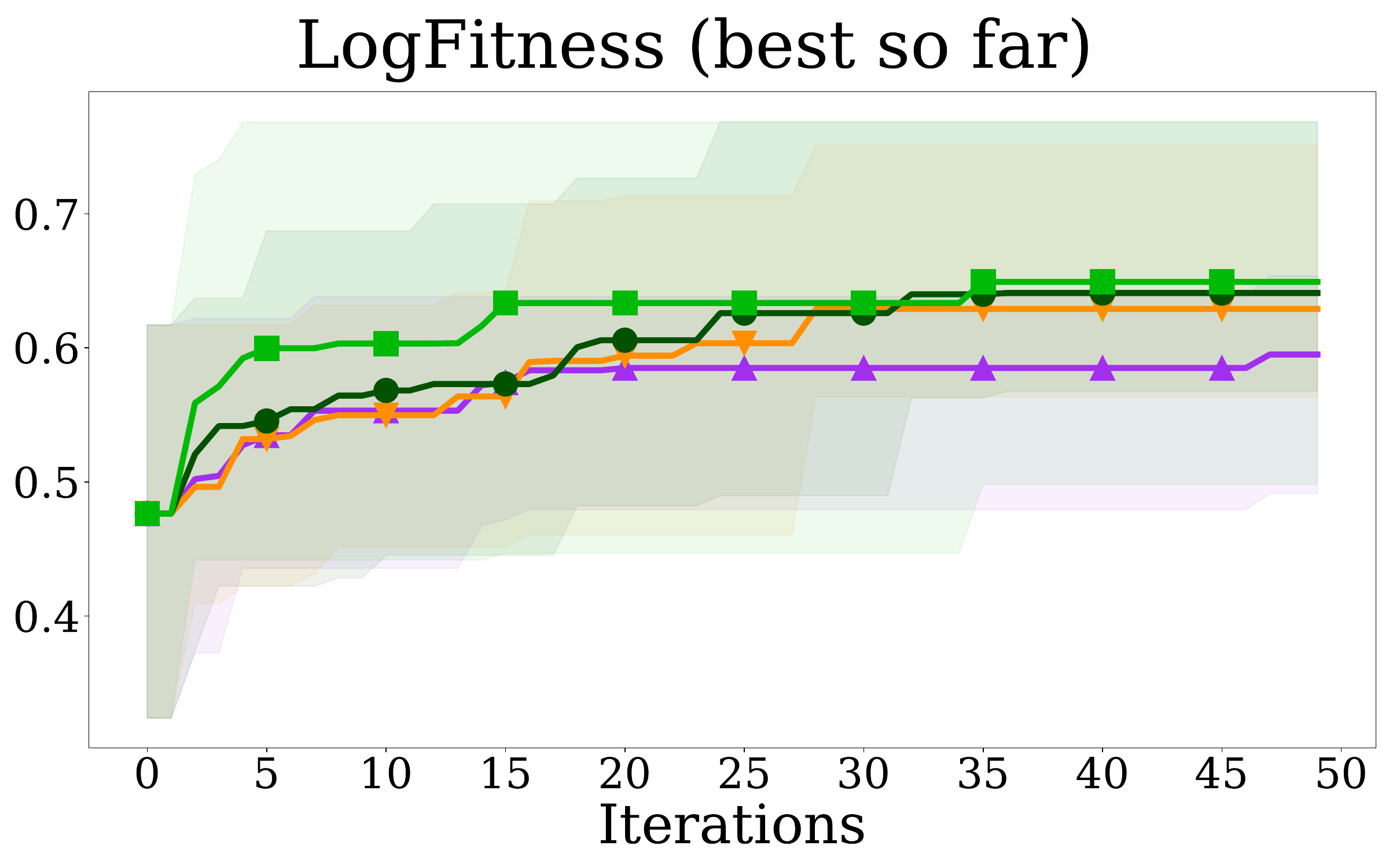}}
    \caption{Convergence speed (BSF).}
    \label{fig:avg-perf-plot-d}
    \end{subfigure}%
    \begin{subfigure}{0.33\linewidth}
    \centering
    {\includegraphics[width=1\linewidth,trim=10 10 0 10,clip]{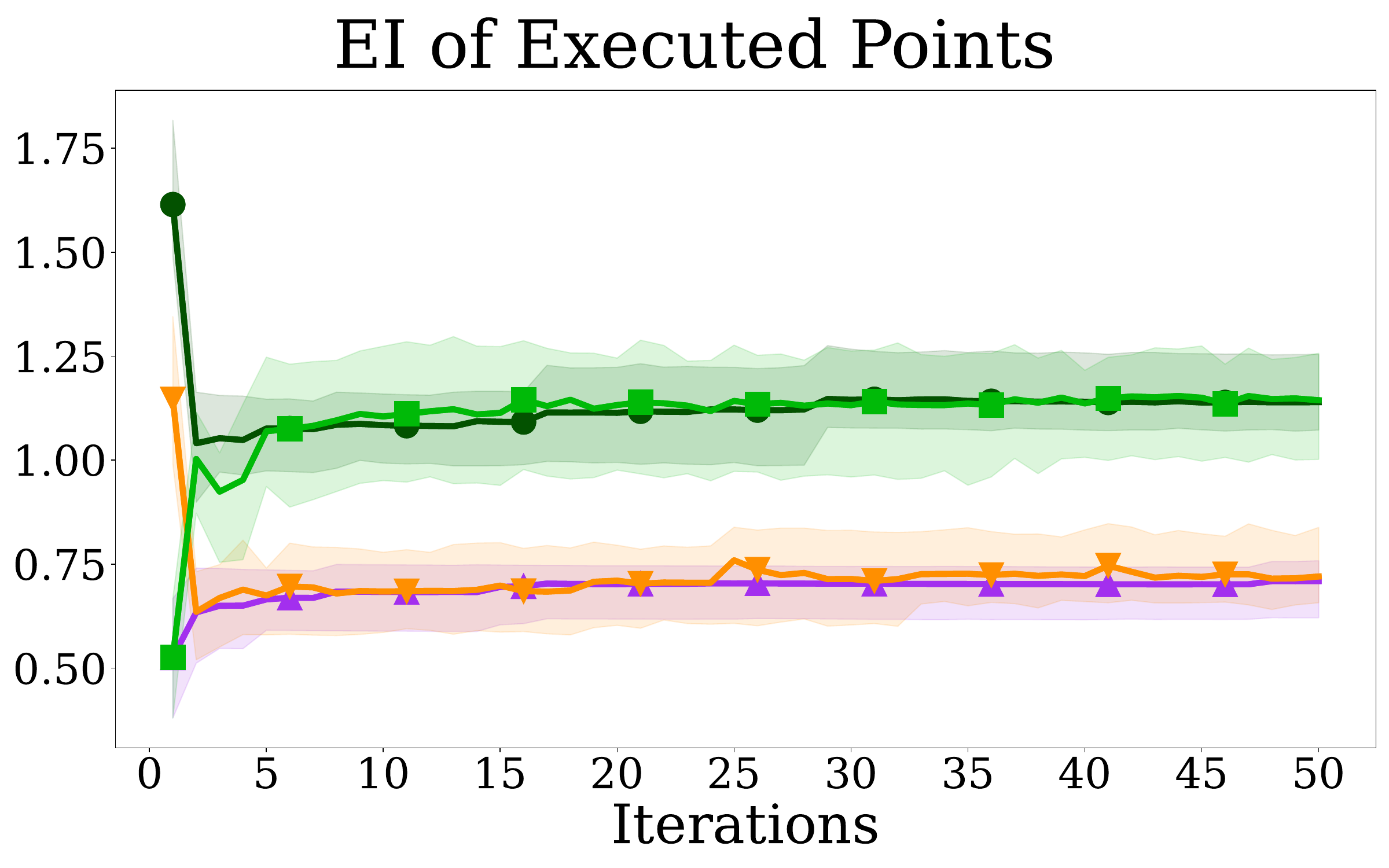}}
    \caption{EI of executed points.}
    \label{fig:avg-perf-plot-d}
    \end{subfigure}%
    \\
    \begin{subfigure}{0.33\linewidth}
    \centering
    {\includegraphics[width=1\linewidth,trim=10 10 0 10,clip]{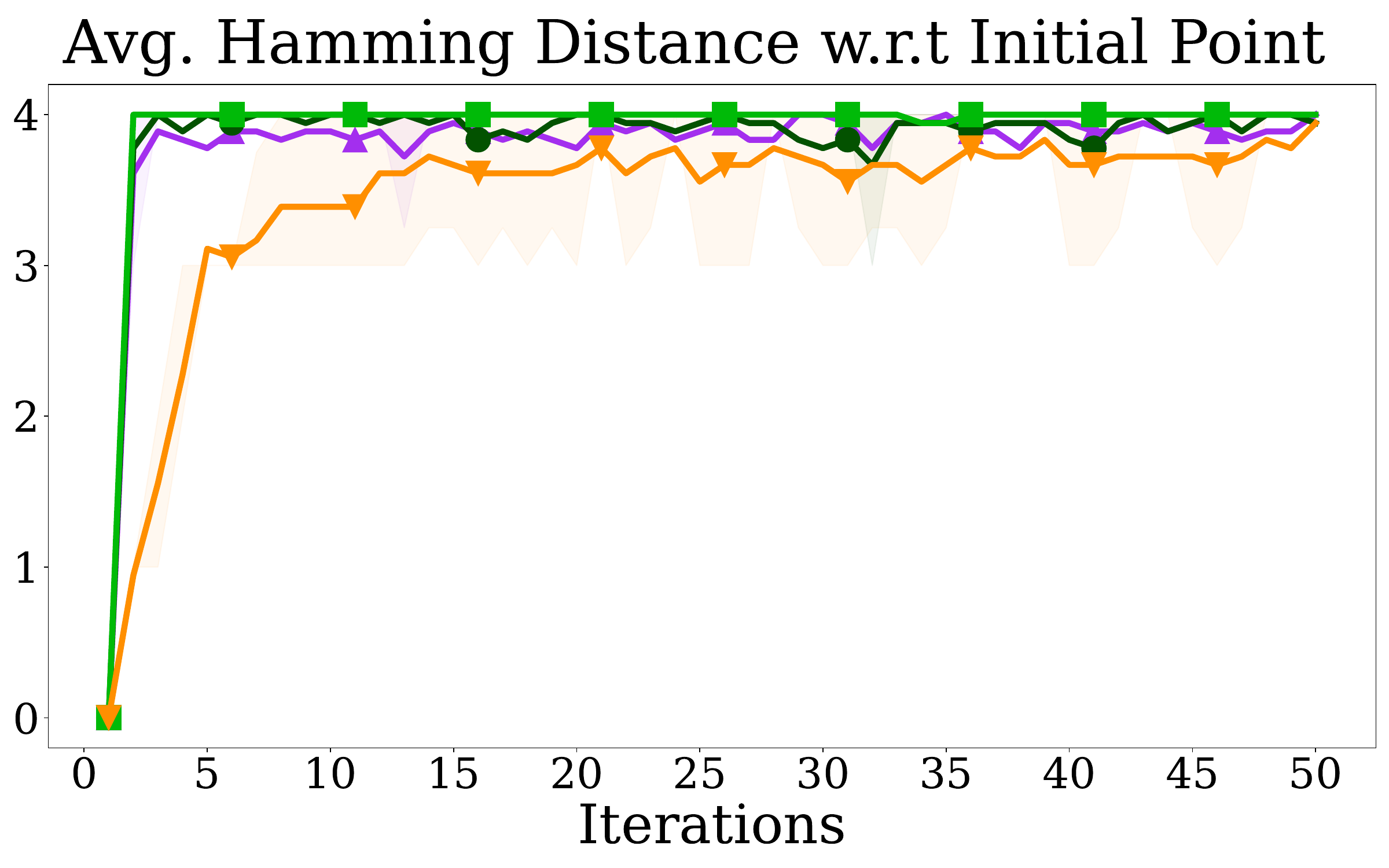}}
    \caption{Batch diversity (w.r.t. past).}
    \label{fig:ei-batch-diversity-wrt-past}
    \end{subfigure}%
    \begin{subfigure}{0.33\linewidth}
    \centering
    {\includegraphics[width=1\linewidth,trim=10 10 0 10,clip]{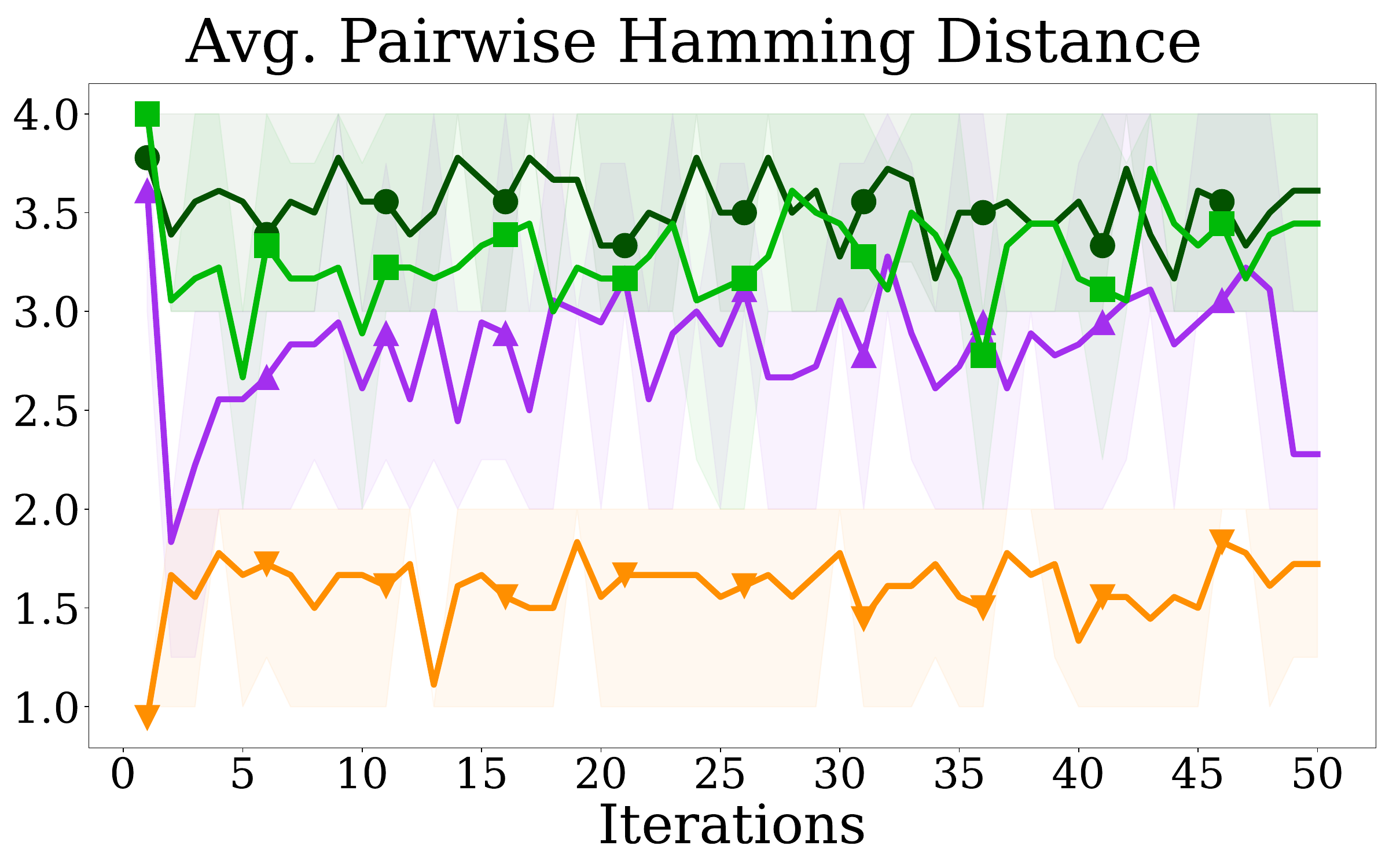}}
    \caption{Batch diversity (pairwise).}
    \label{fig:ei-batch-diversity-pairwise}
    \end{subfigure}%
    \\
    \vspace{0.5em}
    \begin{subfigure}{0.7\linewidth}
    \centering
    {\includegraphics[width=1\linewidth,trim=0 20 450 15,clip]{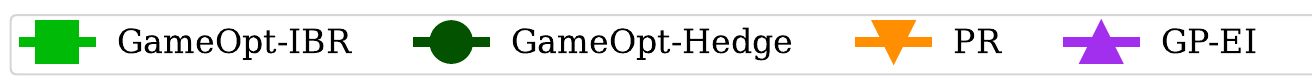}}
    \label{fig:subfig1}
    \end{subfigure}%
 \setlength{\abovecaptionskip}{-5pt}
 \caption{Performance comparison of GP-based methods on the \emph{GB1(4)} dataset using expected improvement (EI) acquisition function and batch size $B =5$. 
 In all experiments \textsc{GameOpt} with \textsc{Ibr} and \textsc{Hedge} subroutines discover better and more diverse protein sequences at a much faster rate.
 }
\label{fig:ei-af-plot}
\end{figure}

Although \textsc{GameOpt} is designed using the UCB acquisition function with a sample complexity guarantee, we provide a further empirical analysis across GP-based methods using a different acquisition function:  expected improvement (EI) \citep{jones1998efficient}. 
As demonstrated by the results on the \emph{GB1(4)} dataset in Figure~\ref{fig:ei-af-plot}, \textsc{GameOpt} variations show superior performance compared to other GP-based baselines. They sample more diverse batches, as given in plots for (\ref{fig:ei-batch-diversity-wrt-past}) sample batch diversity with respect to past and (\ref{fig:ei-batch-diversity-pairwise}) previously executed points.

\subsection{Comparison with latent space optimizers}
\label{app:comp-w-latent-space-optimizers}
Our baselines involve acquisition function optimizers which \textit{directly} operate on large \textit{combinatorial} search spaces. While there is a body of work employing latent space optimizers \citep{gomez2018automatic, tripp2020sample, deshwal2021combining, maus2022local, bo-seq-ae}, our primary focus in this study is to \textit{tractably} optimize the acquisition function directly on combinatorial search spaces, addressing the challenge of exhaustive exploration.

\begin{figure}[h]
    \centering  
    \begin{subfigure}{0.35\linewidth}
    \centering
    {\includegraphics[width=1\linewidth,trim=10 10 0 10,clip]{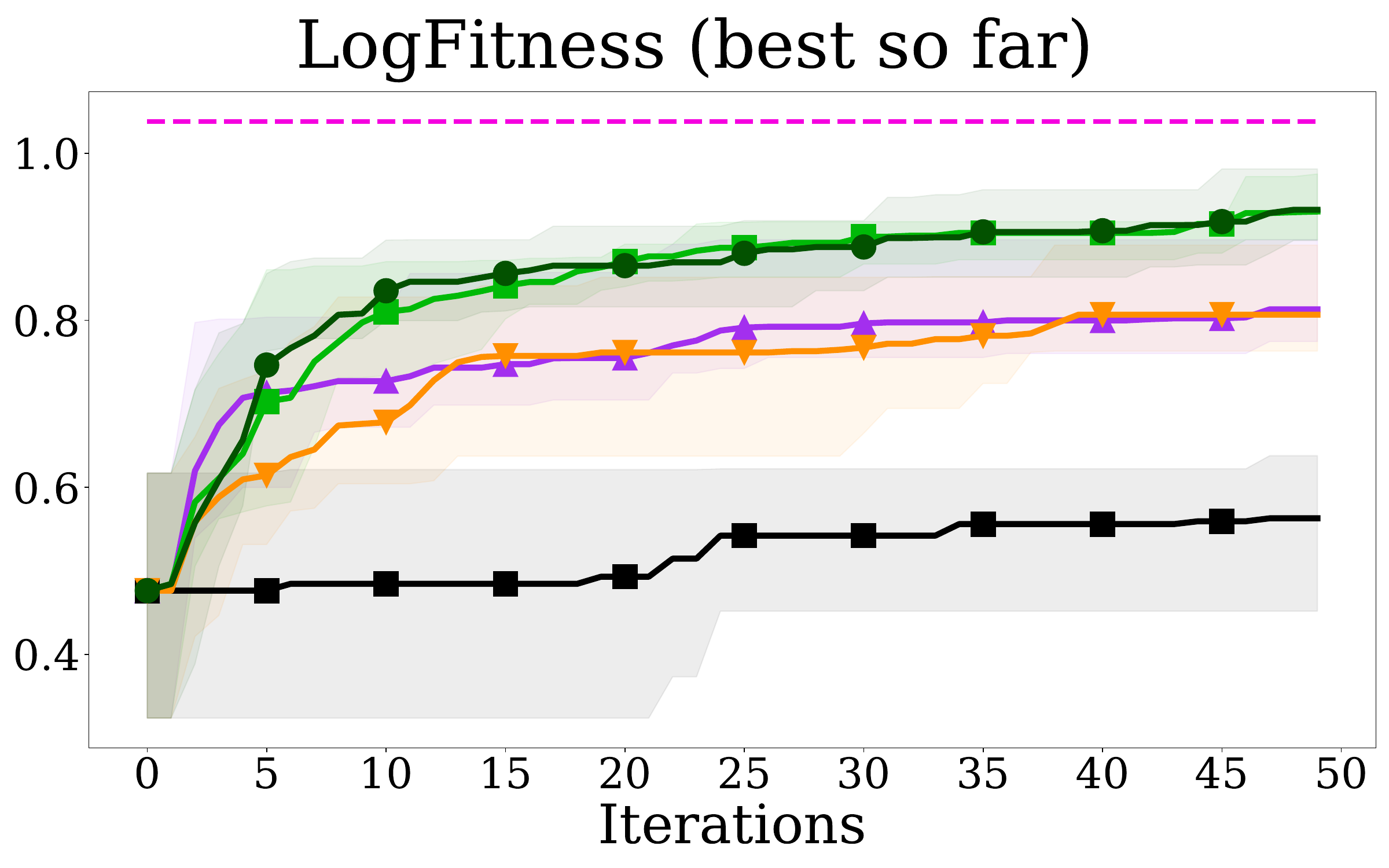}}
    \caption{RBF kernel.}
    \label{fig:latent-fitness-rbf}
    \end{subfigure}%
    \begin{subfigure}{0.35\linewidth}
    \centering
    {\includegraphics[width=1\linewidth,trim=10 10 0 10,clip]{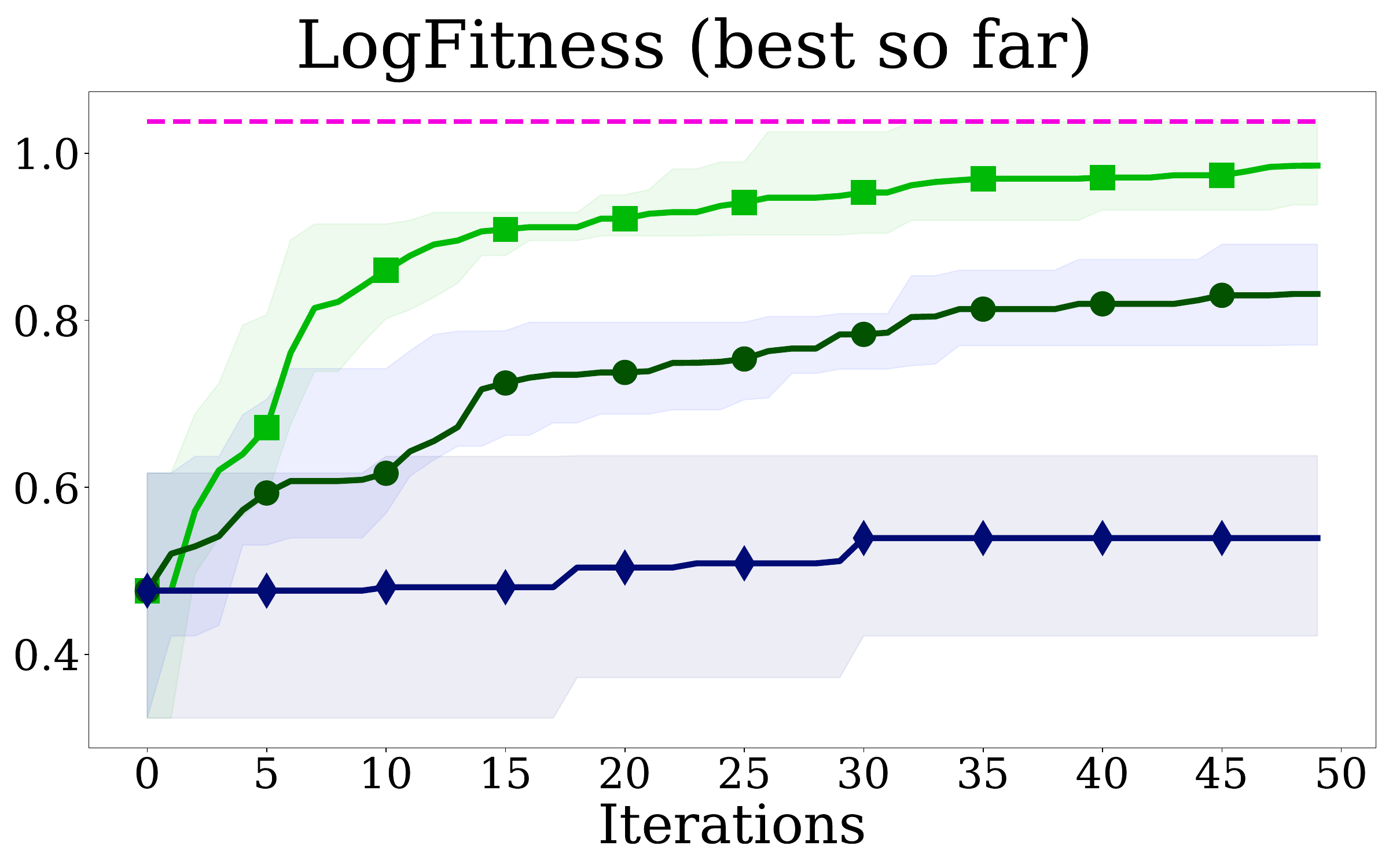}}
    \caption{Structure-coupled kernel.}
    \label{fig:latent-fitness-structure-kernel}
    \end{subfigure}%
    \\
    \begin{subfigure}{0.35\linewidth}
    \centering
    {\includegraphics[width=1\linewidth,trim=10 10 0 10,clip]{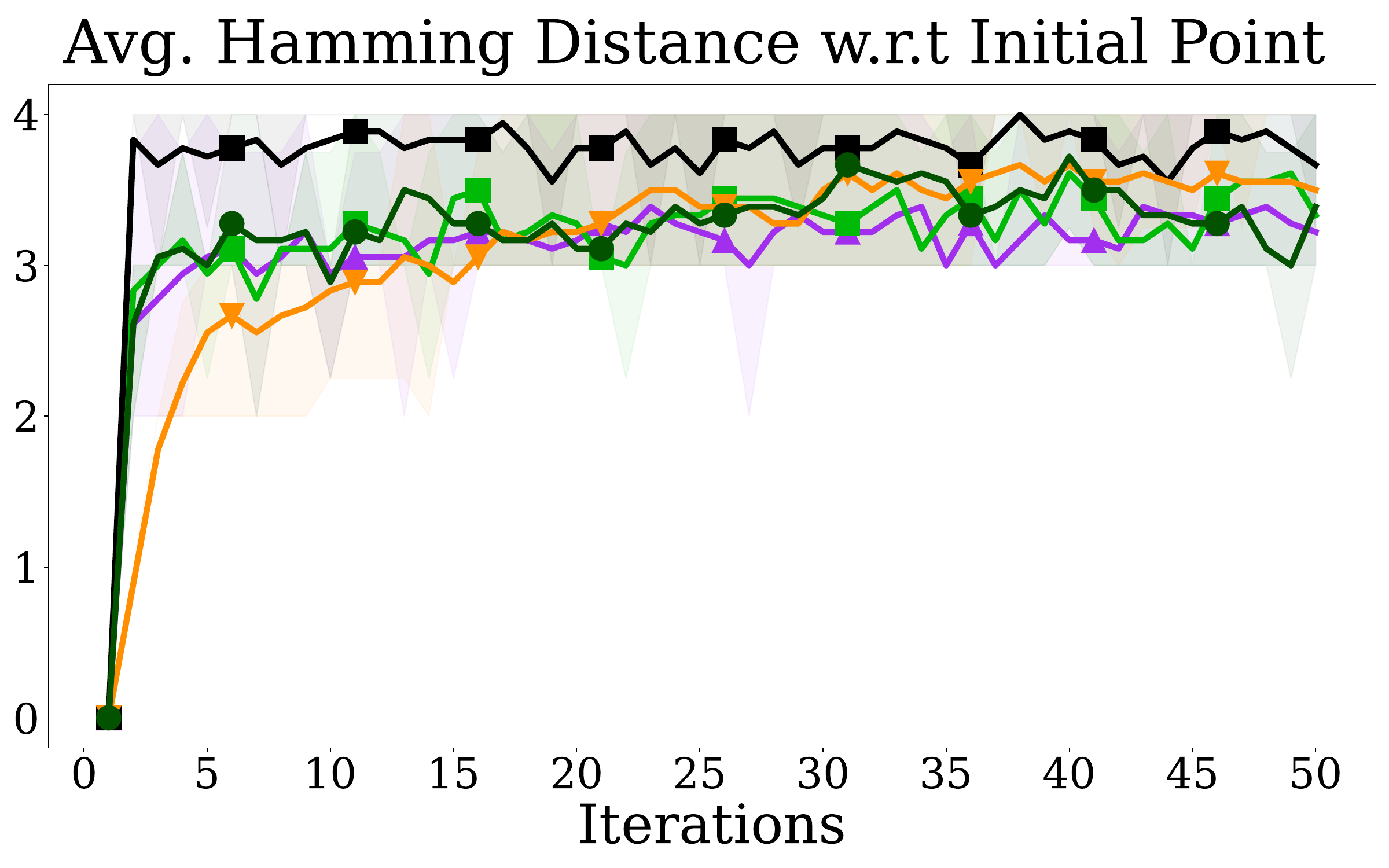}}
    \caption{RBF kernel.}
    \label{fig:latent-hamming-rbf}
    \end{subfigure}%
    \begin{subfigure}{0.35\linewidth}
    \centering
    {\includegraphics[width=1\linewidth,trim=10 10 0 10,clip]{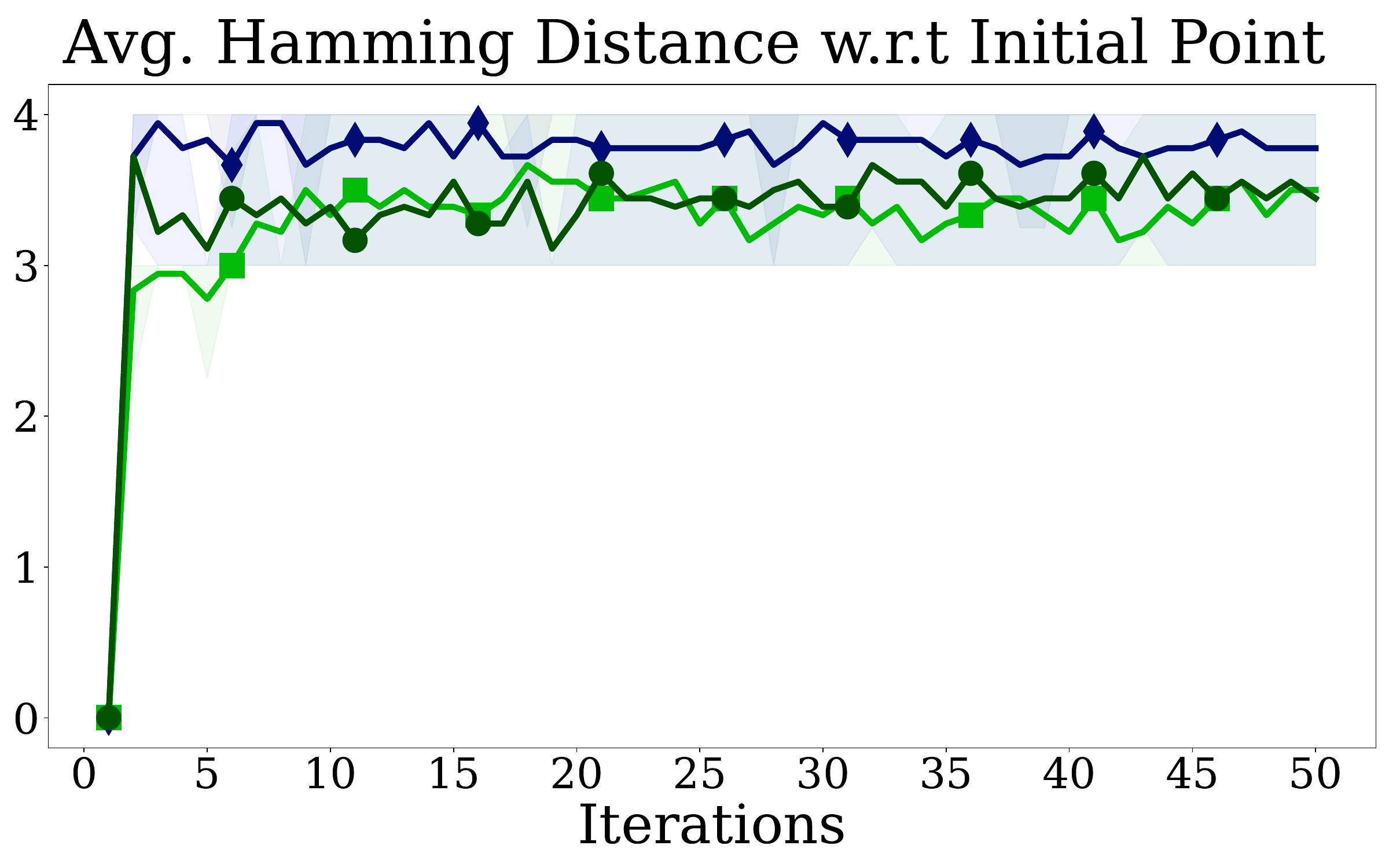}}
    \caption{Structure-coupled kernel.}
    \label{fig:latent-hamming-structure-kernel}
    \end{subfigure}%
    \\
    \vspace{0.5em}
    \begin{subfigure}{1\linewidth}
    \centering
    {\includegraphics[width=1\linewidth,trim=30 20 5 15,clip]{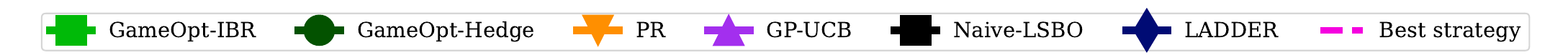}}
    \label{fig:subfig1}
    \end{subfigure}%
 \setlength{\abovecaptionskip}{-5pt}
\caption{Performance comparison of \textsc{GameOpt} against latent space optimizers: Na\"{i}ve LSBO-(L-BFGS-B) \citep{gomez2018automatic} and LADDER \citep{deshwal2021combining} methods under \emph{GB1(4)} domain with batch size $B=5$, using kernels: RBF and structure-coupled kernel \citep{moss2020boss}. Latent space optimizers perform poorly mainly because they rely on a decoder, which \textsc{GameOpt} variations eliminate and directly (tractably) optimize over combinatorial space.  }
\label{fig:com-w-latent-space}
\end{figure}

However, to provide insight, we compare \textsc{GameOpt} against Na\"{i}ve LSBO-(L-BFGS-B) with second-order gradient-based optimizer \citep{gomez2018automatic} and LADDER \citep{deshwal2021combining} methods. The comparison, presented in Figure~\ref{fig:com-w-latent-space}, is performed using the (\ref{fig:latent-fitness-rbf},\ref{fig:latent-hamming-rbf}) RBF kernel and (\ref{fig:latent-fitness-structure-kernel},\ref{fig:latent-hamming-structure-kernel}) a structure-coupled kernel. The structure-coupled kernel is designed using an RBF kernel and a sub-sequence string kernel \citep{moss2020boss}. 
Results indicate that \textsc{GameOpt} variations perform significantly better than the considered latent space optimizers. A notable limitation of latent space optimizers is their dependence on a decoder. In our experiments, we trained a transformer-based decoder using ESM-1v feature embeddings \citep{ESM-1v}, employing beam search for sequence generation. In contrast,  our \textsc{GameOpt} approach does not require such an additional decoder and \textit{directly} optimizes over the sequence space \textit{efficiently}.

\end{document}